\documentclass[journal]{IEEEtran}

\usepackage{cite}
\usepackage{graphicx}
\usepackage{amsmath,amssymb,amsfonts}
\usepackage{algorithmic}
\usepackage{textcomp}
\usepackage{xcolor}
\usepackage{url}
\usepackage[hidelinks]{hyperref}
\usepackage[numbers,sort&compress]{natbib}
\usepackage{bm}
\usepackage{rotating}
\usepackage{hyperref}
\usepackage{multirow}
\usepackage{booktabs}

\begin{document}

\title{Autonomous fixed-wing interception using reinforcement learning}

\author{Daan van Hamersveld, Ewoud Smeur, Iurii Kriachko}

\markboth{D.F. van Hamersveld et al. - MSc. Thesis, Aerospace Engineering, Delft University of Technology, 2026}%
{D.F. van Hamersveld et al.: Autonomous interception of fixed-wing drones}

\maketitle

\begin{abstract}
Medium altitude, long endurance unmanned aerial systems such as Shahed-type strike drones are an unanswered threat in modern conflicts. Conventional air defense systems are too expensive to provide a feasible large-scale solution. Quadcopter interceptors are a cheap, battle-proven alternative, but their extremely limited flight time restricts them to low altitude targets at short range, whereas fixed-wing platforms have the endurance to reach higher altitude targets from longer distances. However, such interceptors are still operated manually. Recent developments in learning-based controllers show potential for automation. In this work, a fixed-wing interceptor is presented that is controlled by a neural network that maps filtered monocular vision observations directly to actuator commands. The policy was trained in a custom Python environment using reinforcement learning and proximal policy optimization. The policy was then deployed on a fixed-wing platform and tested in real interception scenarios. As a baseline, two classical guidance controllers were also developed and tested in simulation. 
The policy is able to intercept targets both in simulation and in real flight. In simulation, the policy outperforms the classical controllers under most conditions. The interceptor was ultimately successfully validated in real flight by intercepting maneuvering fixed-wing targets. This novel system shows the potential of artificial intelligence for autonomous control of fixed-wing interceptor drones. 
\end{abstract}


\section{Introduction}

Medium Altitude Long Endurance (MALE) Unmanned Aerial Systems (UAS) pose a credible threat in modern warfare \cite{Kunertova2023DronesUkraine}\cite{Bilousova2023SanctionsDrones}. In both the Ukraine-Russia conflict as well as the recent Iran-Israel/US war, the systems used can be classified into two broad groups. Firstly, intelligence, surveillance and reconnaissance drones, which are generally fixed-wing UAS with varying imaging capabilities. They typically fly at 2-3 km altitude at speeds up to 150 km/h. Secondly, Shahed-type strike drones, which are delta-wing kamikaze drones that typically fly at 1-2 km altitude at speeds reaching above 250 km/h. They carry a large explosive payload and have high endurance, allowing for long-range (up to 2000 km) targeted ground strikes.

To answer these threats, low cost interceptor drones are emerging as an alternative to expensive conventional air defenses such as fighter jets and missiles. Interceptors carry an explosive payload which is detonated when close enough to neutralize the target. Hence, such systems are single-use, meaning a minimal onboard sensor suite is preferred for affordability. A ground-based radar system can typically provide a mid-course guidance target lock with a few hundred meters error. However, terminal guidance requires meter-level accuracy and hence is currently mainly vision-based, as monocular cameras are inexpensive and easy to source. Traditional guidance laws such as Proportional Navigation (PN) and Pure Pursuit (PP) \cite{shneydor1998missile}\cite{yanushevsky2018modern}\cite{zarchan2012tactical} need a constant Line Of Sight (LOS) and struggle against maneuvering targets. Consequently, current interceptors are mostly manually controlled during terminal guidance.

\begin{figure}[!t]
\centering
\includegraphics[width=0.9\columnwidth]{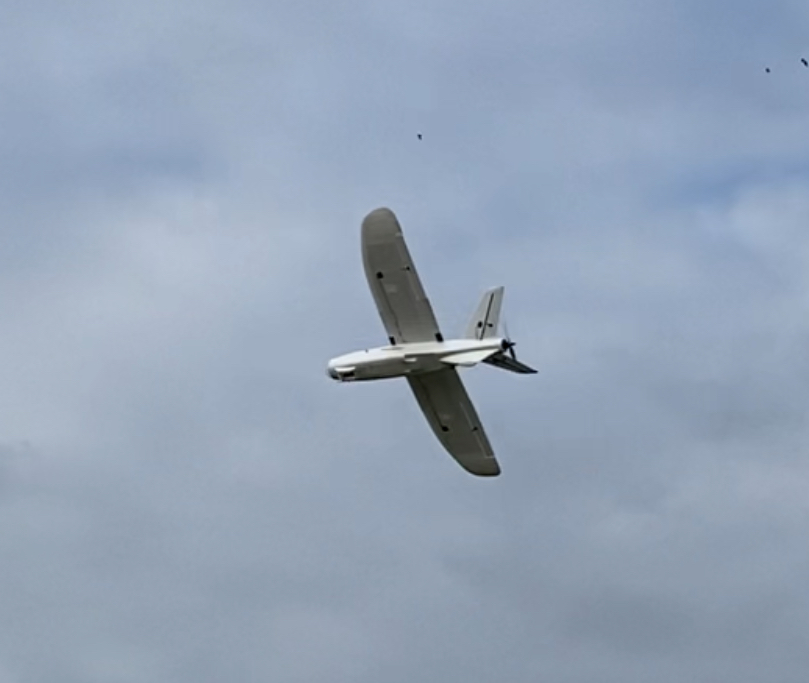}
\caption{The interceptor during a high bank angle.}
\label{fig_Banking}
\end{figure}

Vision-based relative state estimation can be done in many ways for a variety of applications, including drone racing \cite{Kaufmann2023Champion}, target tracking \cite{pothuri2025learningapproachefficientvisionbased}, aerial refueling \cite{Martinez2013VisionAerialRefueling} and also interception \cite{jana2022interception}\cite{patel2025generalpurposemethodrobotic}\cite{vanderAa2024IBVSquadcopterInterception}. For interception purposes, the bounding-box \cite{Barisic_2019}\cite{KFegoMotionGroundTargets2025Kim}\cite{vanderAa2024IBVSquadcopterInterception} is an effective and easy to implement proxy to a relative state. Turning visual observation into effective guidance for interception has been done using classical control approaches such as PP \cite{jana2022interception} and PN \cite{yan2025preciseinterceptionflighttargets}.

Recently, Reinforcement Learning (RL) has been successfully applied to UAS control tasks such as stabilization \cite{Hwangbo2017QuadrotorRL}, attitude control \cite{Koch2018RLUAV}, acrobatics \cite{kaufmann2020deepdroneacrobatics}, drone racing \cite{bahnam2026monoracewinningchampionleveldrone}\cite{Kaufmann2023Champion} and also interception \cite{pothuri2025learningapproachefficientvisionbased}\cite{Capelle2025DroneInterception}\cite{GIRAL2026110985}, with controller authority varying from high level planning to direct actuator command. While promising, for interception the real challenge lies in transferring these results from simulation to a real, uncontrolled environment. Recent work suggests that end-to-end learning-based controllers can achieve high sim-to-real transfer robustness by implementing domain randomization \cite{Ferede2025OneNet}.

In MALE interception scenarios, fixed-wing drones can be superior to quadrotors due to their better endurance, allowing them to intercept targets from longer ranges and at higher altitudes. Studies on RL controllers for fixed-wing platforms are sparse. However, obstacle avoidance \cite{Chai2024MonocularObstacleAvoidance}, path following \cite{Marquis2025RARLFixedWing} evasion \cite{Zhuang2024} and interception with perfect state knowledge \cite{GIRAL2026110985} have been carried out in simulation. In real environments, studies have demonstrated successful attitude holding \cite{B_hn_2024} and landing \cite{Fletcher2021PerchedLandingRL}.

Fixed-wing, vision-based interception using a learning-based controller in a real environment is lacking in the literature. Most interception studies are done in simulation and using a quadrotor model, whilst realistic perception and sim-to-real transfer are usually not addressed.

In this work, a learning-and monocular vision-based fixed-wing interceptor is presented and demonstrated to successfully intercept MALE targets in a real environment. Shown in figure \ref{fig_Banking}, the interceptor is built entirely with off-the-shelf components. An RL controller is trained in simulation which replaces the guidance and control of a conventional autopilot. A neural network directly maps filtered visual and flight controller information to actuator commands. The performance of the RL controller is validated against baseline PN and PP controllers in simulation. Finally, the RL controller is tested on real flight interceptions of fixed-wing, MALE UAS.

\noindent The main contributions of this work are therefore:
\begin{itemize}
    \item Development of an end-to-end neural network enabling autonomous fixed-wing interception of maneuvering targets with limited onboard sensing.
    \item Development of a monocular vision- and learning-based, autonomous, fixed-wing interceptor able to intercept MALE targets in a real environment.
    \item Development of a RL framework inspired by \cite{Capelle2025DroneInterception}, incorporating fixed-wing dynamics and a real computer-vision algorithm.
\end{itemize}
The final controller is tested in real flight interceptions and assessed in terms of performance and constraints.

\section{Preliminaries}

\subsection{Problem formulation}

The interceptor and the target are flying in a North-East-Down (NED) 3-dimensional space. It is assumed that to neutralize the target, an explosive device must be detonated within a maximum radius $r_n$ around the target's position $\mathbf{p}_T$. Hence to intercept the target, the interceptor's position $\mathbf{p}_I$ must satisfy:
\begin{equation}
    (\mathbf{p}_I - \mathbf{p}_T)^\top (\mathbf{p}_I - \mathbf{p}_T) \leq r_n^2.
\end{equation}

Monocular vision suffers from Field Of View (FOV) limitations, occlusions and pixel noise. This sensor requires the problem to be framed as a partially observable Markov Decision Process (POMDP) \cite{Capelle2025DroneInterception}\cite{KAELBLING199899}, incorporating uncertainty as a quantitative observation.

\subsection{Dynamics model}
To train the RL policy in simulation, a six degree of freedom, rigid-body dynamics model for a fixed-wing aircraft was built. The model is based on literature \cite{StevensLewisAircraftControl2015}\cite{Cook1997FlightDynamics} and is described in appendix \ref{app:dynamics_model}. 

\section{Methodology}

\subsection{System identification}
An X-UAV Talon\footnote{{\url{https://www.readymaderc.com/products/details/xuav-talon-pnp}}, {Accessed: 2026‑01‑15.}} body was used for both the interceptor and the target. The interceptor was fitted with a Pixhawk 6C running Ardupilot. The inertial properties of the aircraft were estimated by modeling the fuselage as a cuboid and the wing as a flat plate with uniform density. Cross-coupling terms $I_{xy}$ and $I_{yz}$ were assumed to be zero due to the symmetry of the aircraft. $I_{xz}$ was assumed to be negligible, even though the asymmetry in the horizontal axis would cause this to be non-zero. 

A magnetic angle encoder with a wind-vane was mounted on the side of the fuselage to measure angle of attack (AOA). Three system identification flights were flown to estimate the aerodynamic coefficients of the aircraft using ordinary least squares regression \cite{KleinMorelli2006}. The regression method is described in appendix \ref{app:OLS}.

\subsection{Perception}

\label{sec: perception}
A front-facing IMX462LQR monocular camera was installed in the nose of the interceptor. The videostream was input to a proprietary target detection algorithm running on an onboard Raspberry Pi 5 fitted with a Hailo neural network accelerator. 
\subsubsection{Bounding box}
The algorithm outputs a Bounding Box (BB): a rectangular box encapsulating the target drone.
The BB outputs the following measurement vector at time step $k$:

\begin{equation}
\mathbf{z}_k^\top =
\begin{bmatrix}
x_c & y_c & w & h & c
\end{bmatrix}.
\end{equation}

\noindent Here, $x_c, y_c$ are the BB center coordinates and $w, h$ are its width and height, all in pixels. $c$ is a confidence score $\in [0,1]$ and represents how confident the algorithm is that the bounding box actually encapsulates a drone.

\subsubsection{Pinhole model}
A simulation was created that maps a 3D point in the camera frame to the 2D image plane using the pinhole model adopted from \cite{Krej__2025}. If the dimensions of the object of interest are known, the distance along the optical axis of the camera can be formulated as: 
\begin{equation}
    Z = \frac{f_xW}{w_{px}},
    \label{eq:pinhole1}
\end{equation}

\noindent where $W$ is the width of the object in meters. $w_{px}$ is the width of the object in the image in pixels. Equation \ref{eq:pinhole1} assumes that $W$ is constant when viewed from all angles, which holds only for spherical objects. For a fixed-wing target, the perceived $W$ will depend on the orientation from which it is viewed. To accommodate for this, the simulation computes the effective width as $W_{\mathrm{eff}} = W\cdot\max\bigl(\lvert\cos\psi\rvert,\; 0.6\bigr)$, where $\psi$ is the angle between the target's velocity vector and the LOS. $f_x$ is the focal length of the camera: 
\begin{equation}
    f_x = f_y = \frac{W_{\mathrm{img}}}{2\tan(\mathrm{HFOV}/2)}
    = \frac{1280 \hspace{1mm} \mathrm{px}}{2\tan(60^{\circ}/2)}
    \approx 1108\ \mathrm{px}
    \label{eq:focallength}
\end{equation}
\noindent Once $Z$ is known, $X$ and $Y$, the horizontal and vertical distance from the center of the object to the optical axis, can be formulated as follows:

\begin{equation}\label{eq:pinhole2}
	X = \frac{(u_c-c_x)}{f_x}Z, \\ \hspace{3mm} Y =  \frac{(v_c-c_y)}{f_y}Z.
\end{equation} 
\noindent Here, $u_c, v_c$ are the pixel center coordinates of the object. $c_x,c_y$ are the image center coordinates in pixels. To simulate $u_c, v_c$, the LOS vector $\mathbf{r} \in \mathbb{R}^3$ must be decomposed into $X,Y,Z$ relative to the optical axis of the camera. 

\subsubsection{Camera noise and detection probability}
A monocular camera is inherently noisy and computer vision algorithms are prone to non-detections even with a drone in the FOV. To reflect this in simulation, several error sources were introduced.

First, zero mean pixel noise was added to the BB center and its dimensions.
\begin{equation}
    \sigma_{c} = 3.0 + 0.05\,d \ \mathrm{px},
    \qquad
    \sigma_{s} = 2.0 + 0.02\,d \ \mathrm{px},
    \label{eq:pixelnoise}
\end{equation}
with $d$ the range in meters, standard deviation $\sigma_{c}$ applied to $x_c$ and $y_c$, and standard deviation $\sigma_{s}$ to $w$ and $h$.

Second, computer vision algorithms are prone to anchor the BB consistently wrong for a stretch of frames. For instance, the BB can briefly only encapsulate the fuselage of the target, missing the dimensions of the wing. Hence, a BB bias $\mathbf{b}_{k}$ was introduced:
\begin{equation}
    \mathbf{b}_{k} = \rho\,\mathbf{b}_{k-1}
        + \sqrt{1-\rho^{2}}\;\sigma_{b}\,\boldsymbol{\epsilon}_{k},
    \qquad
    \boldsymbol{\epsilon}_{k} \sim \mathcal{N}(\mathbf{0}, \mathbf{I}_{4}),
    \label{eq:bbox_bias}
\end{equation}

\noindent with $\mathbf{b}_k = [\,b_x,\ b_y,\ b_w,\ b_h\,]^{\top}$ and
$\rho = e^{-\Delta t_{\mathrm{ctrl}}/\tau_{b}}$, where $\tau_{b} = 1\,$s is the
time over which the bias persists and $\sigma_{b}$ its standard deviation. $\boldsymbol{\epsilon}_{k}$ is a vector of four independent zero mean normal samples with standard deviation of one, redrawn at each control step. The bias enters the measurement as:

\begin{equation}
\begin{aligned}
    \tilde{x}_c &= x_c + b_x\,w,  &\qquad  \tilde{y}_c &= y_c + b_y\,h, \\[3pt]
    \tilde{w}   &= (1+b_w)\,w,    &\qquad  \tilde{h}   &= (1+b_h)\,h,
\end{aligned}
\label{eq:bbox_bias_apply}
\end{equation}

\noindent so the center is displaced by a fraction of the apparent box size and
the dimensions of the box are scaled. 

Third, the per-frame detection is probabilistic and degrades with range:
\begin{equation}
    P_{\mathrm{detect}} = p_{\min} + (1-p_{\min})\,e^{-d/d_s},
\end{equation}
with $d_s = 150$ m and $p_{\min} = 0.05$. Detections can only occur if the target is within the FOV. 

Non-detections tend to be clustered. Hence, the detector is simulated as a two-state Markov chain, moving between detecting and blind with per step transition probabilities:
\begin{equation}
    P(\mathrm{blind}_{k+1} \mid \mathrm{detect}_k) = q,
    \hspace{1mm}
    P(\mathrm{detect}_{k+1} \mid \mathrm{blind}_k) = r,
    \label{eq:dropout_markov}
\end{equation}
\begin{equation}
    r = \frac{\Delta t_{\mathrm{ctrl}}}{T_{\mathrm{blind}}},
    \qquad
    q = r\,\frac{1-P_{\mathrm{detect}}}{P_{\mathrm{detect}}}.
\end{equation}
\noindent While blind, the detector recovers with probability $r$ at each control step of duration $\Delta t_{\mathrm{ctrl}}$, giving outages with a mean duration $T_{\mathrm{blind}}$. Outages therefore keep the same mean duration at all ranges while becoming more frequent as $P_{\mathrm{detect}}$ falls. 

Finally, a computer vision algorithm does not instantly produce detections, but instead lags the camera image. This detection latency was also simulated such that the current detection lags the current camera frame. The delayed detection is interpreted with the current attitude of the interceptor, so any rotation appears as a bearing error. 

BB bias, detection dropout and latency were varied through a curriculum as described in section \ref{sec:curriculum}, with specific values described in appendix \ref{app: curriculum}.

\subsection{Kalman filtering}

\label{sec: KF}

The bounding box is noisy and does not provide a continuous estimate of the relative target state. Hence, a Constant Acceleration Kalman Filter (CAKF) was applied to provide robust relative target state estimation. The same filter, with identical parameters, is used in simulation, training, and real flight.
The filter is described in appendix \ref{app:KF}.

\subsection{RL formulation}

In this study, the RL controller is a single neural network that maps the observation vector to an actuation vector $u_k \in \mathbb{R}^4$ that outputs normalized throttle $u_t \in [0, 1]$ and flight control surface $u_a, u_e, u_r \in [-1,1]$ commands. The controller maps a monocular vision-based, filtered relative target state estimate, the filter's confidence in that estimate, and the interceptor's own state directly to actuator-level outputs. Hence, the low-level FC autopilot is bypassed completely. 

\subsubsection{Observable state}
The per-frame observation vector $\mathbf{O}\in\mathbb{R}^{38}$ is built from the KF state estimate and the interceptor's own state:
\begin{equation}
\mathbf{O}^\top =
\begin{bmatrix}
\hat{\mathbf{\bm r}}_{b} &\! d &\! f &\! \mathbf r_{b} &\! \dot{\mathbf r}_{b} &\! \mathbf v_{b} &\! \mathbf q &\! \bm\omega_{b} &\! \bm \delta \\
t_{\text{lost}} &\! \log(\bm{\sigma}_p^2) &\! \log(\bm{\sigma}_v^2) &\! \bm{\delta}_{k-1} &\! V_c &\! V
\end{bmatrix}
\end{equation}
where:
\begin{itemize}
  \item $\hat{\mathbf{r}}_{b}\in\mathbb{R}^3$, $d\in\mathbb{R}$ are the unit line-of-sight vector and range estimates in the body frame,
  \item $f\in\mathbb{R}$ is a detection flag ($1$ if detected this step, else $0$),
  \item $\mathbf{r}_{b},\dot{\mathbf{r}}_{b}\in\mathbb{R}^3$ are the relative target position and velocity estimates in the body frame,
  \item $\mathbf{v}_{b}\in\mathbb{R}^3$ is the interceptor's velocity in the body frame,
  \item $\mathbf{q}\in\mathbb{R}^4$ is the interceptor's attitude quaternion,
  \item $\bm{\omega}_{b}\in\mathbb{R}^3$ are the body angular rates,
  \item $\bm{\delta}\in\mathbb{R}^4$ are the physical control-surface positions $[\delta_e,\delta_a,\delta_r,\delta_t]$ after passing through a first-order lag model as described in appendix \ref{app:actuatorlag},
  \item $t_{\text{lost}}\in\mathbb{R}$ is the time since the last successful detection,
  \item $\log\bm{\sigma}_p^2,\log\bm{\sigma}_v^2\in\mathbb{R}^3$ are the log position and velocity variances from the KF covariance,
  \item $\bm{\delta}_{k-1}\in\mathbb{R}^4$ is the previous commanded action in physical units before actuator lag,
  \item $V_c\in\mathbb{R}$ is the closing speed (positive $=$ closing),
  \item $V\in\mathbb{R}$ is the interceptor's airspeed [m/s].
\end{itemize}
To provide temporal information, $n_\text{stack}=3$ frames spaced $0.5\,$s apart (a $1\,$s window) are flattened into the network input $\mathbf{O}\times n_\text{stack}\in\mathbb{R}^{114}$. While this introduces aliasing for content with frequency higher than the 1$\,$Hz Nyquist limit, this is acceptable as the newest frame is sampled at the control rate (50$\,$Hz) and all rate quantities are observed directly instead of being inferred by differencing frames.

Although $\hat{\mathbf{r}}_{b}$ and $d$ are technically redundant with $\mathbf{r}_b$, they are deliberately both included in the observation vector. From $\hat{\mathbf{r}}_{b}$ and $d$, $\mathbf{r}_b$ can be constructed. However, this is an operation that the neural network would need to learn. $\hat{\mathbf{r}}_{b}$ is range invariant, making it easy for the policy to interpret. $\mathbf{r}_b$ preserves the true geometry, which is relevant when combining other observable states such as relative target velocity. $d$ is also included separately to inform about closing dynamics without relying on bearing. 

The observation uses the KF estimate rather than the raw bounding box. Training directly on $\mathbf{z}_k$ and its finite difference approximation of velocity was attempted, but the policy did not converge to successful interception behavior.

\subsubsection{Reward function}
The reward function $r_k$ is designed to drive the policy towards three behavioral goals: interception, perception and stable flight, and can be formulated as:
\begin{equation}
\begin{aligned}
    r_k = f_{\text{norm}} \bigg(\;
        & c_d\, r_{\text{dist}} + \mathbf{1}_{\text{int}}\, c_{\text{int}}\,(t_{\max}-t_k)
        && \text{(interception)}\\[3pt]
        & {}+ c_c\, r_{\text{cam}} - c_\Phi\, p_\Phi
        && \text{(perception)}\\[3pt]
        & {}- c_\Omega\, p_\Omega - c_u\, p_u - c_\alpha\, p_\alpha \;\bigg)
        && \text{(stable flight)}
\end{aligned}
\end{equation}
\noindent where $\mathbf{1}_{\text{int}}$ activates on the intercept step and
$f_{\text{norm}} = R_{\max}/t_{\max}$ normalizes the return across episode lengths and control rates. Here, $R_{\max}=1000$ is the maximum achievable return for a single episode and $t_{\max}$ is the time limit upon which timeout triggers.
The shaping terms are:
\begin{equation}
    r_{\text{dist}} = \tanh\!\left(\frac{d_{k-1}-d_k}{v_{cl,\text{safe}}\,\Delta t_{\text{ctrl}}}\right),
    \qquad d_k = \lVert \mathbf{p}_T - \mathbf{p}_I \rVert,
\end{equation}
\begin{equation}
    p_\Omega=\sum_{i=1}^{3}\Big(\min\big(\tfrac{|\Omega_i|}{\Omega_{r_{\max}}},1\big)\Big)^2,
    \qquad
    p_u = \tfrac{1}{16}\sum_{i=1}^{4}(\Delta u_i)^2,
\end{equation}
\begin{equation}
    r_{\text{cam}} = c\,\max\!\Big(0,\; 1 - 4\big[(c_x-\tfrac{1}{2})^2 + (c_y-\tfrac{1}{2})^2\big]\Big),
\end{equation}
\begin{equation}
    p_\Phi = \frac{\min(\tau_{\text{lost}},\,\tau_{\max})}{\tau_{\max}},
\end{equation}
\begin{equation}
    p_\alpha = \frac{e^{K\,\bar\alpha}-1}{e^{K}-1},
\end{equation}
\begin{equation}
    \bar\alpha = \min\!\Big(\max\big(\tfrac{|\alpha|-\alpha_{\text{warn}}}{\alpha_{\max}-\alpha_{\text{warn}}},0\big),1\Big),
    \quad K=-3.
\end{equation}

\noindent The distance-progress term $r_{\text{dist}}$ rewards a timestep reduction in true relative distance $d_k = \lVert \mathbf{p}_T - \mathbf{p}_I\rVert$, normalized by the distance the interceptor would cover in one control step at a nominal safe closing speed $v_{cl,\text{safe}}$. The $\tanh$ saturates this signal, so closing faster than $v_{cl,\text{safe}}$ yields no additional reward, which discourages aggressive high-speed dives and instead promotes steady, controlled closure. The interception bonus $c_{\text{int}}(t_{\max}-t_k)$ is granted only on the step the target is intercepted ($\mathbf{1}_{\text{int}}=1$). Its magnitude decreases with the interception timestep $t_k$, rewarding earlier interceptions, while the $(t_{\max}-t_k)$ form removes any incentive to loiter near the target and accumulate shaping reward.

The perception terms reward behavior that keeps the target inside camera view. The centering reward $r_{\text{cam}}$ is maximal when the target is imaged at the frame centre ($c_x=c_y=\tfrac{1}{2}$) and decays parabolically to zero at the frame edge. It is gated by the detection confidence $c$, vanishing when no detection is
available. It provides a gradient toward keeping the target in view and hence stimulates pure pursuit-like tailchase interception behavior. The FOV penalty $p_\Phi$ is its reactive counterpart: $\tau_{\text{lost}}$ counts the control steps since the last successful detection, and the penalty grows linearly to unity at the loss horizon $\tau_{\max}$, beyond which the episode is terminated. Together they encourage the agent to keep the target framed before it is lost, and to reacquire it quickly if it is.

The stable flight terms aim to shape the policy towards behavior that is realistic and non-damaging for the airframe and also not stall the aircraft. The body-rate penalty $p_\Omega$ discourages aggressive angular rates $\Omega_i$ (roll, pitch, yaw), each normalized by $\Omega_{r_{\max}}$ and squared. The smoothness penalty
$p_u$ penalizes large timestep changes in the four actuator commands $\Delta u_i$, pushing for smooth control. The exponential AOA penalty $p_\alpha$ enforces high AOA avoidance, remaining zero below a warning angle $\alpha_{\text{warn}}$ and rising sharply to unity at the AOA limit $\alpha_{\max}$.

The coefficients are calibrated relative to the unit maximum of the progress reward. Penalties weighted above one: $p_\Phi$, $p_\alpha$, and $p_\Omega$ near saturation, can render sustained target loss, incipient stall, or hard manoeuvring net-negative even while the interceptor is closing, whereas $c_c<1$ ensures the centering reward nudges rather than overrides the pursuit geometry. Appendix \ref{app: reward function} describes the exact weights used in training of the policy used in the real flight tests.

\subsection{RL algorithm}
Proximal Policy Optimization (PPO) \cite{schulman2017proximal} was used for training in combination with the Stable-Baselines3 Python library \cite{Raffin2021StableBaselines3}. The algorithm parameters were all kept default, with the exception of clip range $\epsilon$ and discount factor $\gamma$. $\epsilon$ was reduced from 0.2 to 0.05, meaning the policy updates are limited to 5 percent increments per step to increase training stability. $\gamma$ was increased from 0.99 to 0.999 to encourage longer-term path planning in the policy.

\subsection{Training environment}
In \cite{Capelle2025DroneInterception}, a custom environment was developed to run a parallel simulation of more than a thousand quadcopters. In this work, the environment is adapted for fixed-wing aircraft. JAX \cite{bradbury2018jax} and Numba \cite{lam2015numba} are used for efficient compiling, reducing training time and allowing the training to run on a CPU. 
Training occurs in episodes with a single target $T$ and interceptor $I$. $I$ is initialized with conditions close to trim. $T$ is modeled as a moving or accelerating point mass. Episodes are terminated when:
\begin{itemize}
    \item $T$ is intercepted : $(\mathbf{p}_I - \mathbf{p}_T)^\top (\mathbf{p}_I - \mathbf{p}_T) \leq r_n^2$
    \item $T$ is missed : $r_n<d<r_m \hspace{2mm}\& \hspace{2mm}\dot{d}>0$
    \item $T$ is lost : $\tau_{\text{lost}} > \tau_{\max}$
    \item $I$ stalls: $\alpha > 60^\circ$ 
    \item $I$ hits the ground: $h<10$ m  
    \item $I$ goes above altitude ceiling: $h>1000$ m 
    \item $I$ goes out of bounds $p_{I_x},p_{I_y} > 1000$ m
    \item timeout occurs: $t_k > t_{\max}$
\end{itemize}
with $r_n = 3\,$m and miss radius $r_m$ = 5$\,$m. $\tau_{\text{lost}}$ is the number of control steps since the last successful detection and $\tau_{\max} = 200$ steps (4$\,$s at 50$\,$Hz). $t_{\max}$ varies across the curriculum (see appendix \ref{app: curriculum}) to accommodate for longer range approaches and faster targets. 

Contrary to quadrotors, fixed-wing drones can stall at low airspeeds or high AOA. This proved to be one of the main challenges encountered in this work. Instead of exploring post-stall modelling, the approach taken here involved encouraging the agent to avoid stall regimes in the first place. While stall regimes were not clearly defined during the system identification, 25 degree AOA was deemed to be a conservative limit to avoid, regardless of airspeed, as this was near the maximum AOA observed during the system identification flights. At first, episodes were simply cut off when this threshold was crossed, however this resulted in many episodes being lost due to stalls without the agent learning how to avoid them. It is also not possible to penalize AOA, as non-zero AOA is needed to generate lift. Finally, a stall penalty was included that only penalized time steps above a certain AOA threshold. 

To create a RL controller that works in a real environment, the simulation should reflect uncertainties, noise, constraints and variance in the underlying observations and model. Domain Randomization (DR) has been successfully used to train a single neural network that is capable of controlling physically distinct quadcopters \cite{Ferede2025OneNet}. In this work, DR is increased from zero to 30 percent around nominal values as the training progresses.

\subsection{Curriculum learning}
\label{sec:curriculum}
The full interception task involves pursuing a far-away maneuvering target from noisy monocular vision estimates under wind, turbulence and domain randomization of the aircraft parameters. This is too difficult to learn from random initialization all at once. Hence, the training is structured into a 10-stage curriculum. In the initial stage, the interception task is relatively easy. The target is close-by, slow moving and not accelerating. The main goal of this stage is to learn how to fly the aircraft and produce motion towards the target. Progression to a successive stage occurs when the interception rate reaches 75 percent. The policy is evaluated every 1000 episodes, to see whether it can advance to the next stage in the curriculum. The final stage uses an interception target of 101 percent, which is impossible, making it train until the timestep budget runs or the training crashes. When progressing between stages, the best policy is transferred (not the latest) and acts a starting point for the training to continue. Appendix \ref{app: curriculum} describes the individual stages in the curriculum used for training the policy that was ultimately tested in real flight.

\subsection{Real interception test flights}
Flight tests were conducted at Vliegbasis Valkenburg in the Netherlands. It was quickly discovered that testing the entire system together proved to be difficult initially. Target perception was especially challenging. The interceptor needs to be aligned with the target before the RL policy can take control of the actuators. Furthermore, the detection algorithm was not as robust as anticipated, and particularly struggled with tail-based engagements when the background is not clear and the target has a very thin profile. 

\subsubsection{Virtual stationary target}
To separate perception from control, a script was developed that projects a virtual 1.7 meter wide target on the camera image using the pinhole model. This script takes predefined Global Navigation Satellite System (GNSS) coordinates and the interceptor's attitude and position and computes a bounding box onto the camera image. This artificial bounding box then feeds into the CAKF. This way, the pipeline is identical to the deployment scenario, except there are no detection dropouts and the aligning the interceptor with the target becomes much easier. 

\subsubsection{Virtual moving target}
Once stationary virtual targets were successfully intercepted, the script was expanded to also allow a moving virtual target to be projected onto the camera. Instead of a single GNSS coordinate, a series of predefined coordinates plus altitudes and a fixed ground speed simulate a moving target, with the bounding box generation logic identical to the stationary virtual target. The moving target travels between the given coordinates in a looping fashion, progressing to the first coordinate after the last coordinate has been reached, and does so indefinitely, allowing for systematic testing. 

\subsubsection{Real fixed-wing target}
A second X-UAV Talon was developed to function as a moving target. To provide a meaningful interception scenario, the target was autonomously flying a 90 meter altitude rectangular waypoint trajectory using AUTO mode. The pilot was instructed to provide evasive stick inputs whilst in AUTO mode, which allows the target to maneuvre away from the flight mission until the pilot lets go of the sticks. This makes it both harder to track and to intercept the target. Airspeed was managed autonomously with throttle by Ardupilot; the pilot could alter demanded airspeed with stick inputs.  

\subsection{Baseline classical controller}

To compare the RL with a baseline, vision-based Proportional Navigation (PN) and Pure Pursuit (PP) classical guidance controllers were developed. All three controllers (RL, PN and PP) used the same perception pipeline as described in sections \ref{sec: perception} and \ref{sec: KF}, meaning differences in performance would be attributable to differences in control.   

\subsubsection{Guidance laws}

PN commands an acceleration perpendicular to the LOS and proportional to its rotation rate:

\begin{equation}
    \mathbf{a}_{\mathrm{PN}} = N\,V\,\bigl(\boldsymbol{\Omega}\times\hat{\mathbf{r}}_b\bigr),
    \qquad
    \boldsymbol{\Omega} = \mathbf{R}_{b/i}
    \bigl(\hat{\mathbf{r}}_i\times \dot{\hat{\mathbf{r}}}_i\bigr)
\end{equation}
\begin{equation}
    \hat{\mathbf{r}}_i = \mathbf{R}_{i/b}\,\hat{\mathbf{r}}_b,
    \qquad
    \hat{\mathbf{r}}_b = \frac{\mathbf{r}_b}{\lVert\mathbf{r}_b\rVert},
\end{equation}

\noindent where $N$ is the dimensionless gain and $\boldsymbol{\Omega}$ is the inertial LOS rotation rate. $\dot{\hat{\mathbf{r}}}_i$ is the finite difference of $\hat{\mathbf{r}}_i$ over one control step, passed through a first order low-pass filter with a time constant $\tau_{\lambda} = 0.5\,$s. $\boldsymbol{\Omega}$ is computed in the inertial frame and then converted to the body frame using rotation matrix $\mathbf{R}_{b/i}$ such that the interceptor's own rotation does not influence it.
The LOS rate can also be derived analytically in its standard vector form \cite{yanushevsky2018modern}\cite{zarchan2012tactical} as $\boldsymbol{\Omega} = \frac{\mathbf{r}_b\times\dot{\mathbf{r}}_b}{\lVert\mathbf{r}_b\rVert^{2}}$ scaled by $V_c = -\,\hat{\mathbf{r}}_b\cdot\dot{\mathbf{r}}_b$ instead of $V$, however this depends on the estimated relative velocity $\dot{\mathbf{r}}_b$ to set the direction and magnitude of the command. With the current approach, the commands generated depend only on the measured bearing and the interceptor's own speed, and are not affected by the noisy estimate of relative velocity.

PP instead drives the angle $\varepsilon$ between the camera boresight
$\hat{\mathbf{b}}=[1,0,0]^{\top}$ and the line of sight to zero:
\begin{equation}
    \mathbf{a}_{\mathrm{PP}} = K\,V\,\varepsilon\,\frac{\mathbf{e}_{\perp}}{\lVert\mathbf{e}_{\perp}\rVert}, \hspace{1 mm}
    \varepsilon = \arccos\bigl(\hat{\mathbf{r}}_b\cdot\hat{\mathbf{b}}\bigr),\hspace{1 mm}
    \mathbf{e}_{\perp} = \hat{\mathbf{r}}_b - (\hat{\mathbf{r}}_b\cdot\hat{\mathbf{b}})\hat{\mathbf{b}}.
\end{equation}

so that $K$ $[\mathrm{s^{-1}}]$ sets the demanded turn rate per unit of pointing error. LOS should be minimized with respect to the boresight instead of the velocity vector because a fixed camera should be aligned with the target to see. 

\subsubsection{Cascaded autopilot}

Both classical guidance laws produce an acceleration command, which must be realized by an aircraft whose only maneuvering force is lift generated from the wings. To achieve the normal acceleration, the aircraft must hence roll the lift vector onto the demanded acceleration vector. A cascaded autopilot performs this mapping. The acceleration command is first translated into the specific force the accelerometer must register,
\begin{equation}
    \mathbf{f}_{\mathrm{cmd}} = \mathbf{a}_{\mathrm{cmd}} - \mathbf{g}_{b},
\end{equation}
with $\mathbf{g}_{b}$ gravity expressed in body axes. Its magnitude in the plane normal to the optical axis sets how much lift is required, and its direction sets where that lift must point:
\begin{equation} 
    f_{n,\mathrm{cmd}} = \bigl\lVert (f_{y},f_{z}) \bigr\rVert,
    \hspace{1mm}
    \phi_{\mathrm{err}} = \operatorname{atan2}\bigl(f_{y},\,-f_{z}\bigr),
\end{equation}
\noindent with $f_y$ and $f_z$ the $y$ and $z$ components of $\mathbf f_{\mathrm{cmd}}$ in the body frame.
The roll channel drives $\phi_{\mathrm{err}}$ to zero, aligning the lift vector with the demand, while the pitch channel regulates the achieved normal specific force onto $f_{n,\mathrm{cmd}}$ through a feedback loop. Both outer loops feed inner body-rate loops with proportional, integral and feedforward terms. A sideslip loop drives $\beta$ to zero through the rudder.

\begin{figure}[!t]
\centering
\includegraphics[width=1\columnwidth]{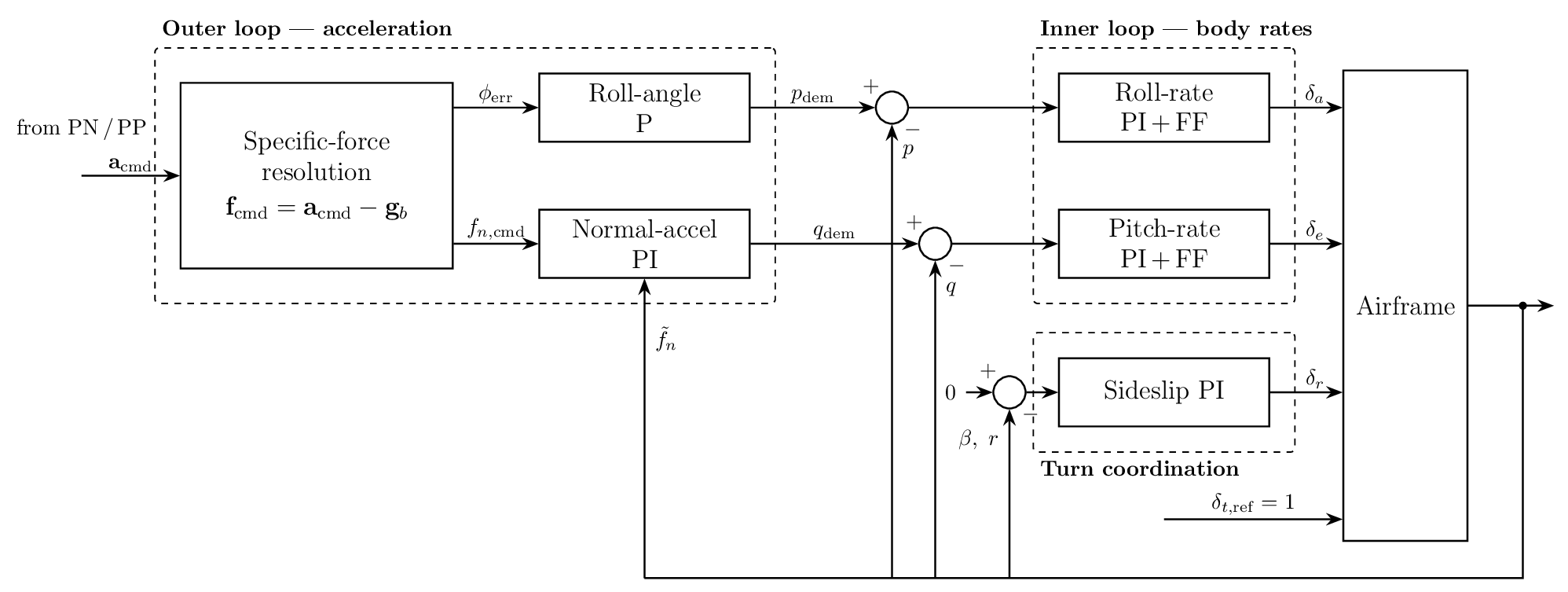}
\caption{A block diagram showing the constructed feedback loops that power the baseline classical autopilot. In simulation, sideslip is computed as $\beta = \arcsin\!\left(v/{\lVert\mathbf{v}_{b}\rVert}\right)$, with $v$ the lateral velocity in the body frame. In real flight, Ardupilot's EKF estimate for sideslip can be used. Note that unlike the baseline controllers, the RL policy does not observe $\beta$.}
\label{fig_autopilot_blockdiagram}
\end{figure}

Because $\phi_{\mathrm{err}}$ is an error rather than an absolute bank angle, and because gravity is accounted for once in $\mathbf{f}_{\mathrm{cmd}}$ rather than through separate corrections in each channel, the mapping remains valid at any attitude. The pitch loop compares the demanded normal specific force $f_{n,\mathrm{cmd}}$ with the actual specific force, and zeroes the difference with a pitch rate command. In simulation, measured specific force is not available and hence it is reconstructed from the velocity in the body frame:
\begin{equation}
    \tilde{\mathbf{f}} = \frac{\mathbf{v}_{b}[k]-\mathbf{v}_{b}[k-1]}{\Delta t}
                 + \boldsymbol{\omega}\times\bar{\mathbf{v}}_{b}
                 - \mathbf{g}_{b}.
\end{equation}
Because the leading term is a numerical derivative, it amplifies noise on $\mathbf{v}_{b}$ by $1/\Delta t$. Therefore, the estimate is low pass filtered by a first order filter with a time constant $\tau_{f}=0.08\,$s. A block diagram of the constructed autopilot is shown in figure \ref{fig_autopilot_blockdiagram}.

A total of sixteen gains were tuned manually, with the inner loops being closed first before the outer loops were tuned. The gains are summarized in appendix \ref{app:autopilot_gains}.

After the cascaded autopilot was tuned, the guidance law gains were then selected by sweeping each against interception rate, giving $N=2$ and $K=1\,\mathrm{s}^{-1}$.

\section{Results}

\subsection{System identification}
The interceptor weighs approximately 2.87$\,$kg. The moments of inertia are estimated to be: $I_{xx}= 0.14$$\,$kg$\cdot$m$^2$, $I_{yy}=0.20$$\,$kg$\cdot$m$^2$ and $I_{zz}=0.34$$\,$kg$\cdot$m$^2$. All cross-coupling terms are assumed to be zero. The surface area of the wing is 0.46$\,$m$^2$, the wingspan is 1.718$\,$m and the mean aerodynamic chord is 0.265$\,$m. The dihedral angle of the V-tail was measured to be approximately 35 degrees. A 12.5 inch diameter, 6.5 inch pitch propeller was used in combination with an X2820 KV570 electric motor. For the thrust model, data from a test report\footnote{\url{http://en.rcsunnysky.com/x-fixedwingseries/1104.html}} was used. The estimated maximum thrust at zero airspeed is 38.23$\,$N. The maximum velocity was estimated to be around 40$\,$m/s based on airspeeds achieved in flight. 

\begin{table}[htbp]
\centering
\caption{Estimated Aerodynamic Stability and Control Derivatives}
\label{tab:aero_derivatives}
\renewcommand{\arraystretch}{1.15}
\begin{tabular}{lcccc}
\hline
\textbf{Category} & \textbf{Coefficient} & \textbf{Flight 1} & \textbf{Flight 2} & \textbf{Flight 3} \\
\hline
\multirow{5}{*}{Pitching moment}
    & $\mathbf{R}^2$        &  0.4794 &  0.6809 &  0.6889 \\
    & $C_{m0}$              & -0.0035 & -0.0113 & -0.0127 \\
    & $C_{m_\alpha}$        & -0.0508 & -0.1885 & -0.1724 \\
    & $C_{m_q}$             & -1.9179 & -3.0066 & -2.4160 \\
    & $C_{m_{\delta_e}}$    &  0.1536 &  0.3647 &  0.2907 \\
\hline
\multirow{4}{*}{Lift}
& $\mathbf{R}^2$        &  0.5943 &  0.7853 &  0.7967 \\
& $C_{L0}$              &  0.2140 &  0.3793 &  0.2963 \\
& $C_{L_\alpha}$        &  1.3442 &  3.1066 &  2.8529 \\
& $C_{L_{\delta_e}}$    &  1.6628 &  0.4925 &  0.4181 \\
\hline
\multirow{3}{*}{Drag}
& $\mathbf{R}^2$        &  0.4316 &  0.5727 &  0.4473 \\
& $C_{D0}$              &  0.0797 &  0.1552 &  0.1232 \\
& $k_{\mathrm{induced}}$&  0.3142 &  0.1835 &  0.1303 \\
\hline
\multirow{6}{*}{Rolling moment}
&$\mathbf{R}^2$        &  0.2281 &  0.2754 &  0.2486 \\
&$C_{l0}$              & -0.0005 & -0.0010 & -0.0007 \\
&$C_{l_\beta}$         & -0.0074 & -0.0176 & -0.0180 \\
&$C_{l_p}$             & -0.0781 & -0.0996 & -0.0910 \\
&$C_{l_r}$             &  0.0156 &  0.1029 &  0.0727 \\
&$C_{l_{\delta_a}}$    &  0.0188 &  0.0367 &  0.0317 \\
\hline
\multirow{6}{*}{Yawing moment}
&$\mathbf{R}^2$        &  0.3204 &  0.3565 &  0.4278 \\
&$C_{n_\beta}$         &  0.0132 &  0.0215 &  0.0206 \\
&$C_{n_p}$             & -0.0198 & -0.0797 & -0.0524 \\
&$C_{n_r}$             & -0.0062 & -0.0509 & -0.0342 \\
&$C_{n_{\delta_a}}$    &  0.0010 &  0.0108 &  0.0059 \\
&$C_{n_{\delta_r}}$    &  0.0242 &  0.0316 &  0.0263 \\
\hline
\multirow{3}{*}{Side force}
&$\mathbf{R}^2$        &  0.1243 &  0.0313 &  0.1075 \\
&$C_{Y_\beta}$         & -0.0954 & -0.0577 & -0.1361 \\
&$C_{Y_{\delta_r}}$    & -0.1689 &  0.1276 & -0.2167 \\
\hline
\end{tabular}
\vspace{3mm}
\footnotesize
\newline \noindent The $\mathbf{R}^2$ values represent the Pearson correlation coefficient for each respective axis, where larger values indicate a better agreement between the measured data and the fitted model.
\end{table}

In table \ref{tab:aero_derivatives}, the estimated aerodynamic coefficients of the aircraft are given for the three system identification flights that were performed. Flight 1 utilized Ardupilot's extended Kalman filter's estimate of the AOA, whereas during flight 2 and 3, a wind vane on an angle encoder was used to measure the real AOA. Additionally, flights 2 and 3 were flown in MANUAL mode, meaning there is no stabilization and each axis was excited separately. Flight 1 was mostly flown in STABILIZED mode, which means yaw and roll are excited simultaneously on lateral stick inputs.

\begin{figure}[!t]
\centering
\includegraphics[width=1\columnwidth]{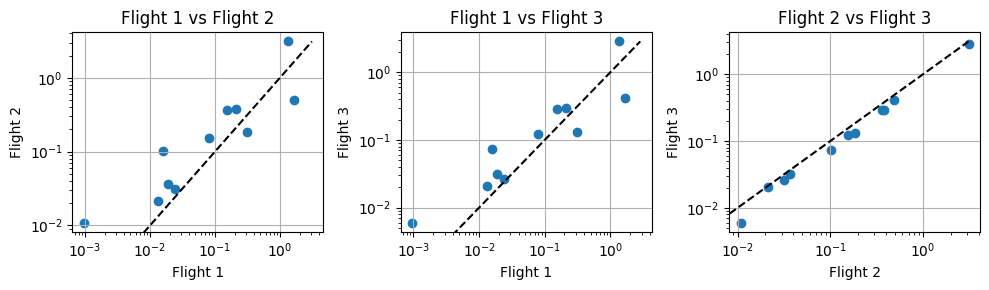}
\caption{A flight by flight comparison of the estimated aerodynamic coefficients. The dashed line indicates identical estimates between two flights. }
\label{fig_Agreement}
\end{figure}

Figure \ref{fig_Agreement} shows the aerodynamic coefficients estimates as a flight-by-flight comparison. Especially flight 2 and 3 agree well. This is likely because of the use of the AOA vane, instead of the Extended Kalman Filter (EKF) estimated AOA onboard the flight controller.

\begin{figure}[!t]
\centering
\includegraphics[width=1\columnwidth]{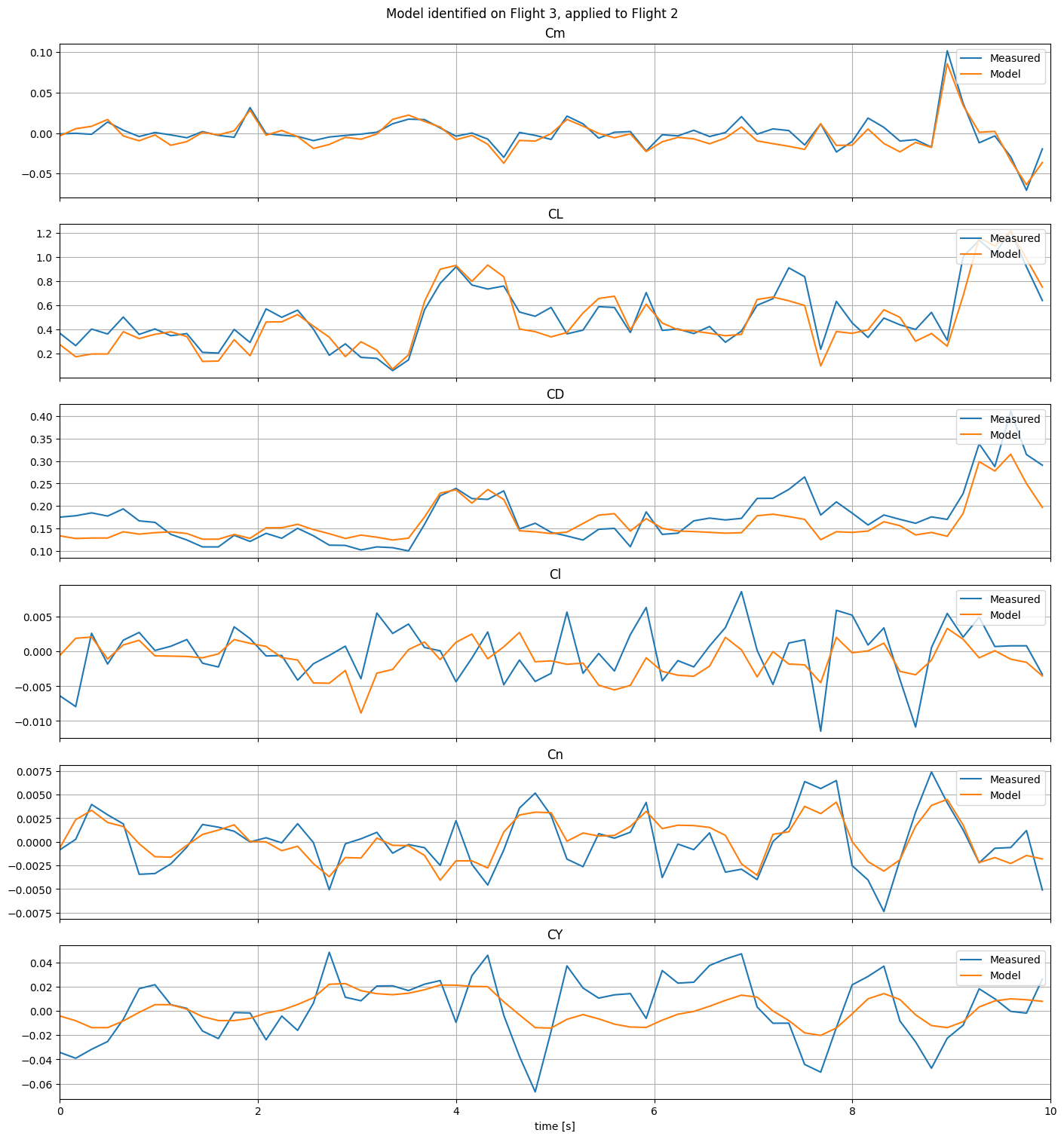}
\caption{A ten second segment from flight 2 showcasing how the model identified using flight 3 (orange line) compares to the measured aerodynamic coefficients from flight 2 (blue line) over time.}
\label{fig_10secondSegment}
\end{figure}

Figure \ref{fig_10secondSegment} shows the modeled versus measured aerodynamic coefficients over a ten second segment of flight 2. The model was identified from flight 3, so the comparison is out of sample. While there are sizeable residuals, the model follows the trend of the measurements.

\begin{figure}[!t]
\centering
\includegraphics[width=1\columnwidth]{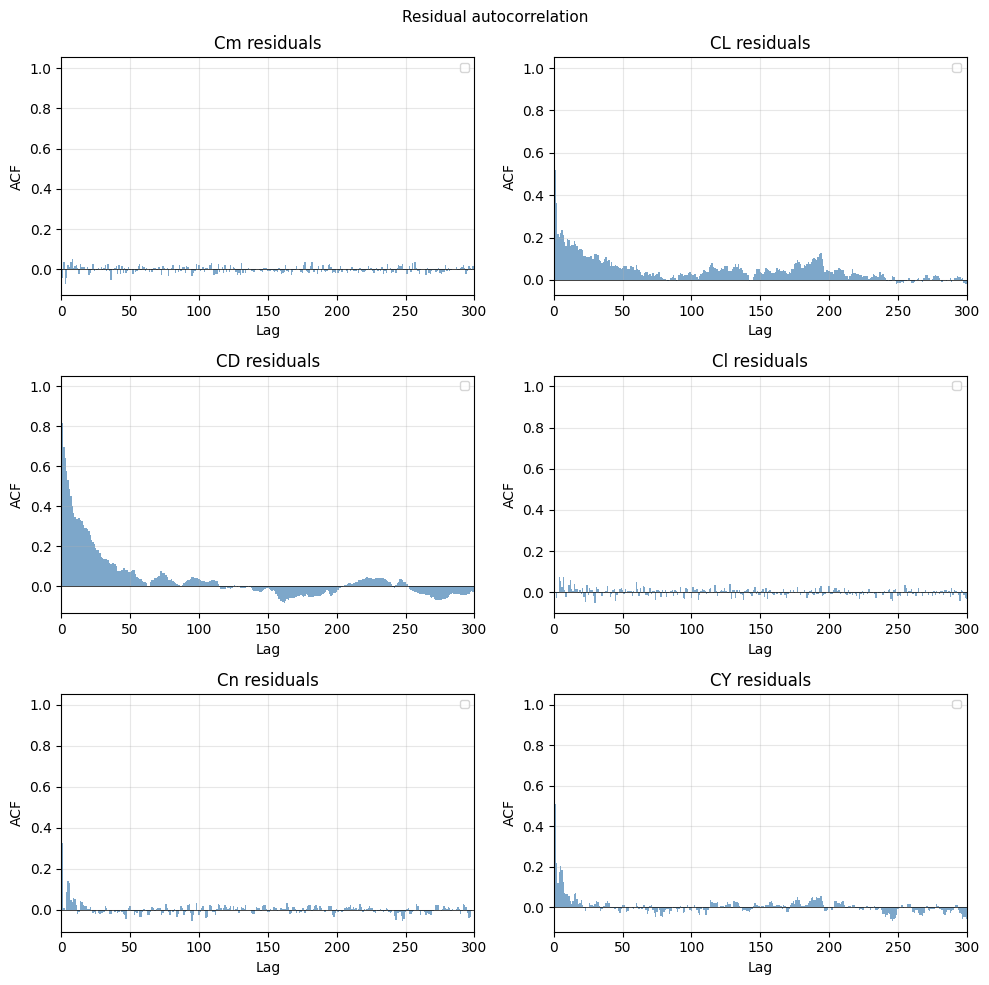}
\caption{The autocorrelation function of the residuals for each aerodynamic coefficient.}
\label{fig_Autocorrelation}
\end{figure}

OLS regression assumes the residuals are white noise. To check whether this assumption holds, the autocorrelation function (ACF) for the residuals was computed for lag 0 to 300 and is shown in figure \ref{fig_Autocorrelation}. For most axes, the residuals are indeed uncorrelated over time. However, $C_L$, and consequently $C_D$, which is dependent on $C_L$, shows significant autocorrelation, especially at lower lag. This means the residuals are potentially not white noise, indicating some deficiency in the aerodynamic model. This was likely due to the thrust model, which neglects advance ratio and other physically correct additions in favor of simplicity. Instead of perfecting the system identification, these results were deemed "good enough" to start training a controller as domain randomization of the aircraft parameters alleviates the need for really accurate system identification. 

For each coefficient, the estimate with the strongest fit and best agreement between flights was adopted. The final values used in training the RL policy are described in appendix \ref{app:sysid}. 

\subsection{Cascaded autopilot for baseline classical guidance}

\label{sec: classical controller in sim}

\begin{figure}[!t]
\centering
\includegraphics[width=1.03\columnwidth]{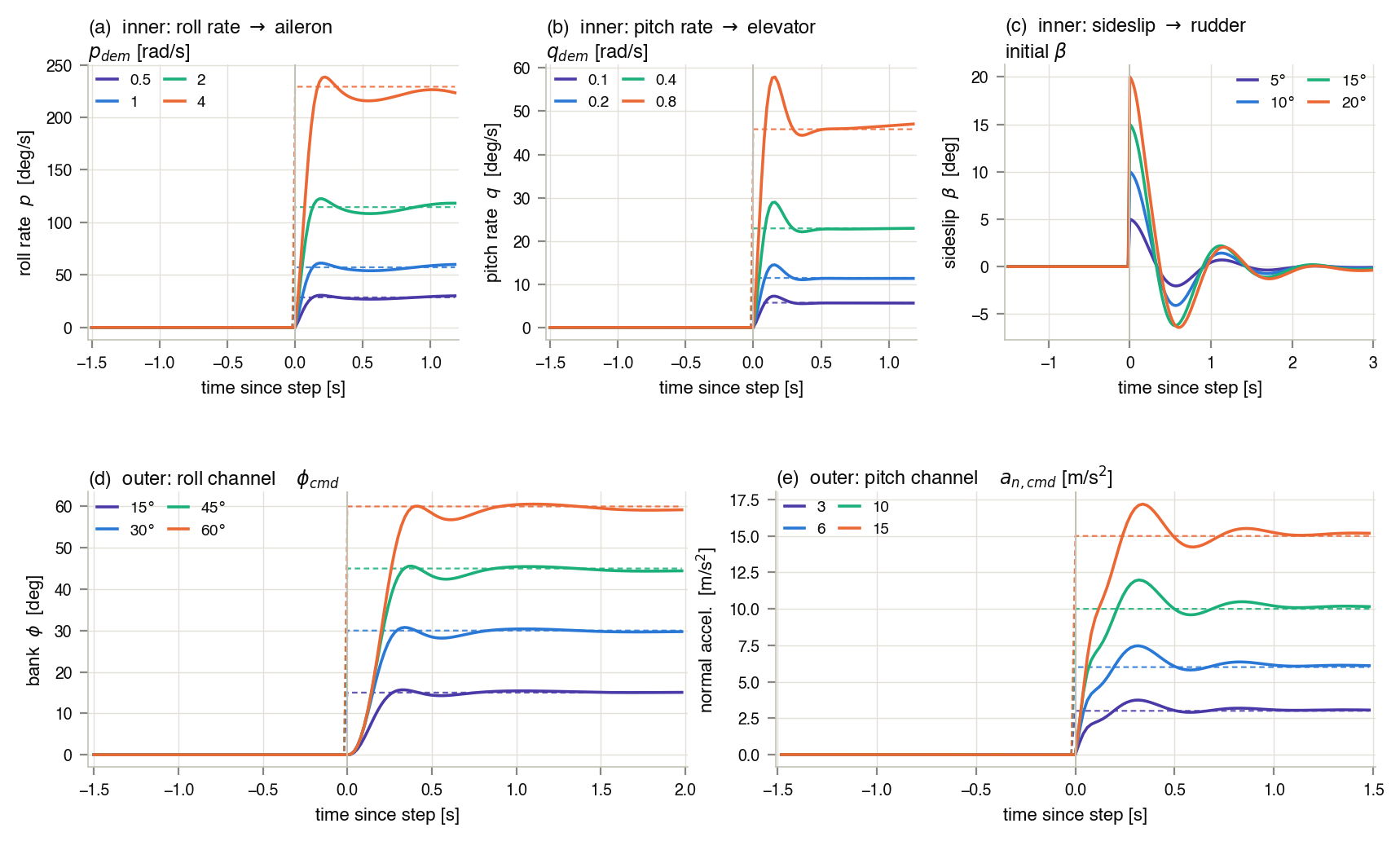}
\caption{The performance of the cascaded autopilot controller in the full non-linear model. (a) The inner roll-rate loop tracking response to a 0.5, 1, 2 and 4$\,$rad/s roll rate demand. (b) The inner pitch-rate loop tracking response to a 0.1, 0.2, 0.4 and 0.8$\,$rad/s pitch rate demand. (c) The inner sideslip angle loop nullifying response to a 5, 10, 15 and 20 degree initial sideslip. (d) The outer roll angle tracking loop response to a 15, 30, 45 and 60 degree roll demand. (e) The outer normal acceleration tracking loop response to a 3, 6, 10 and 15$\,$m/s$^2$ normal acceleration demand.}
\label{fig_autopilot_loops}
\end{figure}

Figure \ref{fig_autopilot_loops} shows the simulated tracking performance of the autopilot used for the baseline classical controller in the full non-linear model. The two inner rate loops are the fastest elements of the cascade. The roll rate loop (a) reaches 90 percent of demand in 0.10$\,$s with at most 7 percent overshoot. The pitch-rate loop (b) is even faster at 0.06 s but more oscillatory, overshooting by 26 percent before settling within 2 percent in roughly 0.44$\,$s. The sideslip regulator (c) is excited by an instantaneous lateral gust rather than a demand step, and returns sideslip angle to zero from initial values of 5$^\circ$ to 20$^\circ$. The first undershoot reaches approximately $-6^\circ$ for the largest lateral gust, with residual oscillations fully decayed by 2.5$\,$s.

The two outer loops are three to five times slower than the inner loops they command. The roll channel (d) reaches 90 percent of a commanded bank angle in 0.16-0.22$\,$s and settles within 2 percent in approximately 0.75$\,$s, with nearly no overshoot. The normal-acceleration loop (e) is the slowest and least damped element: it rises in 0.16-0.20$\,$s but overshoots by 15-26 percent and takes 1.0-1.5$\,$s to settle. 

\subsection{RL learning}
Throughout the course of this research more than 120 different RL policies were trained with varying degrees of success in simulation. The scope of this research was sim-to-real transfer. Hence, only the policy that successfully intercepted targets in real flight is discussed.

\begin{figure}[!t]
\centering
\includegraphics[width=1.05\columnwidth]{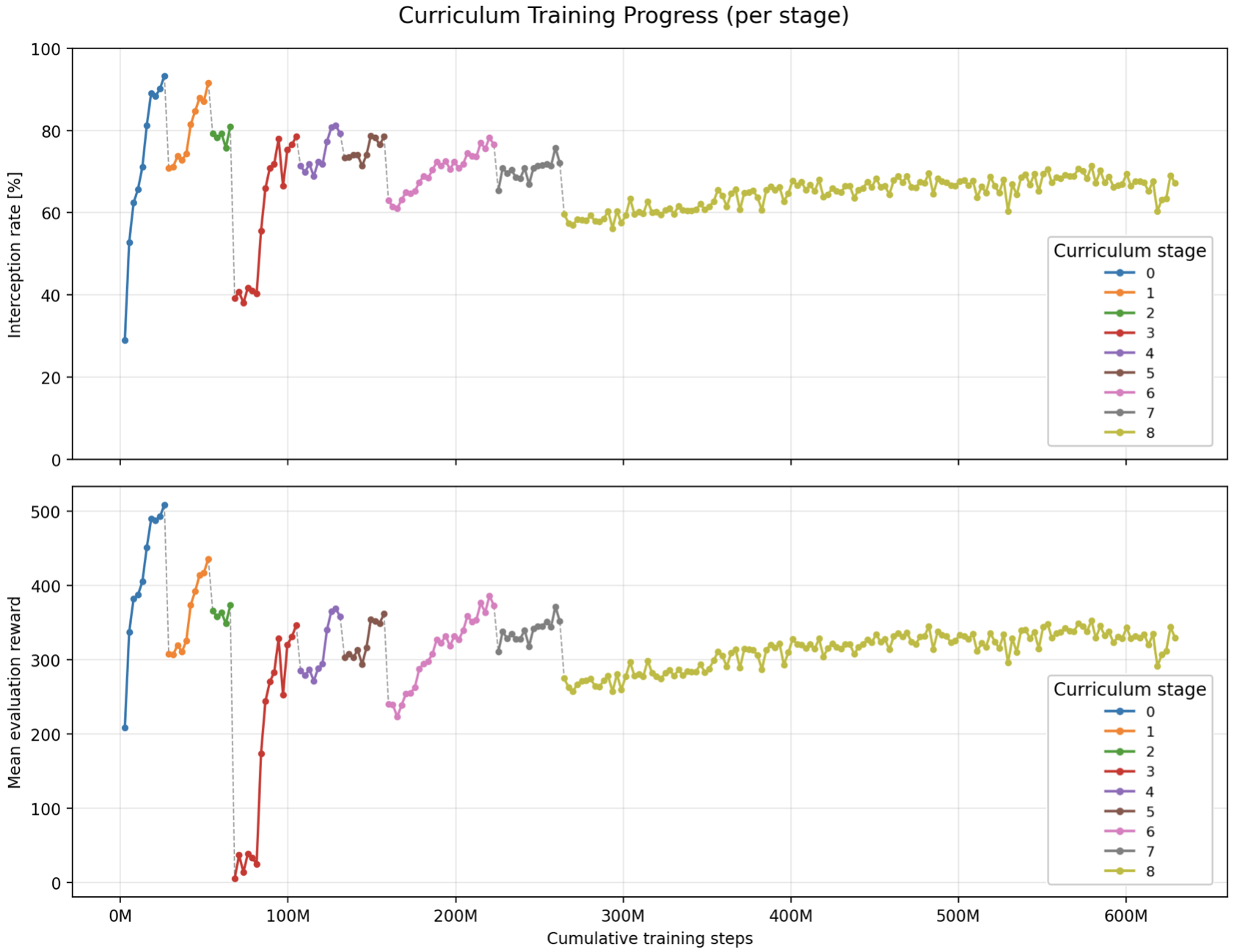}
\caption{The top and bottom figures showcase the progression of the average interception rate and reward throughout the curriculum respectively. Data is plotted in increments of a thousand episodes over the cumulative amount of training steps. Each colour represents a separate stage.}
\label{fig_curriculumprogress}
\end{figure}

Learning was structured into 10 stages (see appendix \ref{app: curriculum}). The progress of the training in terms of average interception rate and mean reward is shown in figure \ref{fig_curriculumprogress}. In the initial stage, the agent quickly progresses from an approximate 30 percent interception rate to nearly 95 percent interception rate. At this stage, the task is simple and maintaining the aircraft's attitude usually results in an interception. As the training progresses through the curriculum, the agent learns to adapt to new circumstances, which causes it to recover from initial drops in performance. An interception rate above 80 percent is not seen after completing stage 4. In stage 8, the task is so difficult that the agent is not able to progress above the threshold of 75 percent interception rate before the training stops (due to CPU constraints).

\subsection{Performance in simulation}

\subsubsection{Single episode analysis}
\label{sec: single episode analysis}
\begin{figure}[!t]
\centering
\includegraphics[width=1\columnwidth]{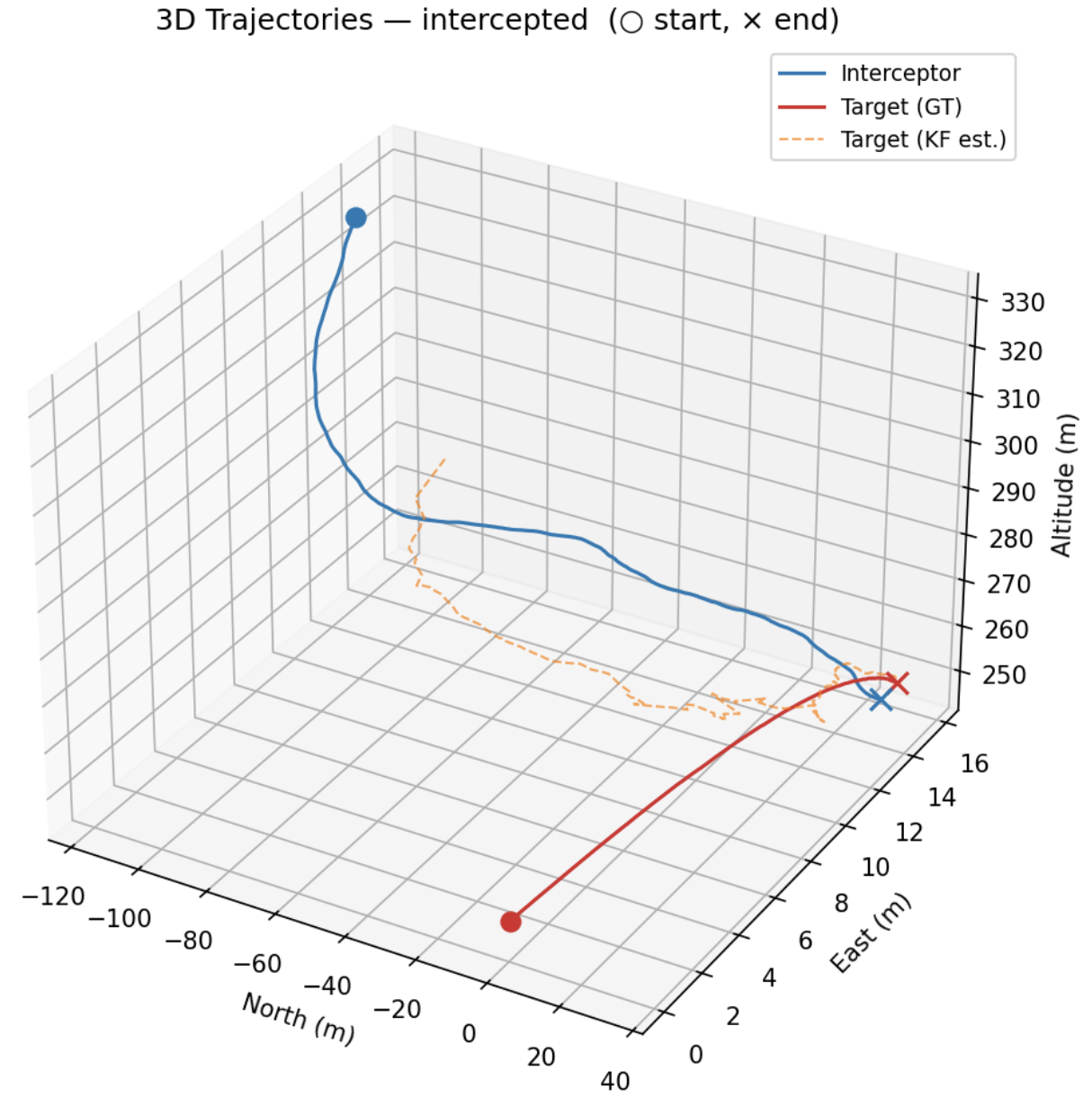}
\caption{The trajectories of the interceptor (blue) and target (red) throughout the course of a single simulated engagement. The circle and cross indicate the start and finish respectively. The dashed orange line indicates the KF estimate of the target's position. }
\label{fig_trajectory}
\end{figure}

Before analyzing performance across many episodes, it is insightful to examine a single episode to understand how the policy behaves. In figure \ref{fig_trajectory}, the trajectories of the interceptor and target in a single engagement are shown. The target inertial velocity in NED was set to 10$\,$m/s ($v_N = 4$$\,$m/s, $v_E = 6$$\,$m/s) and it was manoeuvering ($a_N = $ 1$\,$m/s$^2$, $a_E = $ -1$\,$m/s$^2$). During this episode, there was a steady wind speed of 2$\,$m/s and a 0.5$\,$m/s Gaussian gust at each timestep. The aircraft parameters were randomized ±15 percent around nominal values. Target wingspan was also randomized ±10 percent around 1.7$\,$m. The detection latency was set at 0.06 seconds. The interceptor was able to successfully intercept the target within 6 seconds under these circumstances. 

\begin{figure}[!t]
\centering
\includegraphics[width=1\columnwidth]{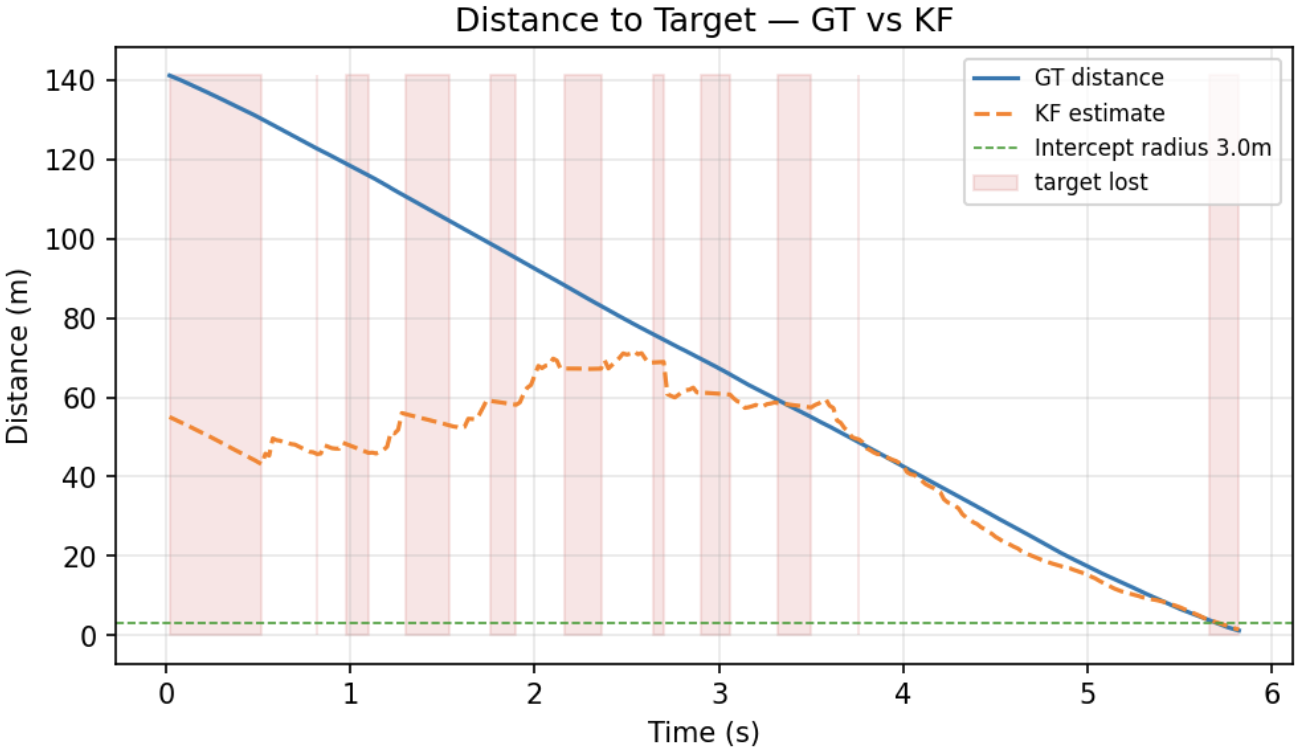}
\caption{The ground truth (GT) range (blue) versus the Kalman Filter (KF) range estimate (dashed orange). The red areas indicate periods where the target is not being detected. The dashed green line indicates the range threshold for interception.}
\label{fig_RangeEstimate}
\end{figure}

\begin{figure}[!t]
\centering
\includegraphics[width=1\columnwidth]{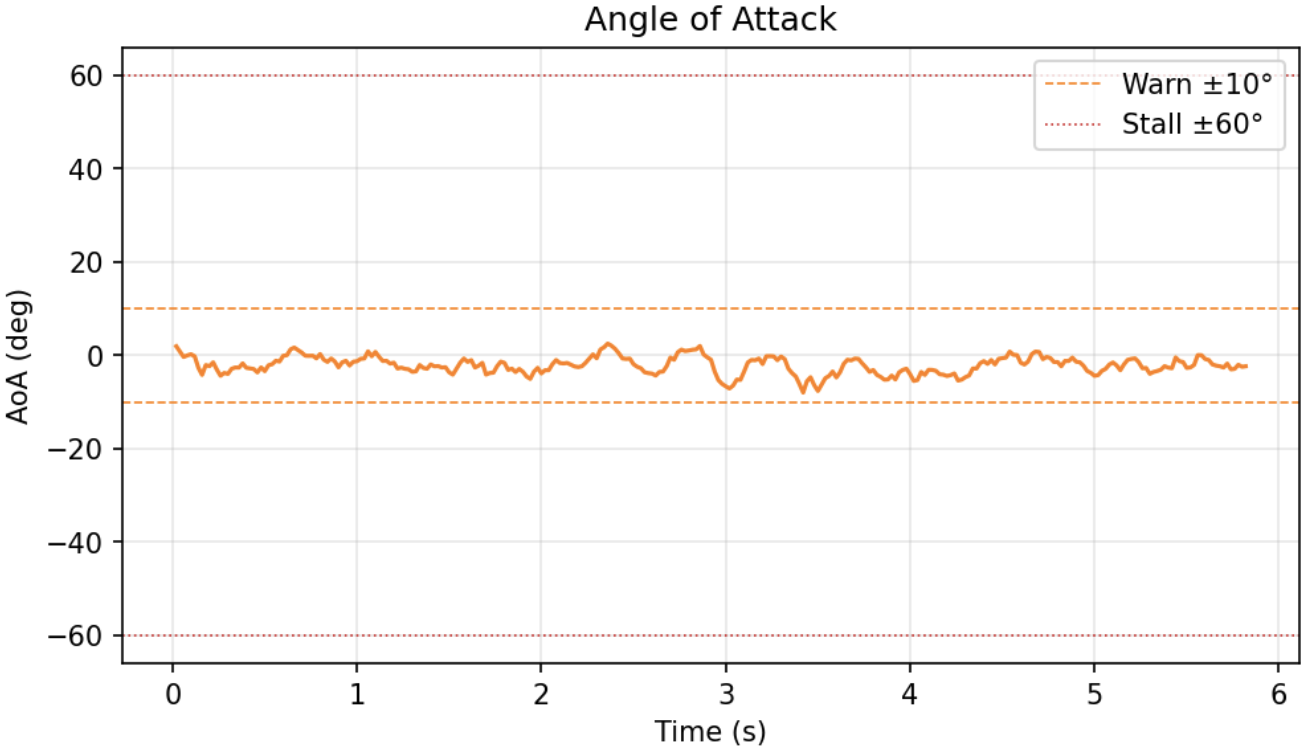}
\caption{The angle of attack throughout the simulated engagement. The dashed lines indicate the angle thresholds beyond which high absolute angle of attack is punished. The red dotted line indicates the cutoff threshold to end the episode early due to stall or airframe collapse.}
\label{fig_AOAinterception}
\end{figure}

\begin{figure}[!t]
\centering
\includegraphics[width=1\columnwidth]{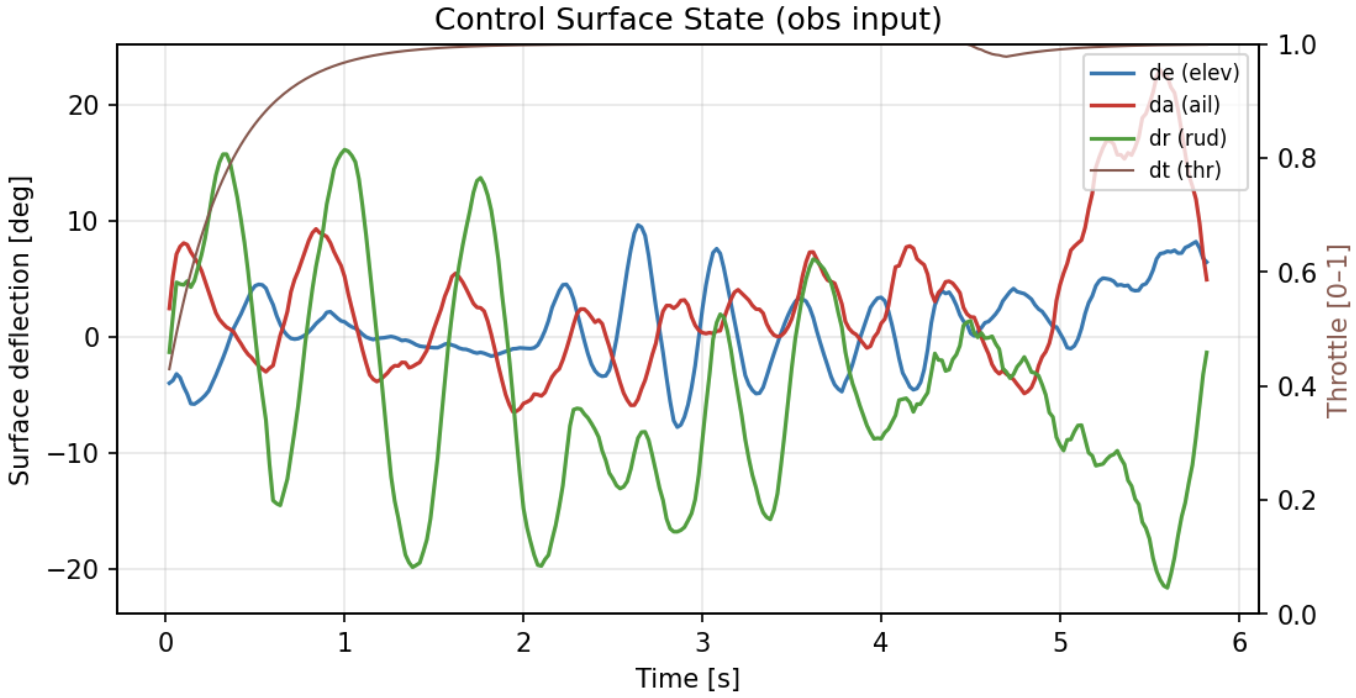}
\caption{The state of the actuators throughout the simulated engagement. Aileron, rudder and elevator deflection are represented by the red, green and blue lines respectively. Throttle percentage [0,1] is shown by the brown line.}
\label{fig_ControlSurfaces}
\end{figure}

\begin{figure}[!t]
\centering
\includegraphics[width=1\columnwidth]{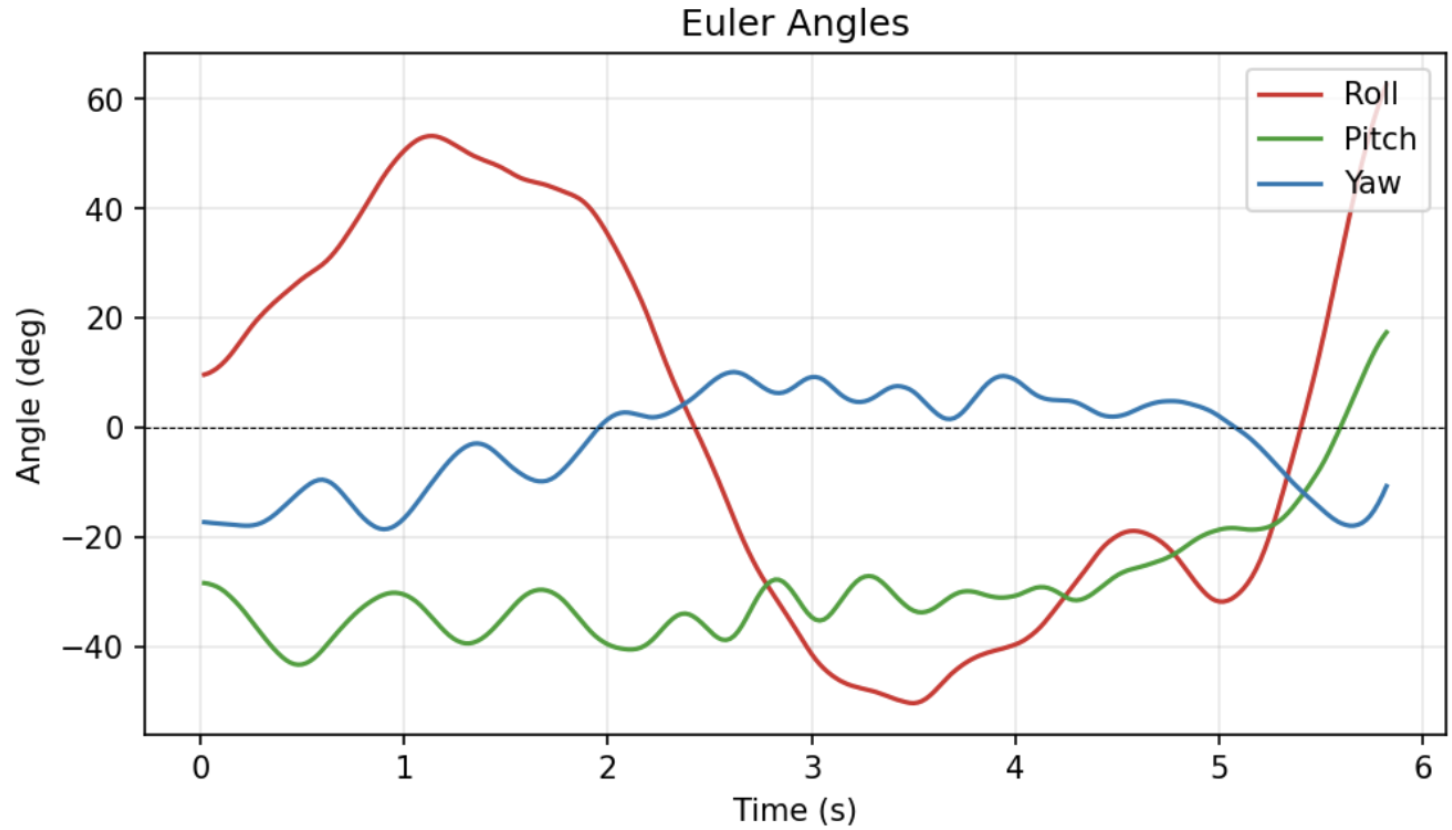}
\caption{The interceptor's attitude in Euler angles throughout the simulated engagement. Roll, pitch and yaw are indicated by red, green and blue lines respectively.}
\label{fig_EulerAnglesInterception}
\end{figure}

Figure \ref{fig_RangeEstimate} shows the ground truth distance to the target versus the estimated distance produced by the KF, with the red areas indicating periods of no detection. Even during these blind moments, the KF produces a reasonable closing distance estimate that follows the ground truth, leading to a successful interception. Figure \ref{fig_AOAinterception} showcases that AOA throughout the course of the engagement stays within the ±10 degree penalty thresholds. This shows the policy has learned to fly conservatively, which is as intended. The interceptor's airspeed varied between a minimum of 21 and a maximum of 33$\,$m/s, which is well above the aircraft's stall speed at these relatively low AOA values. 

Figure \ref{fig_ControlSurfaces} shows the state of the actuators throughout the engagement. The policy commands full throttle immediately. This is a behavior that was consistent across every policy trained throughout the course of this research. However, this particular policy will reduce throttle if the target is lost for a prolonged period of time. This behavior started occurring once the reward for quick interceptions was reduced and the reward for keeping the target inside the FOV was increased. 

Besides throttle, the flight control surfaces all show slight oscillations. Interestingly, the policy aggressively oscillates rudder deflection, which causes subsequent oscillatory yawing behavior as can be seen in figure \ref{fig_EulerAnglesInterception}. Earlier policies that were trained with less strict AOA penalties showed the same behavior but rather in the pitch axis, with the elevator being excited continuously up to its limits. 

\begin{figure}[!t]
\centering
\includegraphics[width=1\columnwidth]{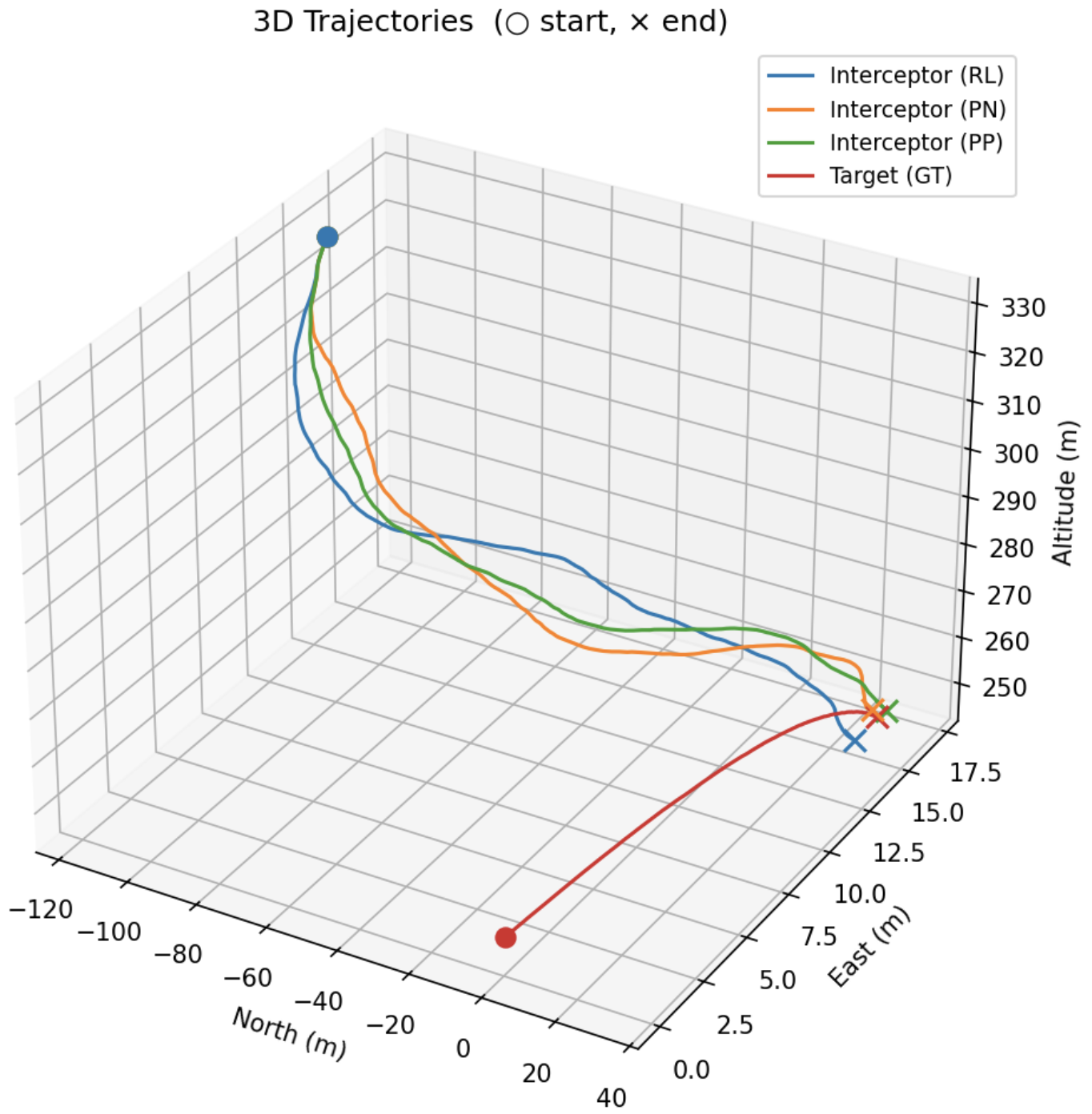}
\caption{The trajectories of the RL interceptor (blue), PN interceptor (orange), PP interceptor (green) and target (red) throughout the course of a single engagement. The circle and cross indicate the start and finish respectively. Noise and other randomness was seeded identically for each interceptor.}
\label{fig_3Ctrajectories}
\end{figure}

\begin{figure}[!t]
\centering
\includegraphics[width=1\columnwidth]{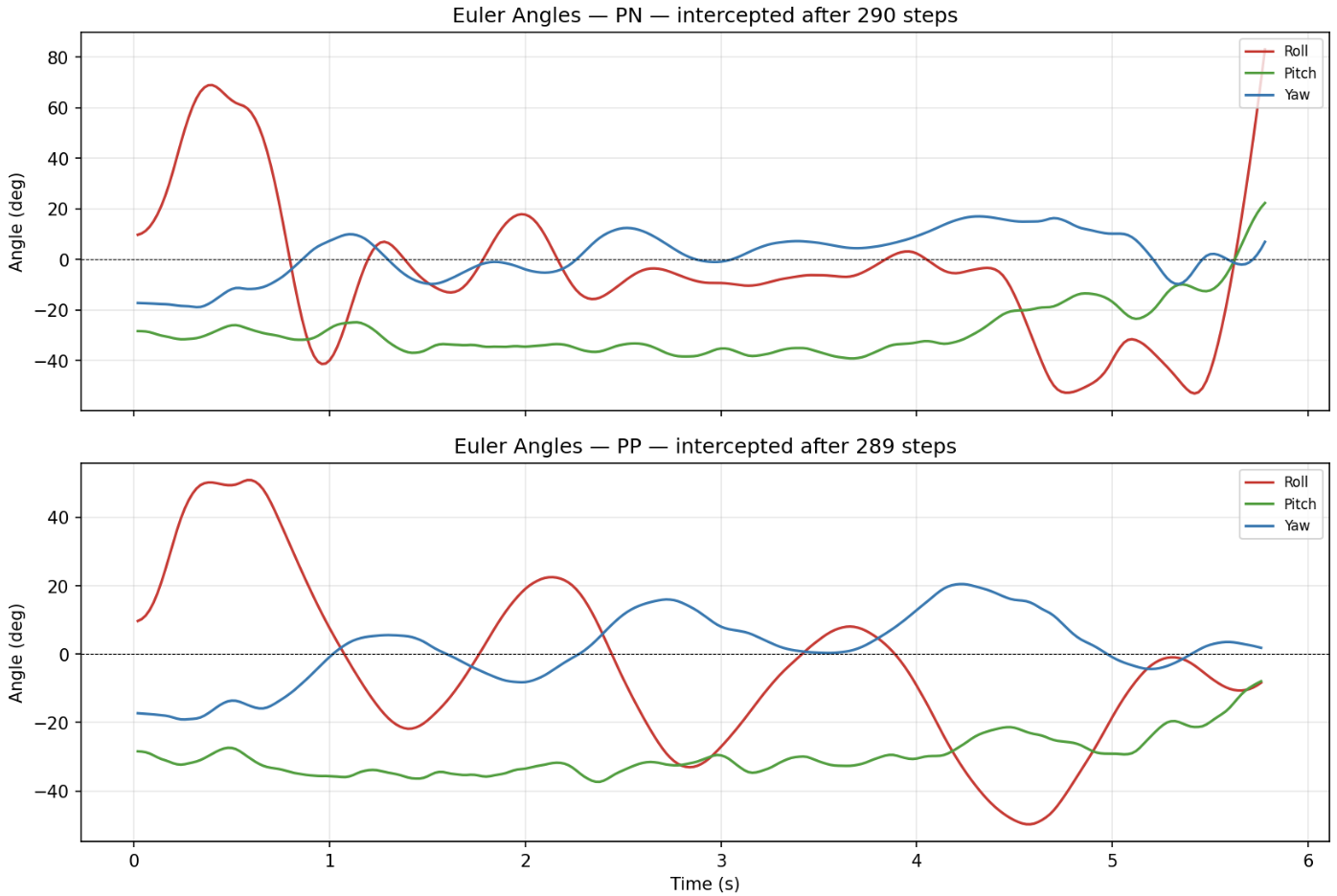}
\caption{The Euler angles of the classically guided interceptors during the same simulated engagement. Roll, pitch and yaw are represented in red, green and blue respectively. The top and bottom subfigures shows the PN and PP interceptors respectively. }
\label{fig_3CEuler}
\end{figure}

\begin{figure}[!t]
\centering
\includegraphics[width=1\columnwidth]{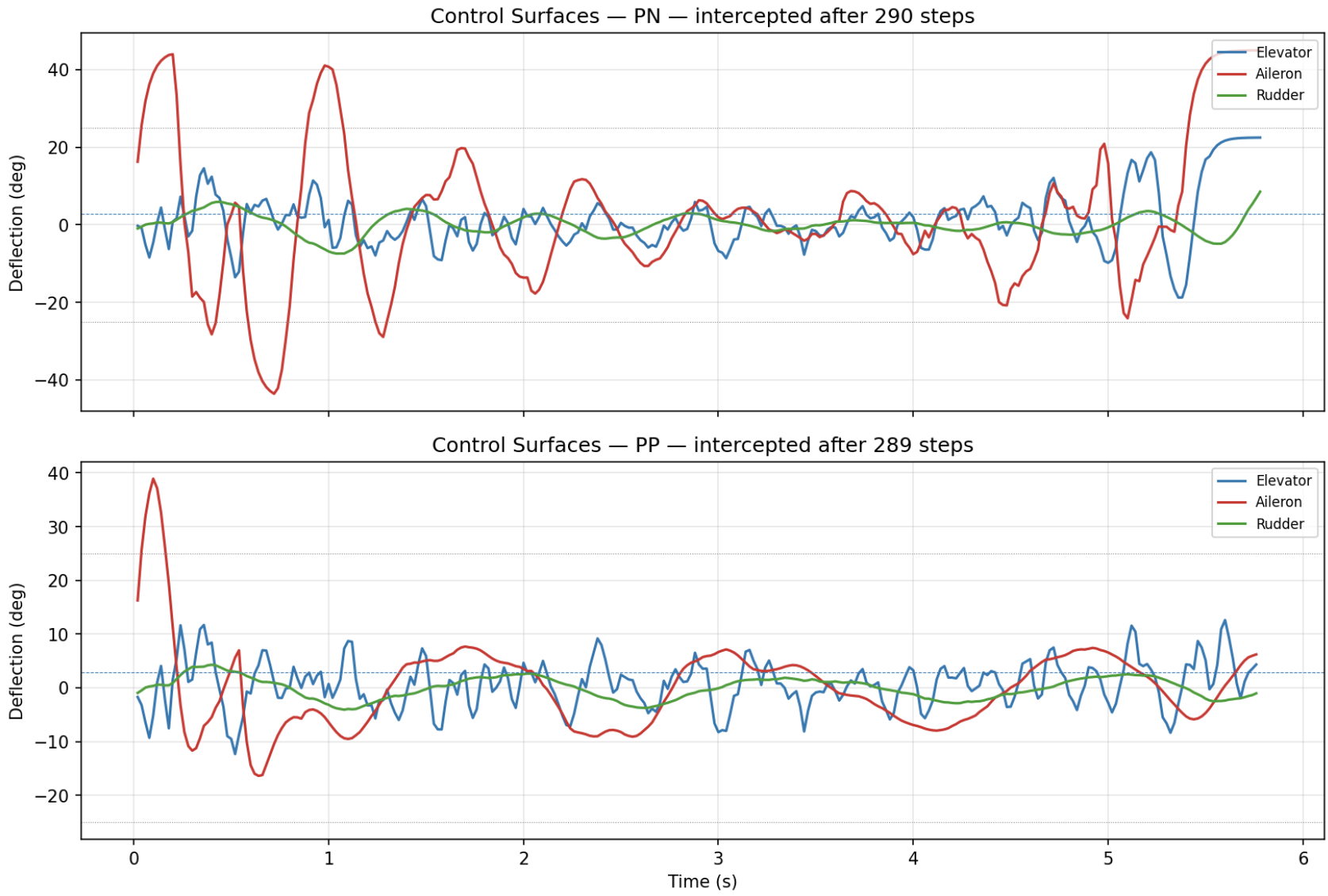}
\caption{The control surfaces of the classically guided interceptors during the same simulated engagement. Elevator, aileron and rudder are represented in blue, red and green respectively. The top and bottom subfigures shows the PN and PP interceptors respectively. Throttle is set to maximum at the start of the engagement for both controllers.}
\label{fig_ClassicalControlSurfaces}
\end{figure}

Figure \ref{fig_3Ctrajectories} shows the same engagement but with the trajectories generated by the baseline classical controllers also included. All three interceptors manage to intercept the target during this engagement. Figure \ref{fig_3CEuler} shows the Euler angles of the classically guided interceptors throughout the engagement. Figure \ref{fig_ClassicalControlSurfaces} shows the control surfaces of the classically guided interceptors over the course of this episode. 

\subsubsection{Multiple episodes analysis}
\label{sec: multiple episode analysis}

\begin{figure}[!t]
\centering
\includegraphics[width=1\columnwidth]{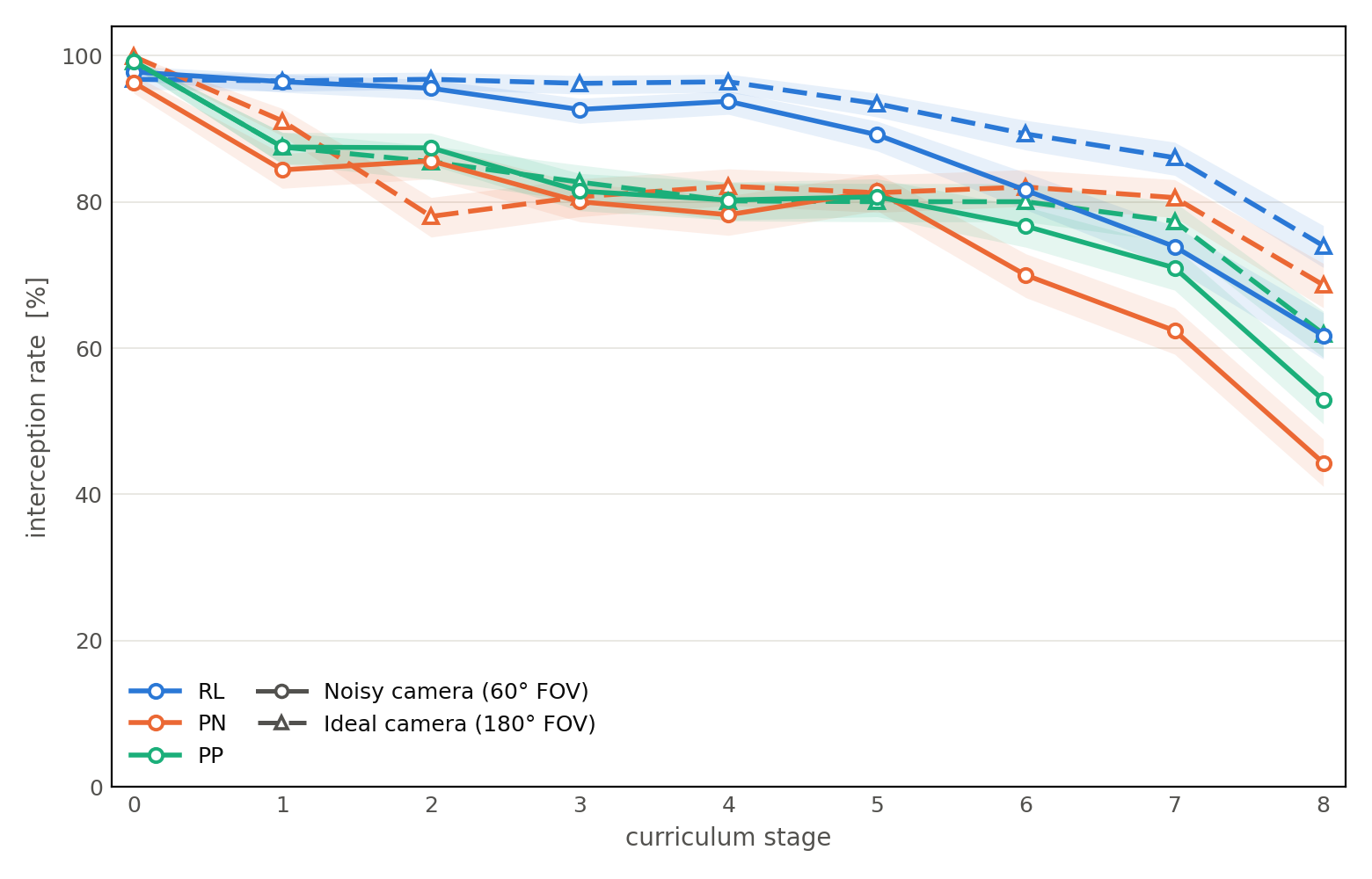}
\caption{The performance of the RL (blue), PN (orange) and PP (green) controllers throughout the nine stages. The solid lines and circles represent performance with a 60$^\circ$ FOV, noisy camera undergoing increasing disturbances as specified in appendix \ref{app: curriculum}. The dashed lines and triangles represent performance with an ideal camera with a 180$^\circ$ FOV, with perfect detections at all stages. Each point represents 896 simulated episodes (64 parallel environments running 14 episodes each). The interception rate plotted on the y-axis represents the fraction of those episodes ending with a successful interception. The shaded area represents the 95 percent confidence interval.}
\label{fig_RLversusClassicalCurriculum}
\end{figure}

The trained RL policy and the baseline classical controllers were tested in the curriculum that the policy was trained in. In figure \ref{fig_RLversusClassicalCurriculum} the performance of each controller throughout the curriculum is shown. To show the effects of limited sensing through the noisy, 60$^\circ$ horizontal FOV monocular camera, the performance is also shown with an idealized 180$^\circ$ FOV camera, which always detects the target as long as it is in the FOV. For each controller, each of the nine curriculum stages and each of the two camera types, 896 episodes were simulated, for a total of 48,384 episodes. 

All controllers perform nearly perfect at stage 0. At this stage, there is very little perceptual noise, no wind or gusts, and the target is near stationary and not accelerating at a low range from the interceptor. As the curriculum progresses, all controllers lose performance, but not equally. Overall, RL performs best and is only matched in the final stages by ideal PP and even outperformed slightly by ideal PN. Ideal RL is even better and degrades much slower. 
PP and PN perform near equally until stage 5, after which PN degrades more rapidly than PP. Real PP is less affected by camera noise than PN as its guidance law commands it to keep the target in view, resulting in robustness despite noisy detections.

To draw meaningful conclusions about the behavior and performance of the controllers in general, 14,256 additional episodes were simulated to investigate the effect of a single variable condition. 336 episodes were generated with nominal conditions similar to stage 5 of the curriculum. Each condition comprises 144 episodes: 5 conditions at increasing target ground speed (5$\,$m/s increments) and no acceleration, 5 at increasing target acceleration (1.15$\,$m/s$^2$ increments), 5 at increasing wind strength (2.25$\,$m/s increments), 5 with increasingly long clustered camera dropout bursts (up to 2.8$\,$s), 4 with 50$\,$m initial range increments, and 5 with varying approach geometry from tail-chase to head-on.

\begin{figure}[!t]
\centering
\includegraphics[width=1\columnwidth]{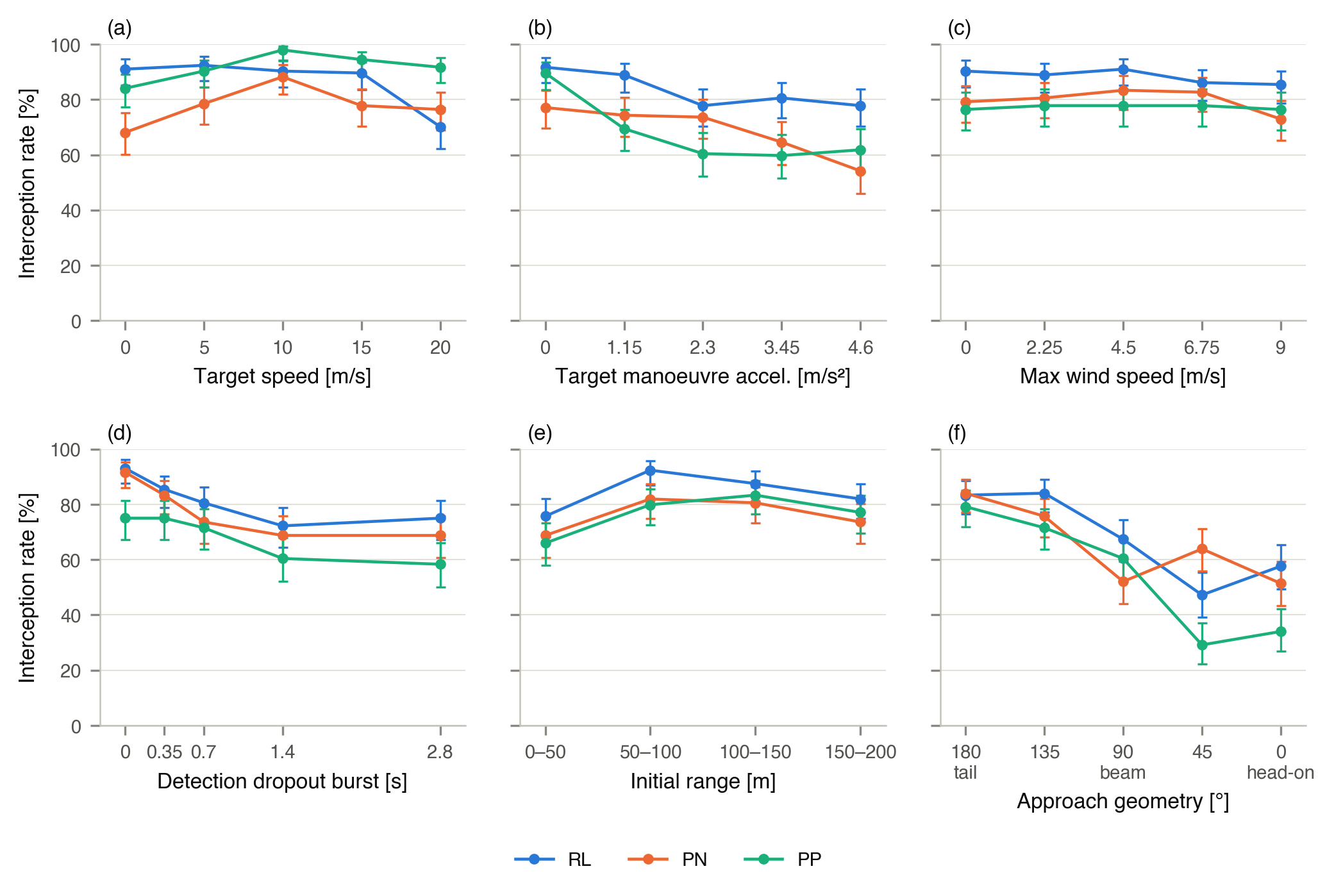}
\caption{Interception rates for increasingly difficult conditions. The error bars indicate the Wilson 95 percent intervals. The nominal conditions of stage 5 hold except for one variable. In (a), the interception rate is shown for increasingly faster, non-accelerating targets. In (b), the target is increasingly accelerating with a mean initial speed of 7.3$\,$m/s. In (c), the max wind speed increases up to 9$\,$m/s. In (d), the mean detection dropout length increases up to 2.8$\,$s. In (e), the range is increased. In (f), the approach geometry is varied from a tail-chase to a head-on engagement.}
\label{fig_Robustness_all}
\end{figure}

Figure \ref{fig_Robustness_all} shows the interception rate under these conditions. Target speed has little effect over most of its range, but the three controllers differ. RL seems to degrade only at target speeds of 20$\,$m/s, whereas PN degrades slightly at target speeds of 15$\,$m/s, but recovers at speeds of  20$\,$m/s. PP outperforms the other controllers at higher target speeds.
On the contrary, target acceleration seems to have a real degrading effect on the performance of the controllers, with higher accelerations leading to lower interception rates for all controllers. However, the classical controllers are more affected than RL in terms of performance.
Wind affects each controller differently. PP and RL are mostly unaffected across the whole sweep, but PN drops below 60 percent interception rate at 9$\,$m/s winds.
Increasingly, long camera dropout bursts lead to decline in performance of all controllers, but surprisingly this decline plateaus beyond 1.4$\,$s dropouts. 
Initial range is non-monotone. Close range initialization provides little time to correct the initial geometry leading to relatively low performance. The highest performance is at 50-150$\,$m for all controllers.
Finally, approach geometry produces the largest spread in the figure, and its minimum lies at 45$^\circ$ rather than head-on for RL and PP. RL falls from 83.3 percent in a tail chase to 47.2 percent at 45$^\circ$, recovering to 57.6 percent head-on. PP is worst affected, dropping to 29.2 percent and 34.0 percent at these two geometries, while PN is the least sensitive and even outperforms RL at 45$^\circ$. PP performing the worst is expected as it always aims to nullify the LOS with respect to the boresight of the camera and hence will result in tail chases. A head-on initialization will therefore likely result in lots of near misses and also loss of target lock, as the interceptor can be too slow in turning around to reacquire the target after passing it. Interestingly, the performance of PN and RL are nearly equal at head-on and 45$^\circ$ engagements, suggesting that RL adapts it pursuit strategy based on the closing velocity. 

\begin{figure}[!t]
\centering
\includegraphics[width=1\columnwidth]{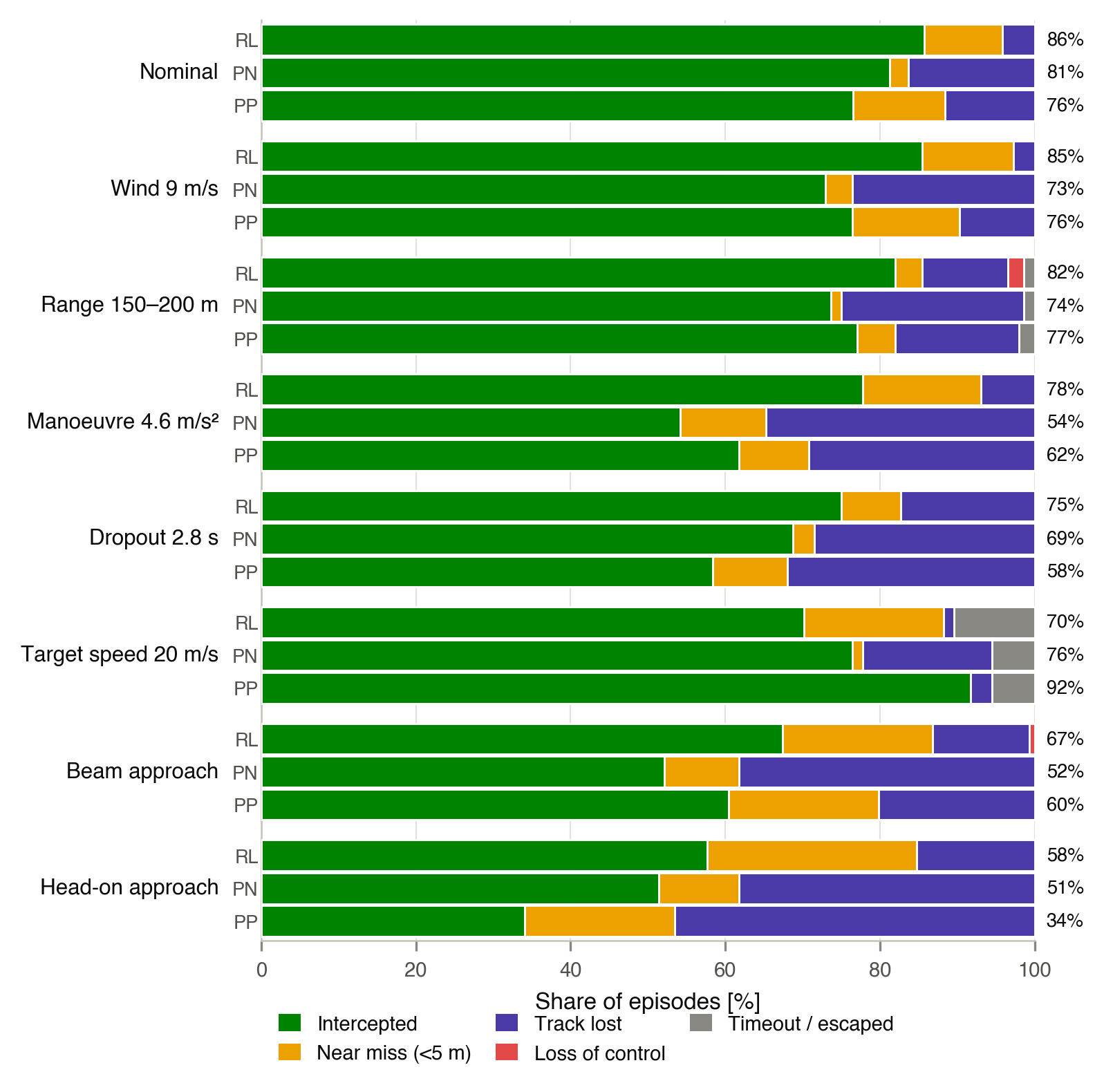}
\caption{A breakdown of the reasons episodes were terminated, categorized by condition. Green, purple, orange, red and grey represent the fraction of episodes terminated due to successful interception, target loss, near misses, loss of control and timeout.}
\label{fig_Outcomes_all}
\end{figure}

Figure \ref{fig_Outcomes_all} gives a breakdown of the reasons for termination of episodes per condition and per controller. Across all conditions, RL outperforms the other controllers with the exception of 20$\,$m/s target speed, in which the classical controllers are better. 
9$\,$m/s wind barely has an impact on performance in RL and PP. PN however degrades from 81 percent to 73 percent interception rate. 

Longer initial range has an impact on performance. With initial ranges between 150 to 200 meters, the performance of RL drops slightly to 82 percent interception rate, with a small portion of episodes surprisingly being terminated due to loss of control. At increasingly higher ranges, the KF range estimate becomes unstable, such that it produces estimates far beyond the training envelope, causing erratic behavior. For the classical controllers, a longer range means more engagements terminate in a target loss, which is expected as the detection probability is lower at longer range.

A quickly accelerating target also reduces interception rates for all controllers, mostly due to target loss. RL is most robust against maneuvering targets and is still able to intercept 78 percent of the time with a majority of the other episodes resulting in a near miss. Surprisingly, a maneuvering target does not cause episodes to be terminated in a loss of control. Unsuccessful terminations are due to near miss and target loss. PN and PP struggle with maneuvering targets with interception rates of 54 and 62 percent respectively.
Extended periods of camera dropouts impact the interception rate of all controllers, leading mainly to more target loss.
In training, the agent sees mainly tail-based engagements, where the target is moving away from the interceptor's initial position. However for evaluation, beam (sideways) and head-on approaches were also simulated. These show a significant reduction in performance, with both a large amount of near misses and target losses. PP is the worst affected during head-on approaches. 

Finally, the condition where the target has a speed of 20$\,$m/s (excluding acceleration) is examined. Unsurprisingly, some episodes terminate in a timeout, as the interceptor can lack the speed to catch the target before the time runs out, however RL is more affected than PN and PP. The latter even improves as target speed increases. This is likely due to PP resulting in a tail-based engagement. When the target has a higher speed, the relative velocity will be lower, allowing for more precise terminal guidance. This is reflected by the fact that there are also no near misses for this condition.

\begin{figure}[!t]
\centering
\includegraphics[width=1.05\columnwidth]{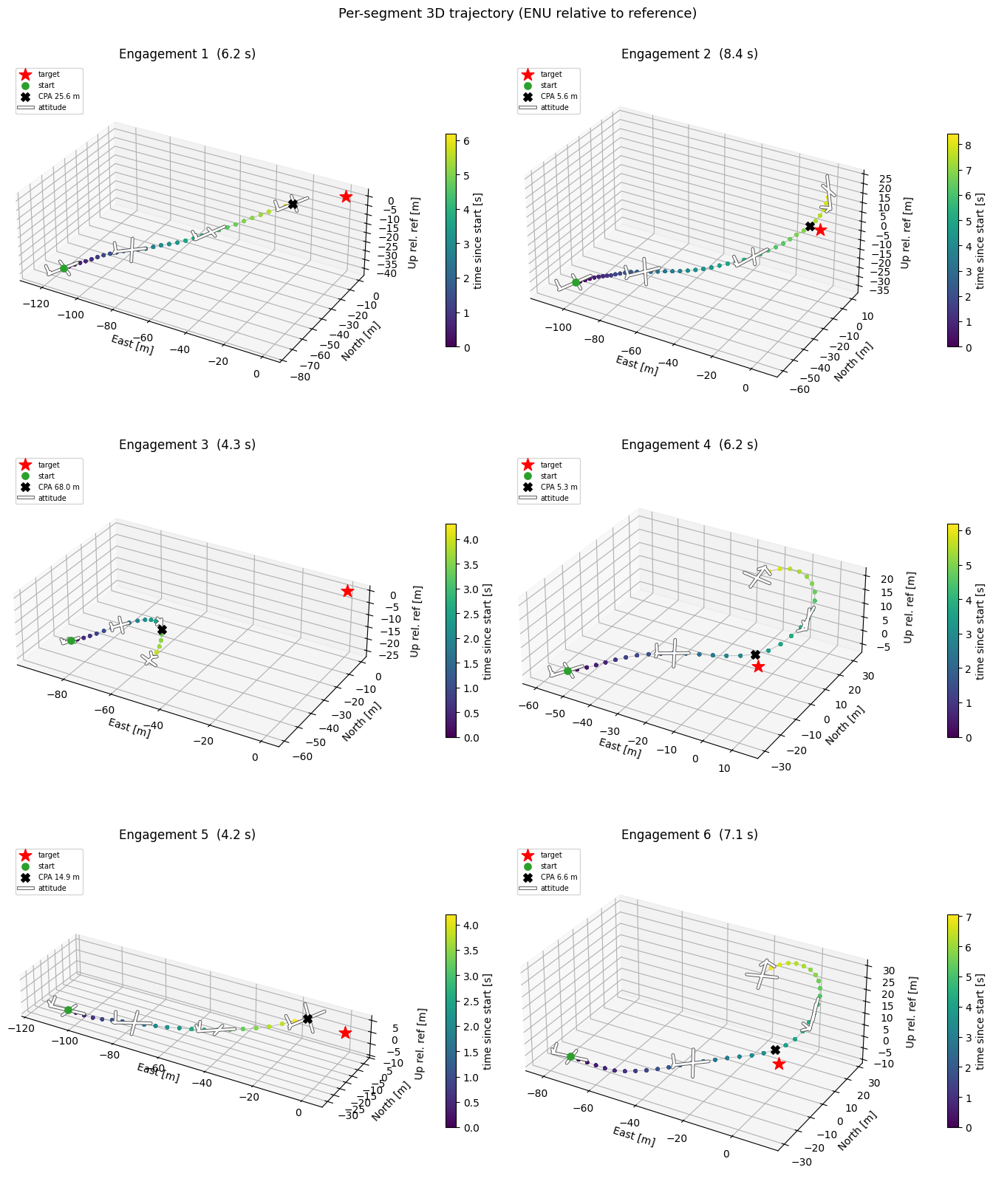}
\caption{3D trajectories of the interceptor during 6 real flight engagements of a stationary virtual target. The green dot represents the start of the engagement, that is the interceptor is flying in MANUAL mode and the RL controller is overriding RC. The red star indicates the relative position and altitude of the virtual target. The black cross indicates the closest point of approach (CPA) in meters. The viridian color scheme shows how the trajectory progresses through time. The aircraft in white with grey outline is a schematic representation of the interceptor's attitude at discrete time steps. The aircraft model is not to scale.}
\label{fig_StationaryTarget}
\end{figure}

\subsection{RL controller in real flight}
Initial in flight testing of the RL controller proved to be difficult. A moving fixed-wing target was hard to keep in the FOV whilst engaging the RL policy. Furthermore, unsuccessful attempts could either be due to a sim-to-real gap in the controller, or the perception could be insufficient. 

\begin{figure}[!t]
\centering
\includegraphics[width=1\columnwidth]{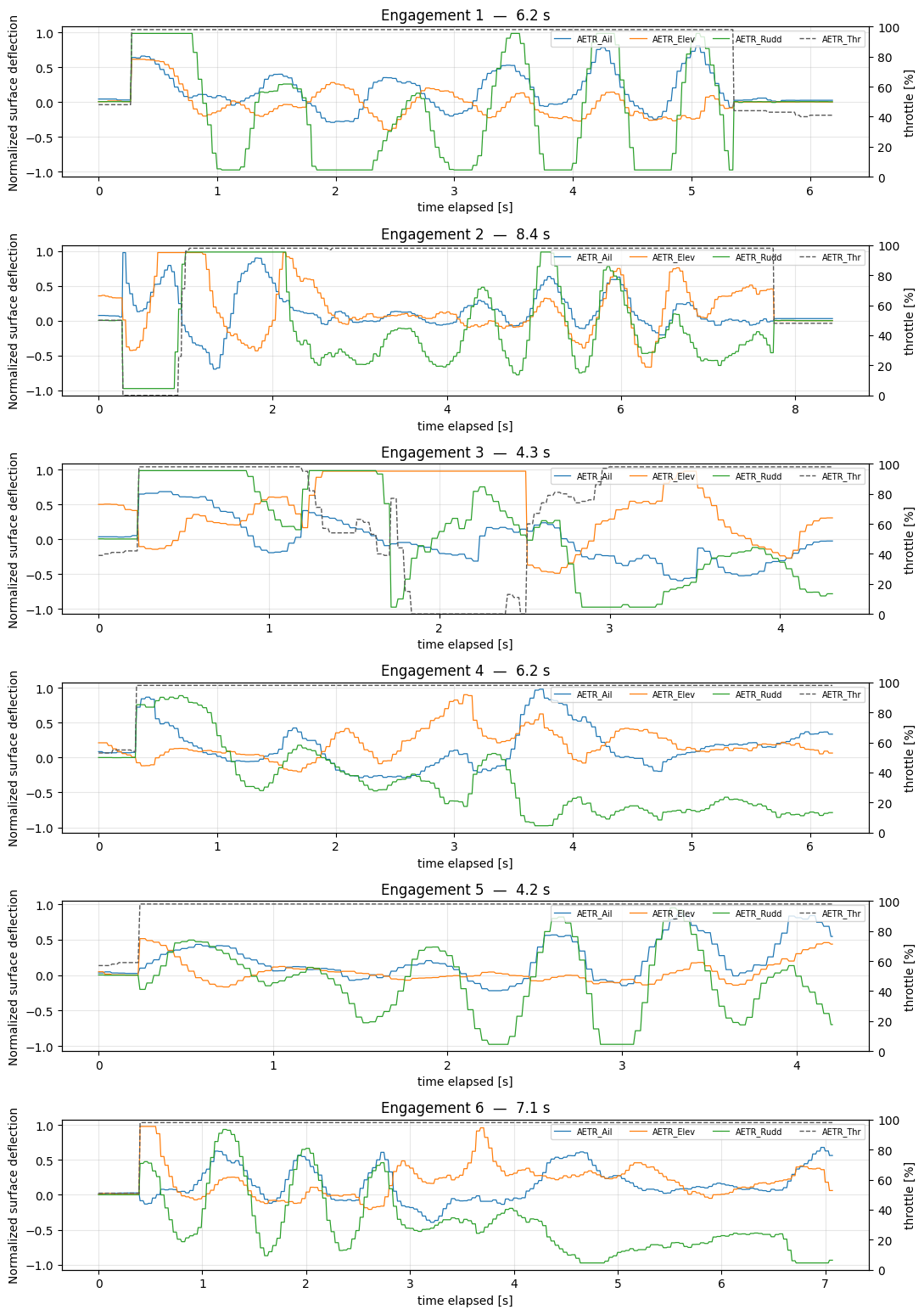}
\caption{The commanded flight surface deflections and throttle during each real flight engagement of a virtual stationary target. In blue, orange and green, the normalized surface deflection of the ailerons, elevator and rudder respectively. The dashed grey line represents the commanded throttle.}
\label{fig_ControlsStationaryTarget}
\end{figure}

\subsubsection{Stationary virtual target}

Flight tests with a virtual target resolved this issue. During the first successful interception tests, the video\footnote{\url{https://youtu.be/zMv574in_xE}} showed the interceptor passing through the target 5 out of 6 times. 

Figure \ref{fig_StationaryTarget} shows the trajectories of the interceptor during all 6 real flight engagements of a stationary virtual target. Engagements 2, 4 and 6 show the interceptor passing through the virtual target barring a few meters error. Engagements 1 and 5 were stopped just before the target was being passed through because the pilot switched to STABILIZED mode, meaning the RL controller is interrupted. However as the onboard video recording shows, the interceptor still passed through the target shortly after without pilot stick inputs. Only engagement 3 shows a clear failure with the interceptor seemingly turning away at about 70 meters from the target. This could be due to a bad KF estimate of the target, however this can not be confirmed as the KF estimation was not logged during this flight. Also visible in figure \ref{fig_StationaryTarget} is the fact that the controller does not stop after the target is intercepted. In training and simulation, the engagement stops, but in real flight, the controller experiences a flight through the target and after passing quickly tries to reacquire it in its FOV with an aggressive maneuver (visible in engagements 2, 4 and 6).

Figure \ref{fig_ControlsStationaryTarget} shows the actuator commands by the RL controller during each engagement. As observed in simulation, the policy heavily oscillates the rudder. This is also observed on the onboard video which shows the plane yawing back and forth significantly. Throttle is mostly full, except in engagement 2 and 3. In engagement 2, the throttle drops immediately after the policy takes control, only to recover quickly to full throttle after 0.7$\,$s. In engagement 3, the throttle is reduced for a longer period of time. This points to a loss of target track, as the same behavior is observed in simulation when a target is lost for a prolonged period. 

With the RL policy being able to intercept a virtual target, the sim-to-real transfer has been successful controller-wise. The policy produces flight behavior that mimics simulation. It successfully avoids high body rates and erratic control inputs. One of the key insights was to ensure that the deployment scripts running on the interceptor provided the same telemetry frequency as the policy was trained on. Earlier flight tests provided 10 to 20 Hz telemetry, whereas the policy was trained on 40-50 Hz. 

\subsubsection{Moving virtual target}

\begin{figure}[!t]
\centering
\includegraphics[width=1\columnwidth]{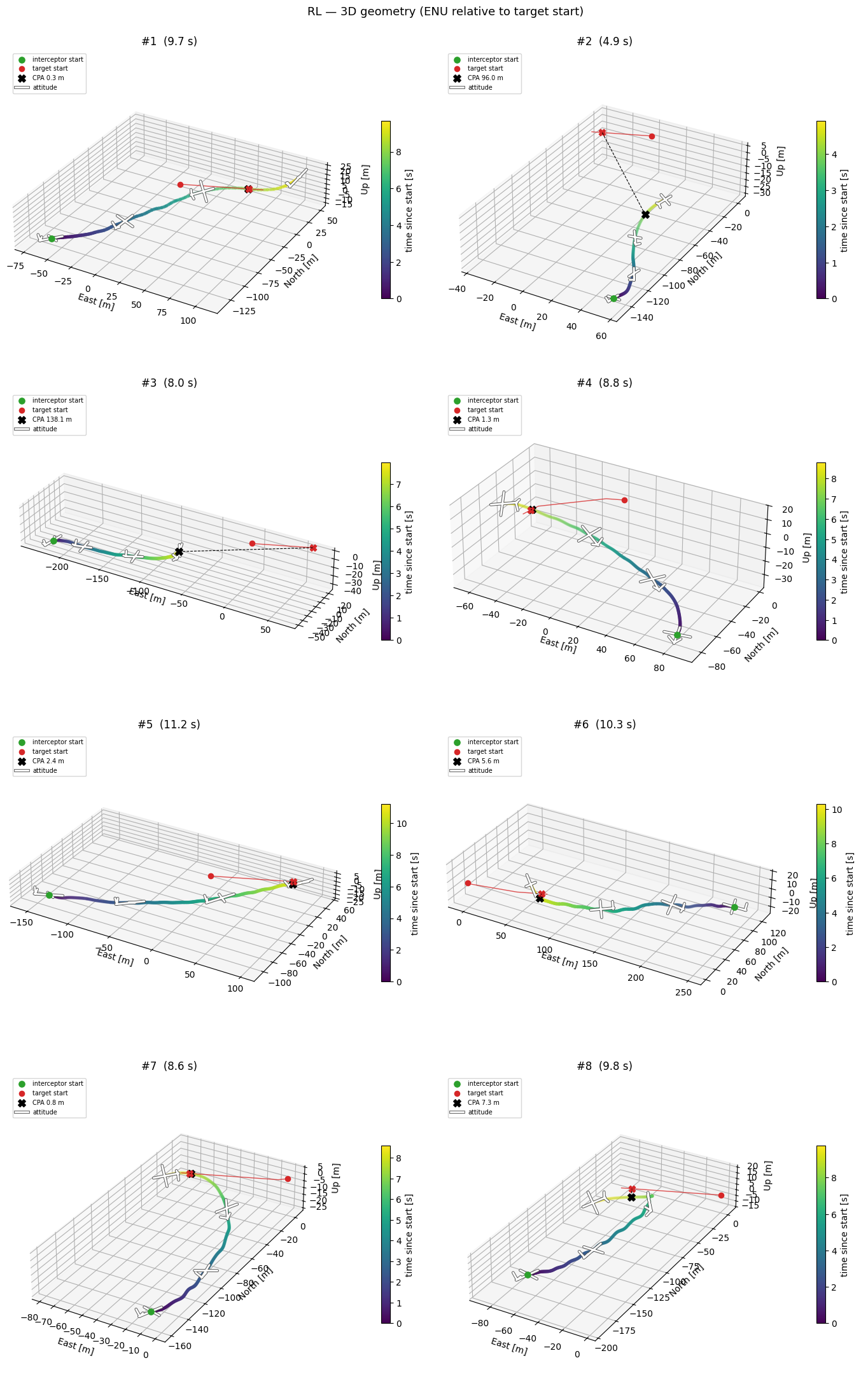}
\caption{3D trajectories of the interceptor during 8 real flight engagements of a moving virtual target. The green and red dot represent the interceptor's and target's position respectively at the start of the engagement. The red and black cross indicate the closest point of approach (CPA) of the target and interceptor respectively. The grey dashed line shows connects the target and interceptor at the CPA. The red line shows the target's trajectory. The viridian color scheme shows how the interceptor's trajectory progresses through time. The aircraft in white with grey outline is a schematic representation of the interceptor's attitude at discrete time steps. The aircraft model is not to scale.}
\label{fig_movingtargettrajectories}
\end{figure}

\begin{figure}[!t]
\centering
\includegraphics[width=1\columnwidth]{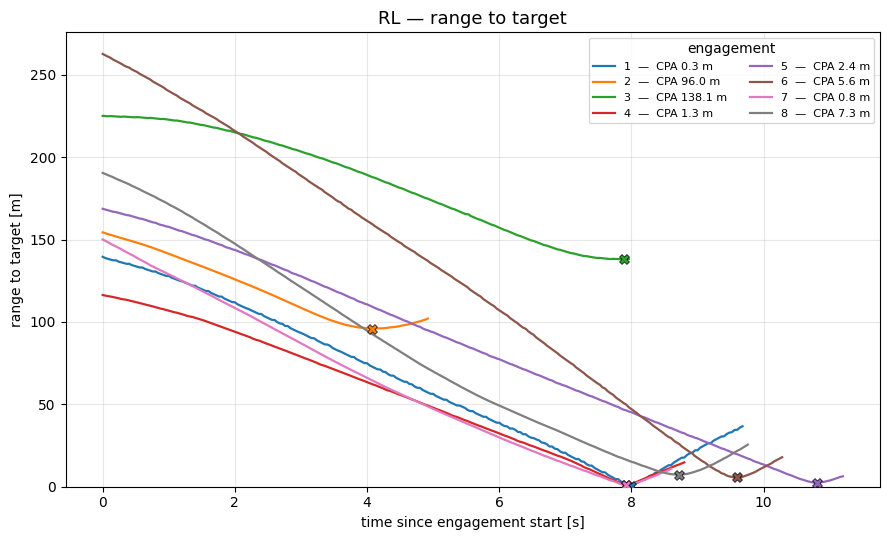}
\caption{The distance between the interceptor and the target throughout the course of the eight engagements. Each colour indicates a separate engagement, with the cross in the corresponding colour representing the shortest distance between the interceptor and the target. CPA: closest point of approach.}
\label{fig_rangemovingtarget}
\end{figure}

After successfully intercepting virtual stationary targets, the controller was challenged with a moving virtual target. The target was flying a 90 meter altitude loop at a fixed ground speed of 8$\,$m/s through a set of coordinates that defined an oval-shaped track. Eight attempts were made, of which four resulted in an interception (distance $\leq$ 3$\,$m), two resulted in a miss (3$\,$m $\leq$ distance $\leq$ 8$\,$m) and two were aborted due to divergence. The trajectories of the interceptor and virtual target during each engagement are shown in figure \ref{fig_movingtargettrajectories}. In figure \ref{fig_rangemovingtarget}, the distance between the interceptor and the target throughout the engagement is shown. 

Overall, the controller performed well with a moving target. During the first engagement, a closest point of approach (CPA) of 0.3$\,$m was achieved. This is tenfold below the chosen interception threshold of 3$\,$m, but also far below the wingspan of the target (1.7$\,$m), indicating precision even as the target moves. During engagement four, five and seven, similar performance was observed. Engagement six and eight also showed the interceptor closing in quite well, but ultimately remained above the threshold of 3$\,$m. Engagement two and three were aborted due to the interceptor steering away from the target. 

In simulation, it was observed that tail-based approach geometries had the highest success rate. In real flight, approach geometry does not necessarily negatively influence performance. Engagement four shows an approximately sideways engagement, resulting in a successful interception (CPA: 1.3$\,$m). Engagement seven and eight start nearly head-on, with engagement seven resulting in a precise interception (CPA: 0.8$\,$m) and engagement eight resulting in a near miss (CPA: 7.3$\,$m). Engagement eight shows the interceptor making a very sharp move to try and align itself with the target. Engagement six is fully head-on and the interceptor does not try to align itself with the target's trajectory but just flies into it head on (CPA: 5.6$\,$m). 

Engagement two and three were aborted because the interceptor started to steer away from the target. This was likely due to the KF producing an unreasonable range and position estimate. Once the estimates are far beyond what the policy has seen in training, it starts to diverge. During these two engagements, the policy reduces throttle, which is consistent behavior occurring when it has lost track of the target. 

\label{sec: moving virtual target}

\subsubsection{Real target}
The ultimate test of the system was the interception of a real target. For this test, a fixed-wing target was flying a rectangular pattern at 90 meters altitude in AUTO mode. Meanwhile, the operator of the target was instructed to give stick inputs to maneuver and actively avoid being collided into. 

The interceptor was aimed at the target manually. The controller would be switched on after a target detection was confirmed on the camera feed. After switching on the controller, RC commands generated by the RL policy take over the aircraft, with pilot stick inputs having no effect until the controller is switched off again. The relative trajectory was monitored by both pilots from the ground. When the interceptor got into collision range, the controller was switched off and both pilots maneuvered in STABILIZED mode to avoid an actual impact.

A total of six engagements were flown with the RL controller engaged. Three engagements resulted in a successful interception with both pilots taking over to avoid a collision. During the first engagement\footnote{\url{https://youtu.be/zG3759-pM_g}}, the interceptor detected a bird or another aircraft, which was mistakenly assumed by the human operators to be the target drone. This resulted in normal behavior, but chasing a phantom target of which detection faded as the controller was turned on. The third interception was the first successful engagement\footnote{\url{https://youtu.be/ptbnmpXbg0c}}. Both pilots actively took over control manually to avoid a collision. The fifth engagement was a successful interception. The sixth and final attempt was also a successful interception. The interceptor banked 130 degrees during this chase (as visible in the flight logs) and was able to keep the target in sight before the pilot took back control. 

\begin{figure}[!t]
\centering
\includegraphics[width=1\columnwidth]{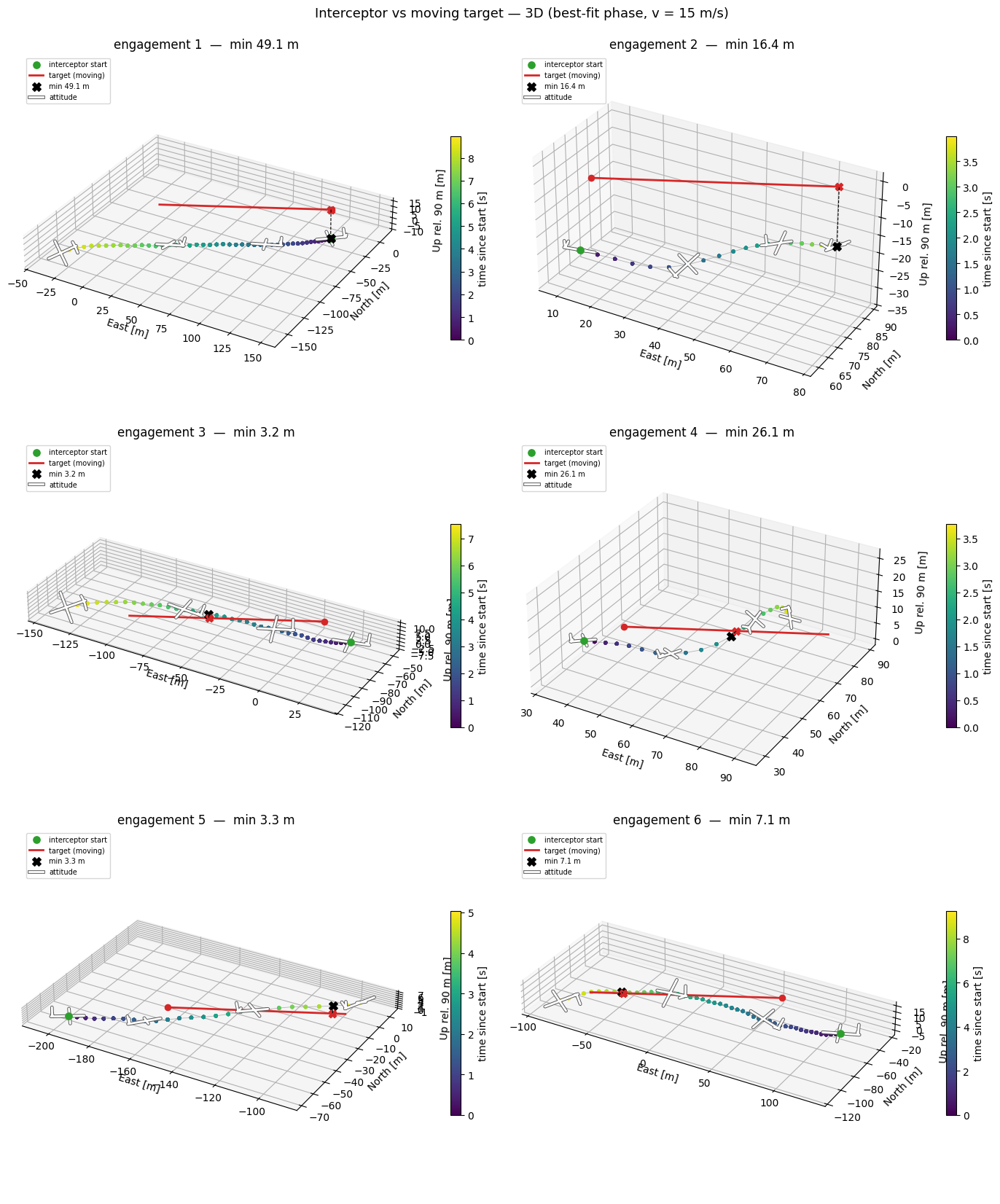}
\caption{3D trajectories of the interceptor during 6 real flight engagements of a fixed-wing target. The green and red dot represent the interceptor's and target's position respectively at the start of the engagement. The red and black cross indicate the closest point of approach (CPA) of the target and interceptor respectively. The grey dashed line shows connects the target and interceptor at the CPA. The red line shows the target's mission flight path. The CPA here is computed based on the proximity to the mission flight path, not the actual target trajectory. The viridian color scheme shows how the interceptor's trajectory progresses through time. The aircraft in white with grey outline is a schematic representation of the interceptor's attitude at discrete time steps. The aircraft model is not to scale. The target's trajectory shown here is an estimated trajectory, assuming a constant 15 m/s speed and no maneuvering or deviation from the predetermined flight path. Actual CPA may have been smaller because the target was also maneuvering which is not represented in this figure. The target trajectory is computed based on the assumption that the target had a constant speed and the red line is simply an extrapolation of where the target would have been if the theoretical CPA was indeed the closest point of approach.}
\label{fig_realInterceptions}
\end{figure}

Unfortunately, the target's flight logging got corrupted a mere 17 seconds after launch. Hence, there are no GNSS coordinates of the target's trajectory. To still get an indication of CPA, a mixture of information was used. Firstly, the target was mostly flying a rectangular pattern between four predetermined GNSS coordinates with a desired altitude of 90 meters. The target was maneuvering to make it more difficult for the interceptor, but it can be assumed that the target was approximately flying the mission. For each engagement, the CPA with this trajectory can be computed. By assuming a constant airspeed of 15$\,$m/s, the reconstructed target trajectory based on the mission can be compared to the interceptor's trajectory as shown in figure \ref{fig_realInterceptions}. 

Secondly, the video recordings can be analyzed to see how much the target drone takes up the frame. Using the pinhole model equations (\ref{eq:pinhole1}, \ref{eq:focallength}), a depth estimate can be reconstructed from the image.

\begin{figure}[!t]
\centering
\includegraphics[width=1\columnwidth]{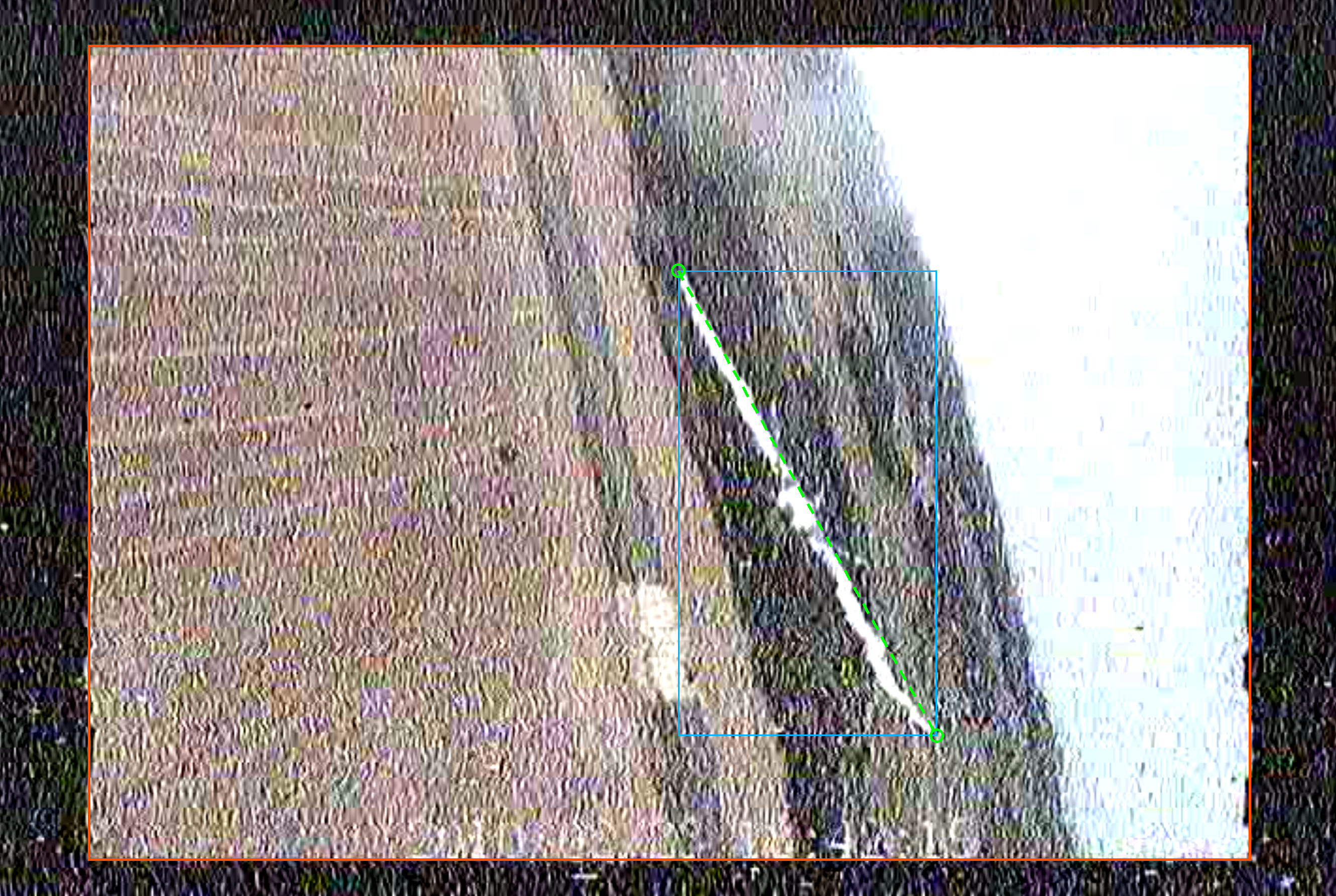}
\caption{Engagement 3. 3.3 meter range estimate. The date and time stamp are incorrect and can be neglected; they are only included to show the full frame.}
\label{fig_overlay1}
\end{figure}

\begin{figure}[!t]
\centering
\includegraphics[width=1\columnwidth]{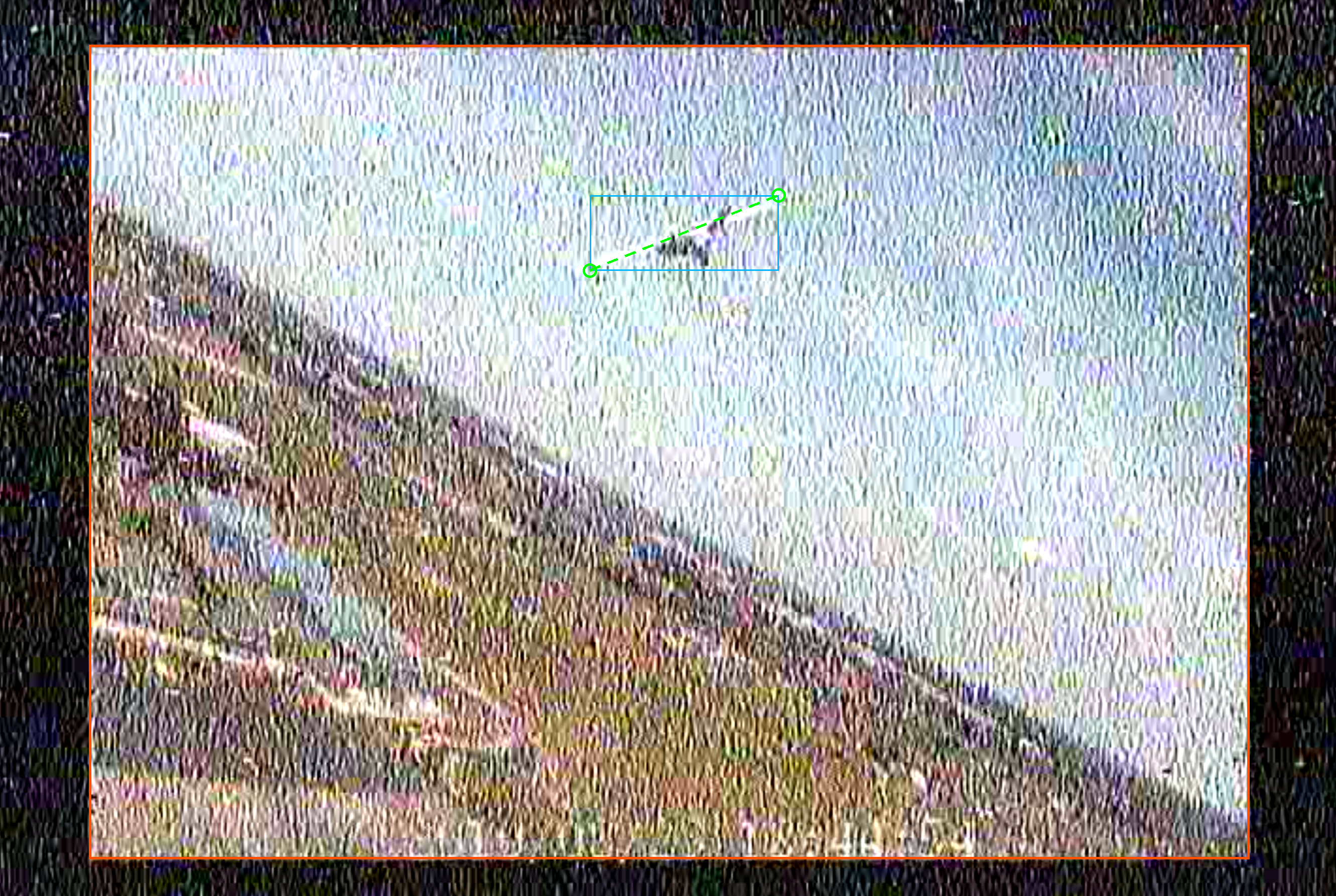}
\caption{Engagement 5. 6.0 meter range considering approximately 45 degrees of yaw with respect to the optical axis of the interceptor. The date and time stamp are incorrect and can be neglected; they are only included to show the full frame.}
\label{fig_overlay2}
\end{figure}

\begin{figure}[!t]
\centering
\includegraphics[width=1\columnwidth]{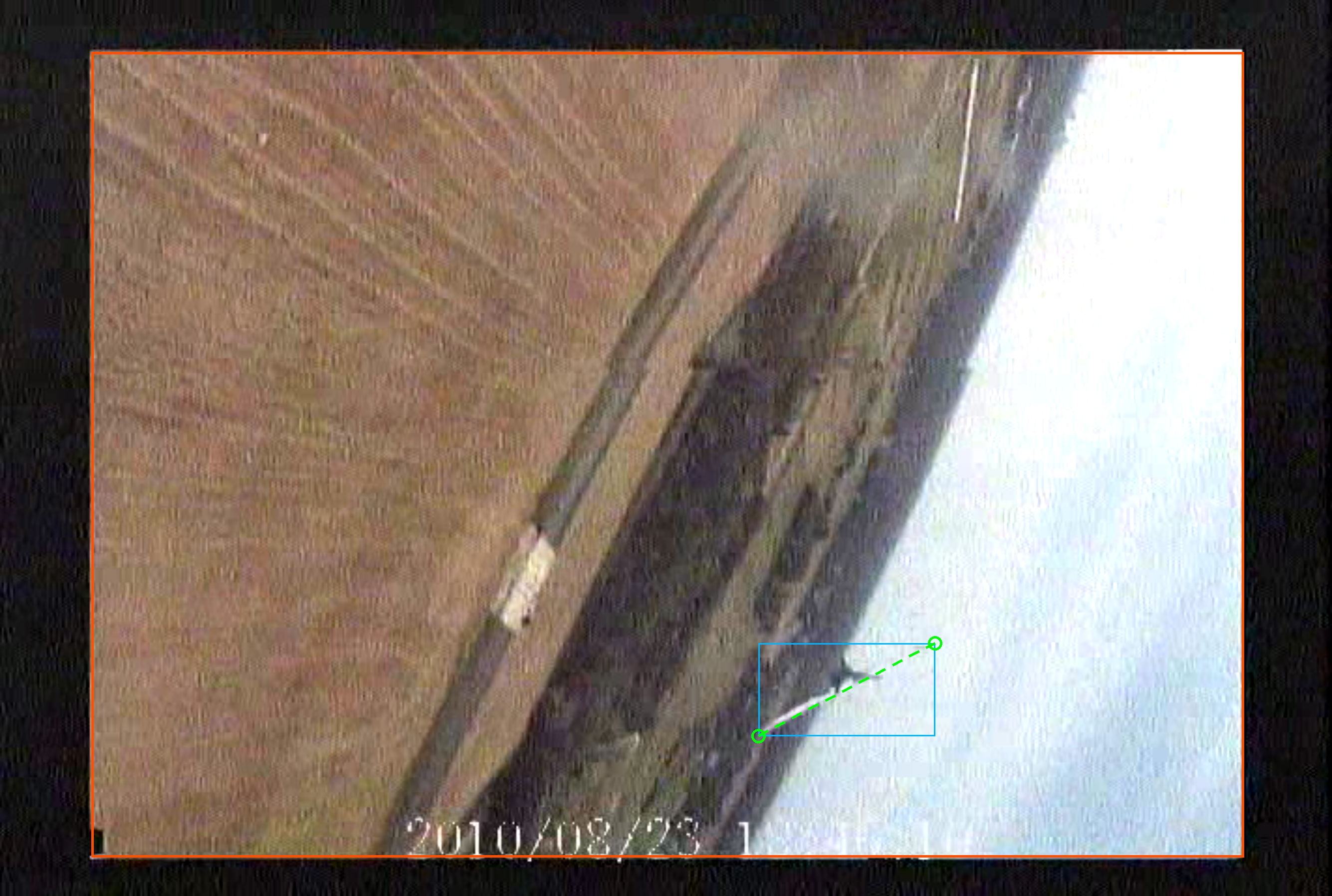}
\caption{Engagement 6. 8.1 meter range considering approximately 20 degrees of yaw with respect to the optical axis of the interceptor. The date and time stamp are incorrect and can be neglected; they are only included to show the full frame.}
\label{fig_overlay3}
\end{figure}

In figures \ref{fig_overlay1}, \ref{fig_overlay2} and \ref{fig_overlay3}, still images of the target being intercepted during engagement three, five and six respectively are shown. The interceptor comes even closer on video, but the images shown here contain relatively little static noise and also show the entire wingspan of the aircraft. For engagement three, the interceptor and target bank opposite ways to avoid a collision, meaning this range estimate will be quite accurate as the closest point of approach. For engagement five and six, the interceptor flies underneath the target. Therefore the CPA is likely closer than these images suggest.

The wingtips are identified manually in each frame, and the tip-to-tip separation $L$ measured in screen-capture pixels is converted to native camera pixels through $s = 1280/W_{\mathrm{act}}$, where $W_{\mathrm{act}}$ is the width of the active video area. With $f_x = 1108.5$~px from Eq.~\eqref{eq:focallength} and $W = 1.718$$\,$m, Eq.~\eqref{eq:pinhole1} gives $Z = f_x W / w_{px}$ with $w_{px} = sL$:
\begin{center}
\begin{tabular}{llrrr}
\hline
 & $L$ [px] & $s$ & $w_{px}$ [px] & $Z$ [m] \\
\hline
(3) & $\sqrt{515^2+927^2}=1060.4$ & $0.552$ & 585.6 & 3.25 \\
(5) & $\sqrt{376^2+150^2}=404.8$  & $0.553$ & 224.0 & 8.50 \\
(6) & $\sqrt{354^2+186^2}=399.9$  & $0.554$ & 221.5 & 8.60 \\
\hline
\end{tabular}
\end{center}
A target yawed by $\psi$ relative to the camera axis presents an effective width $W\cos\psi$, so the values above are upper bounds. Estimating $\psi \approx 45^{\circ}$ in engagement 5 and $\psi \approx 20^{\circ}$ in engagement 6 yields corrected ranges of $6.0$$\,$m and $8.1$$\,$m, respectively.

\begin{figure}[!t]
\centering
\includegraphics[width=1\columnwidth]{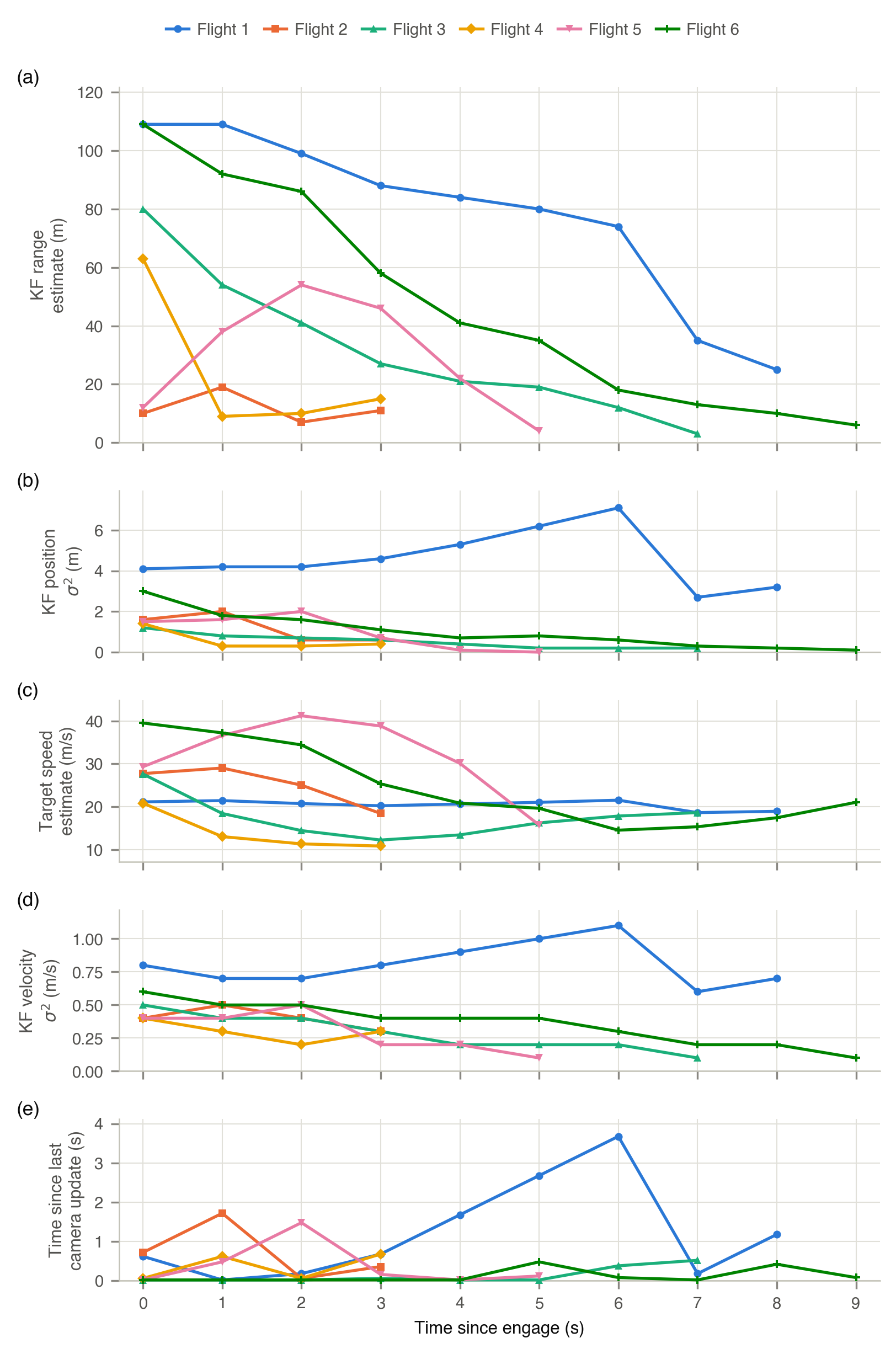}
\caption{1 Hz KF logs of the estimated range (a), position estimate variance (b), estimated target velocity (c), velocity estimate variance (d), time elapsed since the last detection (e). Each colour represents a single engagement plotted on a common time axis. }
\label{fig_KF1Hzlogs}
\end{figure}

Lastly, besides the mission and the images, the KF running on the interceptor gives a 1 Hz log containing the range estimate, which gives an indication of perceived range. In figure \ref{fig_KF1Hzlogs}, various KF parameters are plotted over time during each engagement. The top subfigure shows the KF range estimate. For engagements 3, 5 and 6 the final range estimate is 3, 4 and 6 meters respectively. This corroborates the range estimates obtained using other methods. Besides the final estimates, these plots also show how target state estimation progresses over time during real interception flights. Engagements 1, 3 and 6 show a very similar pattern in the range estimate, which starts high and gradually decreases over time. This explains why during flight 1, the interceptor did not cut throttle: the filter's estimate remained consistent with a closing engagement throughout. However, the video shows the detector had locked onto something else in the sky. The detection dropout plot shows that six seconds into the engagement, the target has been lost for 3.7 seconds. Surprisingly, the interceptor keeps flying normally on the filter's estimate rather than breaking off. This shows the strength of introducing detection dropouts to the training environment. As long as the target is far away and not maneuvering unpredictably, a long detection dropout can still result in a successful pursuit and possibly regain of target lock if the target being chased is real. 

Engagement 5 is interesting because the range estimate is actually very low upon initialization (12$\,$m), but then quickly climbs to 54$\,$m, only to drop all the way to 4$\,$m when the target is being intercepted. The perception model possibly locked onto different illusionary targets mid-engagement, reducing the reliability of the KF estimate. This is also reflected by the other plots in figure \ref{fig_KF1Hzlogs}, which show a climbing target speed estimate, as if the target is accelerating away. Also, the position and velocity estimate variances are high compared to the actual range and velocity estimates. The detection dropout plot corroborates these findings as it shows that 2 seconds into the engagement, the target was lost for 1.7 seconds. After that it (re)acquires the real target and proceeds to intercept successfully. 

\subsection{Rudder behavior in RL}

\begin{figure}[!t]
\centering
\includegraphics[width=1\columnwidth]{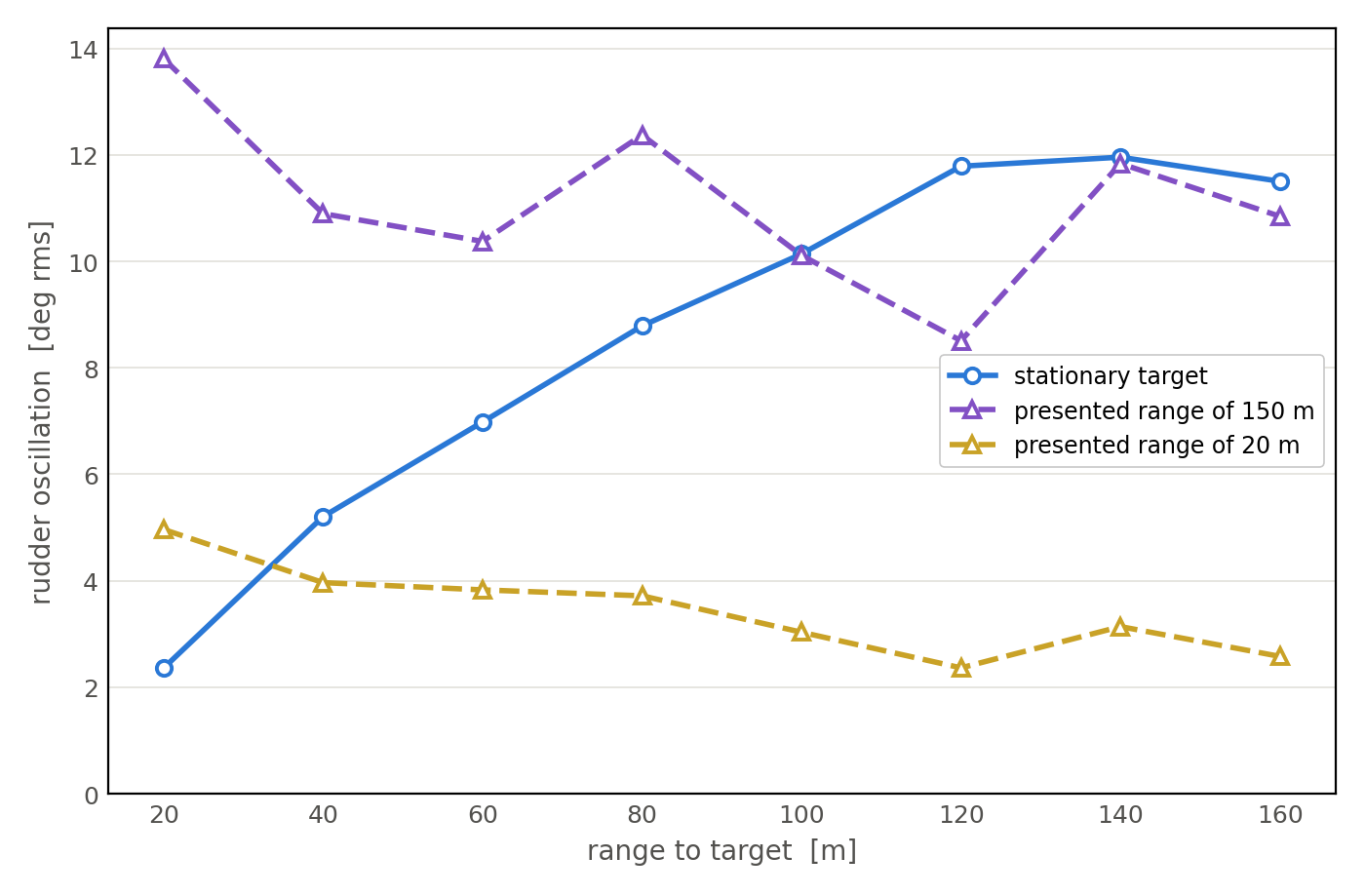}
\caption{The root mean square (rms) in degrees of rudder deflection versus the range between the target and the interceptor. The blue line shows rudder deflection amplitude against real range as seen when intercepting stationary targets. The purple and yellow dashed lines show the rudder deflection amplitude at various ranges based on presented artificial range estimates.}
\label{fig_osc_vs_range}
\end{figure} 

The RL policy showed significant oscillation of the rudder, both in simulation and in real flight, resulting in an oscillatory yaw rate that decreases in amplitude as the target is approached. This was investigated and seems to be correlated with perceived range to the target as shown in figure \ref{fig_osc_vs_range}. Rudder oscillation amplitude, quantified as the root mean square rudder deflection in degrees, correlated nearly linearly with perceived range. If the policy is presented artificial high or low range estimates, the rudder oscillation amplitude will be high or low respectively, regardless of actual range. 

\section{Discussion}

\subsection{Performance}

In simulation, RL outperformed the classical controllers under almost all circumstances.
Against fast, non-maneuvering targets, PP overtakes RL. At a target speed of 20$\,$m$\cdot$s$^{-1}$ it achieves 91.8 percent interception rate against the policy's 70.2 percent. The policy's training distribution reached only 12$\,$m$\cdot$s$^{-1}$ target speed per axis, favoring to leave a velocity margin for accelerating targets. Hence, this condition lies partly outside the training distribution.
RL commanded rudder independently, PP and PN only use rudder to coordinate turns. PN underperformed RL but mostly matched PP, only being outperformed in the later stages of the curriculum where perception is severely limited by noise. This is mainly due to the sensor being a strapdown monocular camera. Traditionally, PN is used on missiles with seeker heads which ensure the sensor is pointing at the target. That assumption is not valid for the interceptor used in this work. Notably, PN is implemented in a range-invariant fashion. This differs from textbook approaches \cite{yanushevsky2018modern}\cite{zarchan2012tactical} which scale the commands by closing velocity. This approach was initially attempted, but abandoned as it was observed that the closing velocity estimate from apparent bounding box dimensions was too noisy to properly scale the guidance commands. The performance of the baseline classical guidance algorithms also depends on the underlying cascaded autopilot that was manually tuned. 

In real flight, the RL policy achieved sim-to-real transfer. With virtual targets its performance showed high accuracy at a variety of initial geometries, attitudes and wind speeds. 
The performance was ultimately successfully validated by the interceptions of real, maneuvering fixed-wing MALE targets. Human operators had to intervene repeatedly to prevent an actual collision between the interceptor and its target.

\subsection{Perception}
Perception was a key challenge in this work. Initial results showed that simply providing bounding-box data as input was not sufficient to let the policy converge to reliable interception behavior in training. The Kalman filter resolved this by providing relative target state estimates. Additionally, the Kalman filter also made the classical controllers more robust. With the bounding-box only approach, these controllers would have no input to act on when the target left the FOV. However, the filter produces estimates even when the target is not detected. 

In training, the challenge was to simulate all real-life perception noise. Since the performance of the computer vision algorithm was not quantitatively known, the approach was to train the controller with as much noise as possible. Initially, mainly white noise in the pixels and frame-by-frame detection dropouts were provided, but this is easily compensated for by the filter. By clustering detections dropouts into larger chunks, it becomes harder for the filter to create reliable estimates. Consequently, the policy learns to behave conservatively. In real flight, the perception might be less noisy, meaning more optimal approaches could be flown. During interceptions of real targets, it was indeed observed that detection can drop out over prolonged periods. Figure \ref{fig_KF1Hzlogs} shows that during engagement 5 with a real target, there were no detections for at least 1.7 seconds at one point. Despite this, the policy was able to successfully intercept the target, showing robustness during perception degradation.

One noise source that was not simulated in training is the detection of false positives or even other potential wrong targets such as birds, clouds, or even high-altitude civilian aircraft. This was especially visible during real target engagement 1, where the RL controller was engaged whilst locking onto a different target. The policy currently does not discern between a real target and something else that still results in detections feeding into the Kalman filter. 

The Kalman filter currently uses all detections with a confidence score above 0.3. Using these scores as weights could actually result in a better prediction with a lower uncertainty. In the current implementation, the detection algorithm's confidence score is lost during the filtering. 

Range estimation is based on the Kalman Filter back projecting bounding box dimensions onto a known wingspan of 1.7 meters. While the simulation and training environments randomize the actual target dimensions around this nominal value, the range estimate will breakdown for targets that are much smaller or larger. This does not necessarily mean that performance degrades however. In real flight, spurious odd-sized detections did not result in sudden changes in course.

\subsection{Flight model}
To effectively train a policy that is transferable to a real platform, robustness is needed rather than accuracy. The dynamics model used in training should cover the real life physics, but also not over-complicate things. To identify the aircraft parameters, three flights were flown, with two having an AOA vane mounted on the aircraft. The results in table \ref{tab:aero_derivatives} are not perfect, with no correlation scores above 0.8. This could probably be greatly improved by flying more flights with an AOA vane and specifically exciting different axis independently. Additionally, a different regression method could be used. OLS was used here for simplicity, but figure \ref{fig_Autocorrelation} showed that for lift and drag, the residuals are autocorrelating, which violates the OLS assumption that the noise is independent. Some non-linear identification method could possibly generate a more valid estimate. A better thrust model could potentially also resolve the autocorrelation problem. 

However, the scope of this work was sim-to-real transfer. Rather than focusing on creating a very high fidelity simulation, domain randomization and other techniques were successfully used to overcome gaps between simulation and real flight.

\subsection{Training}

Once a policy was successfully used to intercept stationary virtual targets in real flight, further development and training of other RL policies was halted. This was not because it could not have been improved, but rather the choice was deliberately made to test the policy with as many different real flight experiments as possible whilst keeping performance and limitations comparable across experiments. 

One of the downsides of reinforcement learning is that it requires training of the neural network, which can take a long time. This makes iteration cycles long. To alter the observations of the policy or the curriculum usually means retraining the policy from scratch, costing multiple hours of CPU time. Transfer learning allows the weights of one policy to be transferred to the next. This is helpful when quickly testing late-stage curriculum changes, where the policy already should have learned how to fly. However, to obtain reproducible results, the curriculum and parameters should not be altered mid-training. 

The analyses in sections \ref{sec: single episode analysis} and \ref{sec: multiple episode analysis} were performed after a bug fix in the training environment. Specifically, the pre-fix training environment mixed aircraft acceleration in the inertial frame with aircraft acceleration in the body frame (equations \ref{eq: v_body} and \ref{eq: v_inertial}). This lead to the additional term $- \boldsymbol{\omega}_b\times\mathbf{v}_b$ fighting the aircraft during turning in the training environment. In other words, the RL policy was trained in an environment in which turning was approximately twice as hard as would be realistic. This explains why the performance in the corrected environment is lower than what is suggested by the curriculum learning progression in figure \ref{fig_curriculumprogress}, which is in the environment where the bug is still present. The bug was discovered after real flight tests were successfully performed and hence instead of retraining a new policy, the current policy was maintained for comparability at the cost of performance.
Nonetheless, this error shows the robustness of the RL policy. Not only did performance degrade only slightly in simulation after the fix, the policy also managed to intercept targets in a real environment, showcasing sim-to-real transferability even when trained in an imperfect physics model. The performance of the policy in the original training environment is included in appendix \ref{app:pre-bug-fix}, which unsurprisingly shows higher performance across all domains.

One interesting but unexplored idea to potentially improve performance of the policy is to introduce a reward for producing a correct relative target state estimate. Currently the policy is driven towards perception by keeping the camera aimed at the target and the target in the FOV, but this might not always be the optimal solution. The reward function could stimulate low covariances in the state estimates or even penalize proportional to the error in the state estimate. 

\subsection{Transferability}
Initially, time since start of the engagement was also provided to the policy, but this proved to hinder sim-to-real transfer. During real flight testing, the controller is running continuously, but its authority over the aircraft is being toggled on and off by a human operator. Time since start of the engagement quickly blows up beyond the training envelope, leading to unpredictable behavior. Additionally, the parameter adds no value to the agent's decision making.  

By using a framework that trains policies with up to 30 percent domain randomization in simulation, the controller becomes robust to sim-to-real gaps in the flight model. This controller is end-to-end, meaning even low-level dynamics needed to be simulated.

\subsection{RL controller flight behavior}

One of the more surprising observations during this research was the continuous rudder oscillations produced by the policy both in simulation and in real flight. The aircraft is visibly heavily yawing back and forth, even in level wings flight, which means it is unlikely to be attributable to turn coordination. The oscillatory yawing fades as perceived range shrinks, as if the interceptor is straightening itself out before impact.
At first it was thought that the policy might have learned to triangulate for better range estimates, however replacing the noisy camera with an ideal one reduced KF position error to nearly zero, whilst rudder oscillation was unchanged.
Then it was hypothesized that perhaps the oscillatory motion aids in reacquisition of target lock after a detection dropout, as the oscillation effectively widens the FOV. However this was unlikely, as reacquisition seemed independent of rudder oscillation. 
It was thought that maybe the policy uses the generated sideslip as an airbrake. This could keep throttle at a maximum, which is preferred to avoid its ramp up, whilst slowing closing velocity allowing for preciser maneuvering. This theory was refuted as airspeeds were comparable with or without the oscillations. 
It could also have been an artifact of the pre-correction training environment physics, which artificially made banking a lot harder, but this appears unlikely as removing rudder actually reduced performance less in the pre-correction training environment than in the corrected training environment.
Since replacing the RL rudder control with the sideslip damper used in the cascaded autopilot for the baseline controllers did not reduce the interception rate significantly, it was hypothesized that the behavior must be optimizing some reward, however this was not confirmed.
In the end, this analysis also highlights one of the downsides of using neural networks for control: its behavior can be a black box to an extent.

\subsection{Future work}
This work has demonstrated fixed-wing, vision-based interception using an RL controller in real flight. The focus here was to demonstrate the proof-of-concept; once a policy was successfully flight tested, development of other policies stopped. Future work could continue this research. The approach in this study was to create a very conservative policy, with the highest chance of successful sim-to-real transfer, but this is at the cost of performance. It would be very interesting to see exactly when the policy fails to transfer to reality. In the current setup, policies are trained with up to 30 percent domain randomization of the aircraft parameters. This value is based on other work, but such high domain randomization also degrades performance and it is not certain that it is needed to such a degree to ensure sim-to-real transfer. Additionally, the policy described has been steered significantly by including large incentives to keep the target in sight. While this makes the policy more robust in the case of lots of camera noise or bad computer vision, it also hinders interception performance. It would be insightful to see where the limits of all of these considerations lie and where the policy breaks down in real flight. 

Previous work has investigated quadrotor-based interception in simulation. This work now shows it is possible both in simulation and real flight for fixed-wing drones. This indicates that the possibilities are not gated by the platform used. The work outlined here could be extended for other platforms like tailsitters or X-wing interceptors. 

One caveat of the work demonstrated here is that upon initializing an interception, for both the RL and classical controllers, the target needs to be detected. The KF produces a relative target position estimate. If there are no prior detections, the KF can not produce a position estimate, meaning the controllers have nothing to act on. Having the controller acquiring the target autonomously would greatly improve combat readiness. 

The classical controllers were not tested in real flight. While their they were outperformed in simulation, for a complete comparison they would also need to be tested in real flight. 

Currently, the relative state of the target is determined solely by visual observations. For midcourse guidance, a rough, ground-based radar estimate of the relative position of the target can be used. Future controllers could fuse such estimates with onboard detections, leading to a more robust behavior and also allowing autonomous target (re)acquisition for terminal guidance. For terminal guidance, monocular cameras provide a cheap and established way to detect targets. However, other sensors such as acoustic sensors, stereovision cameras, event-based cameras and infrared sensors can also be used. Multiple monocular cameras of different focal lengths can even be used to fuse a sharp focused image with a wide, unsharp image. This can make target reacquisition much easier in case of a missed interception. 

\section{Conclusion}

In this work, an end-to-end neural network was trained using reinforcement learning to control a fixed-wing interceptor aircraft tasked with taking down an uncooperative target. The policy maps filtered, monocular vision observations directly to actuator commands. In simulation it outperformed baseline classical guidance algorithms. The controller achieved sim-to-real transfer, showcasing a comparable performance during real flight experiments. The system was ultimately successfully validated by intercepting real fixed-wing, medium altitude, long endurance targets, showing the potential of autonomous, intelligent interceptors in a modern air-defense context. 

\newpage
\appendices
\section{Aerodynamics model}
\label{app:dynamics_model}

This section briefly describes the aerodynamics model that was used in the Python-based simulation and training environments.

\subsubsection{State vector}
The state vector $\mathbf{x}$ can be defined as:
\begin{equation}
\begin{aligned}
\mathbf{x}^\top &=
\begin{bmatrix}
\mathbf{p}_i & \mathbf{v}_i & \mathbf{q} & \boldsymbol{\omega}_b & \delta_e & \delta_a & \delta_r & \delta_t
\end{bmatrix} \\
\end{aligned}
\end{equation}
\noindent where $\mathbf{p}_i$, $\mathbf{v}_i$ $\in \mathbb{R}^3$ are the position and velocities in the North-East-Down (NED) inertial frame, $\mathbf{q} = \begin{bmatrix} q_0 & q_1 & q_2 & q_3\end{bmatrix}^\top$, is the unit quaternion attitude representation, $\boldsymbol{\omega}_b \in \mathbb{R}^3$ are the angular rates in the body frame, $\delta_e, \delta_a, \delta_r \in$ [-1, 1] are the normalized control surfaces actuator states which are scaled by the actuator deflection limits in radians and $\delta_t \in$ [0, 1] is the normalized throttle state. All actuator states undergo first-order lag and hence: $\dot{\delta}_j = \frac{\delta_{j,\mathrm{cmd}} - \delta_j}{\tau_j}, \qquad j \in \{e, a, r, t\}$, with $\tau_{e,a,r} = 0.05$ s and $\tau_t = 0.4$ s.

\subsubsection{Translational dynamics}
The translational dynamics in the body frame can be defined as:
\begin{equation}
    \dot{\mathbf{v}}_b = \tfrac{1}{m}\mathbf{F}\mathrm{_{tot}} - \boldsymbol{\omega}_b\times\mathbf{v}_b.
    \label{eq: v_body}
\end{equation}
\noindent Here, $m$ is the mass of the aircraft and $F_{tot}$ is the total force acting on the body.
The simulator propagates position and velocity in NED inertial coordinates, giving:
\begin{equation}
        \dot{\mathbf{p}}_i = \mathbf{v}_i, \hspace{3mm} \dot{\mathbf{v}}_i = \tfrac{1}{m}\mathbf{R}{_{b/i}}^{\top}\mathbf{F}\mathrm{_{tot}}.
        \label{eq: v_inertial}
\end{equation}

\noindent Vectors are resolved either in the NED inertial frame, whose axes point north, east and downward, or in the body frame, whose axes $x_b$, $y_b$, $z_b$ point forward along the fuselage, out the right wing and downward respectively. The corresponding body-frame velocity components are denoted $u$, $v$ and $w$:
\begin{equation}
\mathbf{v}_b = \mathbf{R}_{b/i}\,\mathbf{v}_i =
\begin{bmatrix} u & v & w \end{bmatrix}^\top, \quad
V = \|\mathbf{v}_b\|
\end{equation}

\noindent The rotation matrix $R_{b/i}$ maps the inertial frame to the body frame:
\begin{equation}
\mathbf{R}_{b/i} =
\begin{bmatrix}
1 - 2(q_2^2 + q_3^2) & 2(q_1 q_2 + q_0 q_3) & 2(q_1 q_3 - q_0 q_2) \\[1pt]
2(q_1 q_2 - q_0 q_3) & 1 - 2(q_1^2 + q_3^2) & 2(q_2 q_3 + q_0 q_1) \\[1pt]
2(q_1 q_3 + q_0 q_2) & 2(q_2 q_3 - q_0 q_1) & 1 - 2(q_1^2 + q_2^2)
\end{bmatrix}.
\label{eq:Rbi}
\end{equation}

\noindent The matrix is orthonormal, so its inverse is its transpose,
$\mathbf{R}_{b/i}^{-1} = \mathbf{R}_{b/i}^{\top} = \mathbf{R}_{i/b}$.

The total force is comprised of the aerodynamic forces, the thrust generated by the propeller and the force due to gravity: 
\begin{equation}
\mathbf{F}_\mathrm{tot} = \mathbf{F}_\mathrm{aero} + \mathbf{F}_\mathrm{thrust} + \mathbf{F}_\mathrm{gravity},
\end{equation}
\noindent where:
\begin{equation}
\mathbf{F}_{\mathrm{gravity}} =
R_{b/i}
\begin{bmatrix}
0 & 0 & mg
\end{bmatrix}^\top,
\end{equation}
\begin{equation}
\mathbf{F}_{\mathrm{thrust}} =
\begin{bmatrix}
\delta_t\, T_{\max}\sqrt{\dfrac{\rho}{\rho_0}}\left(1 - \hat{V}^{\,2}\right) & 0 & 0
\end{bmatrix}^\top,
\end{equation}
\begin{equation}
\mathbf{F}_\mathrm{aero}=\bar{q}S \hspace{-1mm}\begin{bmatrix}
\cos\alpha \cos\beta & -\sin\beta & -\sin\alpha \cos\beta \\
\cos\alpha \sin\beta & \cos\beta & -\sin\alpha \sin\beta \\
\sin\alpha & 0 & \cos\alpha
\end{bmatrix}\hspace{-2mm}
\begin{bmatrix} - C_D \\ \phantom{-} C_Y \\ - C_L \end{bmatrix}.
\end{equation}
\noindent Here, $g$ is the gravitational acceleration. $T_{max}$ is the maximum thrust achievable at full throttle with zero airspeed. $\hat{V} = \min\!\left(\frac{V}{V_{\max}},\, 1\right)$, with $V$ being the airspeed and $V_{\max}$ the airspeed at which thrust is zero. $\rho$ is the air density (ISA troposphere model) and $\rho_0 = 1.225\ \mathrm{kg\,m^{-3}}$ is the reference air density at sea level. In this research, $\sqrt{\dfrac{\rho}{\rho_0}}$ is effectively 1 due to the low altitude testing (max 120 m). $S$ is the wing surface area and $\bar q$ is the dynamic pressure. $\alpha = \arctan\frac{w}{u}$ is the Angle Of Attack (AOA), $\beta = \arcsin\frac{v}{V}$ is the sideslip angle. $C_L, C_D$ and $C_Y$ are the non-dimensional lift, drag and sideforce coefficients and can be described as: 
\begin{equation}
C_L = C_{L_0} + C_{L_\alpha}\,\alpha + C_{L_{\delta_e}}\,\delta_e,
\end{equation}
\begin{equation}
C_D = C_{D_0} + kC_L^2,
\end{equation}
\begin{equation}
C_Y = C_{Y_\beta}\,\beta + C_{Y_{\delta_r}}\,\delta_r.
\end{equation}
\noindent Here, $k$ is the induced drag factor. $C_{L_0}, C_{L_\alpha}, C_{L_{\delta_e}}$ are coefficients that describe how the lift is shaped by AOA and elevator deflection. $C_{D_0}$ is the zero-lift drag coefficient. $C_{Y_\beta},  C_{Y_{\delta_r}}$ are coefficients that shape how the sideforce is shaped by sideslip angle and rudder deflection.

\subsubsection{Rotational dynamics}
The rotational dynamics are given by:
\begin{equation}
    \dot{\boldsymbol{\omega}}_b = \mathbf{J}^{-1} \Big( \mathbf{M}_b - \boldsymbol{\omega}_b \times (\mathbf{J} \, \boldsymbol{\omega}_b) \Big),
\end{equation}
\noindent where $\mathbf{J}$ is the inertia tensor of the aircraft and $\mathbf{M}_b$ is the total moment vector which is approximated to contain only aerodynamic moments:
\begin{equation}
\mathbf{M}_b = \bar qS
\begin{bmatrix}
bC_l & cC_m & bC_n
\end{bmatrix}^\top.
\end{equation}
\noindent Here, $b$ is the wingspan and $c$ is the mean aerodynamic chord. $C_l, C_m$ and $C_n$ are the dimensionless aerodynamic moment coefficients that describe how airflow generates rotational moments about the body axes: 
\begin{align}
C_l &= C_{l_0}
+ C_{l_\beta}\beta
+ C_{l_p}\frac{p b}{2V}
+ C_{l_r}\frac{r b}{2V}
+ C_{l_{\delta_a}}\delta_a
+ C_{l_{\delta_r}}\delta_r,
\end{align}
\begin{align}
C_m &= C_{m_0}
+ C_{m_\alpha}\alpha
+ C_{m_q}\frac{q c}{2V}
+ C_{m_{\delta_e}}\delta_e,
\end{align}
\begin{align}
C_n &= C_{n_0}
+ C_{n_\beta}\beta
+ C_{n_p}\frac{p b}{2V}
+ C_{n_r}\frac{r b}{2V}
+ C_{n_{\delta_a}}\delta_a
+ C_{n_{\delta_r}}\delta_r.
\end{align}
\noindent Here, $p, q, r$ are the body angular rates. 

\subsubsection{Attitude kinematics}
The attitude kinematics are defined by:
\begin{equation}
\dot{\mathbf{q}} = \tfrac{1}{2}
\begin{bmatrix}
-q_1 & -q_2 & -q_3 \\
\phantom{-}q_0 & -q_3 & \phantom{-}q_2 \\
\phantom{-}q_3 & \phantom{-}q_0 & -q_1 \\
-q_2 & \phantom{-}q_1 & \phantom{-}q_0
\end{bmatrix}
\boldsymbol{\omega}_b .
\end{equation}

\subsubsection{V-tail configuration}
The interceptor built for this study has a V-tail ruddervator configuration instead of a conventional rudder and elevator. Hence:
\begin{align}
\delta_{e,\text{eff}} &= \frac{1}{2}\cos(\Gamma)\,(\delta_{v_L} + \delta_{v_R}), \\
\delta_{r,\text{eff}} &= \frac{1}{2}\sin(\Gamma)\,(\delta_{v_L} - \delta_{v_R}),
\end{align}
\noindent where $\Gamma$ is the dihedral angle and $\delta_{v_L}, \delta_{v_R} \in $ [-1, 1] are the normalized left and right ruddervator deflection.

\section{Ordinary Least Squares regression}
\label{app:OLS}

Ordinary Least Squares describes the regression problem as follows:
\begin{equation}
    \mathbf{z}=\mathbf{X}\bm{\theta} + \mathbf{v}.
\end{equation}
\noindent Here, $\mathbf{z}$ is the vector of the measured dependent variables. $\mathbf{X}$ is the regressor matrix formed from measured independent variables. $\bm{\theta}$ is the unknown true parameter vector. Finally, $\mathbf{v}$ is the residual vector.
The OLS parameter estimate $\hat{\bm{\theta}}$ is obtained by minimizing the sum of squared residuals:
\begin{equation}
    J(\bm{\theta}) = (\mathbf{z}-\mathbf{X}\bm{\theta})^T(\mathbf{z}-\mathbf{X}\bm{\theta}).
\end{equation}
\noindent The necessary condition for the minimum is: 
\begin{equation}
    \mathbf{X}^T\mathbf{z} =\mathbf{X}^T\mathbf{X}\hat{\bm{\theta}}.
\end{equation}
\noindent Solving this yields the solution:
\begin{equation}
    \hat{\bm{\theta}}=(\mathbf{X}^T\mathbf{X})^{-1}\mathbf{X}^T\mathbf{z}.
\end{equation}
\noindent This OLS parameter vector estimate gives the best least-squares fit of the linear model to the gathered flight data and was consequently used to create a high-fidelity simulation to train the RL controller.

\section{Kalman Filter}
\label{app:KF}
This section describes the Kalman filter used to filter the bounding-box signals and output relative state estimates. 

\subsubsection{State vector and process noise model}

The filter tracks the target's dynamic state:
\begin{equation}
  \bm{x}_k =
  \begin{bmatrix}
    \bm{p}_k \\ \dot{\bm{p}}_k \\ \ddot{\bm{p}}_k
  \end{bmatrix}
=
  \begin{bmatrix}
    p_N\hspace{-1mm}&\hspace{-1mm}p_E\hspace{-1mm}&\hspace{-1mm} p_D \hspace{-1mm}& \hspace{-1mm}v_N \hspace{-1mm}&\hspace{-1mm} v_E \hspace{-1mm}&\hspace{-1mm} v_D \hspace{-1mm}&\hspace{-1mm} a_N \hspace{-1mm}&\hspace{-1mm} a_E \hspace{-1mm}&\hspace{-1mm} a_D
  \end{bmatrix}^{\!\top},
  \label{eq:state}
\end{equation}
where $\bm{p}_k,\dot{\bm{p}}_k,\ddot{\bm{p}}_k\in\mathbb{R}^3$ are the target
position, velocity and acceleration in NED. By including acceleration in the state vector, the filter can follow a maneuvering target. By working in a NED inertial frame, the target's motion is decoupled from the interceptor's ego-motion.

Between measurements, the state evolves under a discrete constant acceleration model with additive Gaussian process noise,
\begin{equation}
  \bm{x}_k = \boldsymbol{\Phi}\,\bm{x}_{k-1} + \bm{w}_{k-1},
  \qquad \bm{w}_{k-1}\sim\mathcal{N}(\bm{0},\bm{Q}),
  \label{eq:process}
\end{equation}
with the sampling interval $\Delta t = 1/f_c = 0.02\,\mathrm{s}$ matching the
control and perception frequency $f_c=50\,\mathrm{Hz}$. In training, $f_c$ is randomized between 40 and 50 Hz. With $\bm{I}_3$ and $\bm{0}_3$ the $3\times3$ identity and zero matrices, the transition matrix $\boldsymbol{\Phi}$ is
\begin{equation}
  \boldsymbol{\Phi} =
  \begin{bmatrix}
    \bm{I}_3 & \Delta t\,\bm{I}_3 & \tfrac{1}{2}\Delta t^{2}\,\bm{I}_3 \\[2pt]
    \bm{0}_3 & \bm{I}_3           & \Delta t\,\bm{I}_3 \\[2pt]
    \bm{0}_3 & \bm{0}_3           & \bm{I}_3
  \end{bmatrix}.
  \label{eq:F}
\end{equation}
Real targets do not obey constant acceleration, hence process noise is introduced to explain the change of acceleration (jerk). The process noise is obtained by treating jerk as a zero-mean white process of intensity (power spectral density) $q$ and
integrating over $\Delta t$:
\begin{equation}
  \bm{Q} = q
  \begin{bmatrix}
    \tfrac{\Delta t^{5}}{20}\bm{I}_3 & \tfrac{\Delta t^{4}}{8}\bm{I}_3 & \tfrac{\Delta t^{3}}{6}\bm{I}_3 \\[4pt]
    \tfrac{\Delta t^{4}}{8}\bm{I}_3  & \tfrac{\Delta t^{3}}{3}\bm{I}_3 & \tfrac{\Delta t^{2}}{2}\bm{I}_3 \\[4pt]
    \tfrac{\Delta t^{3}}{6}\bm{I}_3  & \tfrac{\Delta t^{2}}{2}\bm{I}_3 & \Delta t\,\bm{I}_3
  \end{bmatrix},
  \qquad q = 5\times10^{-4}.
  \label{eq:Q}
\end{equation}
\noindent $\bm{Q}$ is the covariance of the error that the prediction model makes in a single step. 

\subsubsection{Measurement model}

Only target pixel coordinates are used, not their temporal derivatives. Each detection where $c \geq 0.3$ is converted to a measured NED position $\bm{m}_k$. $c_x,c_y$ and
normalized width $w$ are back projected through the pinhole model using the image size $W_{\mathrm{img}}\times H_{\mathrm{img}}=1280\times720$ px, and the nominal target wingspan $W=1.7\,\mathrm{m}$ using equations \ref{eq:pinhole1}, \ref{eq:focallength} and \ref{eq:pinhole2} yielding the camera-frame point $\bm{r}_{\mathrm{cam}}=[X,Y,Z]^\top$. This is rotated into the body frame and
then into NED and translated by the interceptor's own NED position $\bm{p}_i$ using the body-to-NED rotation $\mathbf{R}_{i/b}$
formed from the autopilot attitude quaternion $\bm{q}$:
\begin{equation}
  \bm{m}_k
  = \bm{p}_i + \mathbf{R}_{i/b}\,\mathbf{R}_{b/c}\,\bm{r}_{\mathrm{cam}},
  \qquad \hspace{-6mm}
  \mathbf{R}_{b/c} =
  \begin{bmatrix} 0&0&1\\ 1&0&0\\ 0&1&0 \end{bmatrix}.
  \label{eq:meas_constr}
\end{equation}
The resulting measurement is linear in the state
\begin{equation}
  \bm{m}_k = \bm{H}\,\bm{x}_k + \bm{\nu}_k,
  \qquad
  \bm{\nu}_k\sim\mathcal{N}(\bm{0},\bm{R}_k),
  \label{eq:obs}
\end{equation}
\noindent with $\bm{H} = \begin{bmatrix} \bm{I}_3 & \bm{0}_{3\times6}\end{bmatrix}$ such that velocity and acceleration are never measured directly but instead are estimated from the sequence of position fixes through the model in \eqref{eq:F}. $\bm{\nu}_k$ represents the zero mean Gaussian measurement noise with covariance $\bm{R}_k$. 

\subsubsection{Range-dependent measurement covariance}

Range is recovered from apparent size and is far less certain than bearing, with an uncertainty that grows with distance. This is modeled by constructing $\bm{R}_k$ in a LOS frame and rotating it into NED. With range $d=\lVert\bm{m}_k-\bm{p}_i\rVert$ (floored at $0.5\,\mathrm{m}$),
the lateral and depth standard deviations are
\begin{equation}
    \;\sigma_{\mathrm{lat}}
    =\Big|\tfrac{\partial X}{\partial u_c}\Big|\sigma_c
    =\frac{Z}{f_x}\sigma_c
    =\frac{d(\sigma_{c,0}+\sigma_{c,1}d)}{f_x}\;
\end{equation}
and
    
\begin{equation}
    \;\sigma_{\mathrm{depth}}
    =\Big|\tfrac{\partial Z}{\partial w_{px}}\Big|\sigma_s
    =\frac{Z^{2}}{f_x W}\sigma_s
    =\frac{d^{2}(\sigma_{s,0}+\sigma_{s,1}d)}{f_x W},\;
\end{equation}
with $\sigma_{c,0}=3.0\,\mathrm{px}$, $\sigma_{c,1}=0.05\,\mathrm{px/m}$,
$\sigma_{s,0}=2.0\,\mathrm{px}$, $\sigma_{s,1}=0.02\,\mathrm{px/m}$ and $Z \approx d$. The three sources of error are assumed to be uncorrelated. The covariance in the LOS frame $\bm{R}^{\mathrm{LOS}}_k$ is therefore:
\begin{equation}
    \bm{R}^{\mathrm{LOS}}_k =
    \begin{bmatrix}
    \sigma_{\mathrm{depth}}^{2} & 0 & 0 \\
    0 & \sigma_{\mathrm{lat}}^{2} & 0 \\
    0 & 0 & \sigma_{\mathrm{lat}}^{2}
    \end{bmatrix}
    \label{eq:Rlos}
\end{equation}

\noindent and is mapped into NED by
\begin{equation}
  \bm{R}_k = \bm{L}\,\bm{R}^{\mathrm{LOS}}_k\,\bm{L}^{\top},
  \qquad
  \bm{L} = \begin{bmatrix} \hat{\bm{r}} & \bm{v}_1 & \bm{v}_2 \end{bmatrix},
  \label{eq:R}
\end{equation}
where $\hat{\bm{r}}=(\bm{m}_k-\bm{p}_i)/d$ is the LOS unit vector and $\bm{v}_1,\bm{v}_2$ are two unit vectors orthogonal to each other and to $\hat{\bm{r}}$, completing an orthonormal basis. Bearing can be trusted from few measurements, whereas range needs a longer measurement time to be trusted.

\subsubsection{Initialisation}

On the first detection that exceeds the confidence threshold, the filter is seeded from the measurement, with zero velocity and acceleration and large priors on the unobserved states:
\begin{equation}
  \hat{\bm{x}}_{0} =
  \begin{bmatrix} \bm{m}_0 \\ \bm{0} \\ \bm{0} \end{bmatrix},
  \qquad
    \bm{P}_{0} = \begin{bmatrix} \bm{R}_0 & \bm{0} & \bm{0} \\ \bm{0} & \sigma_{v_0}^2\bm{I}_3 & \bm{0} \\ \bm{0} & \bm{0} & \sigma_{a_0}^2\bm{I}_3 \end{bmatrix},
  \label{eq:init}
\end{equation}

\noindent with $\sigma_{v_0}=5\,\mathrm{m/s}$ and $\sigma_{a_0}=2\,\mathrm{m/s^2}$, so
that the first updates are dominated by the measurements. 

\subsubsection{Prediction step}

At every control step, irrespective of whether a detection is available, the
state and covariance are propagated with \eqref{eq:F}--\eqref{eq:Q}:
\begin{align}
  \hat{\bm{x}}_{k\mid k-1} &= \boldsymbol{\Phi}\,\hat{\bm{x}}_{k-1\mid k-1},
  \label{eq:predict_x}\\[2pt]
  \bm{P}_{k\mid k-1} &= \boldsymbol{\Phi}\,\bm{P}_{k-1\mid k-1}\,\boldsymbol{\Phi}^{\top} + \bm{Q}.
  \label{eq:predict_P}
\end{align}

\subsubsection{Correction step}

When an accepted detection is available, the predicted state is corrected by
the measurement $\bm{m}_k$. The innovation (measurement residual) and its
covariance are
\begin{align}
  \tilde{\bm{y}}_k &= \bm{m}_k - \bm{H}\,\hat{\bm{x}}_{k\mid k-1},
  \label{eq:innov}\\[2pt]
  \bm{S}_k &= \bm{H}\,\bm{P}_{k\mid k-1}\,\bm{H}^{\top} + \bm{R}_k.
  \label{eq:S}
\end{align}
The Kalman gain weights the correction by the relative confidence in the
prediction and the measurement,
\begin{equation}
  \bm{K}_k = \bm{P}_{k\mid k-1}\,\bm{H}^{\top}\,\bm{S}_k^{-1},
  \label{eq:K}
\end{equation}
and the state and covariance are updated as
\begin{align}
  \hat{\bm{x}}_{k\mid k} &= \hat{\bm{x}}_{k\mid k-1} + \bm{K}_k\,\tilde{\bm{y}}_k,
  \label{eq:update_x}\\[2pt]
  \bm{P}_{k\mid k} &= \left(\bm{I}_9 - \bm{K}_k\,\bm{H}\right)\bm{P}_{k\mid k-1}.
  \label{eq:update_P}
\end{align}
Although only position enters the innovation \eqref{eq:innov}, the off-diagonal terms of $\bm{P}_{k\mid k-1}$ built up by the dynamics in \eqref{eq:F} make velocity and acceleration observable, so the gain \eqref{eq:K} corrects all nine states.

\subsubsection{Detection handling and coasting}

When there is no detection or detection is below the confidence threshold, the correction step \eqref{eq:innov}--\eqref{eq:update_P} is skipped and only the prediction step \eqref{eq:predict_x}--\eqref{eq:predict_P} executes. The filter then coasts on its motion model, with $\bm{P}$ growing through $\bm{Q}$ to reflect the increasing uncertainty until the target is reacquired.

\subsubsection{Outputs}

The filter exposes the estimated target state and covariance,
\begin{equation}
  \hat{\bm{p}}_k = \hat{\bm{x}}_{k\mid k}^{(1:3)},\quad
  \hat{\bm{v}}_k = \hat{\bm{x}}_{k\mid k}^{(4:6)},\quad
  \hat{\bm{a}}_k = \hat{\bm{x}}_{k\mid k}^{(7:9)},\quad
  \bm{P}_{k\mid k}.
  \label{eq:outputs}
\end{equation}
These are expressed relative to the interceptor and rotated into the body frame,
\begin{equation}
  \bm{r}_{b,k} = \mathbf{R}_{i/b}\,(\hat{\bm{p}}_k - \bm{p}_i),
  \qquad
\dot{\bm{r}}_{b,k} = \mathbf{R}_{i/b}\,(\hat{\bm{v}}_k - \bm{v}_i)
  \label{eq:relative},
\end{equation}
\noindent from which the LOS angles and rates used by the controllers are computed.

\section{Actuator lag model}
\label{app:actuatorlag}
Each actuator is modeled as a first-order lag between the commanded deflection $\delta_{i,\mathrm{cmd}}$ and the achieved deflection $\delta_i$,
\begin{equation}
    \dot{\delta}_i = \frac{\delta_{i,\mathrm{cmd}} - \delta_i}{\tau_i},
    \qquad i \in \{e,a,r,t\},
    \label{eq:actuator}
\end{equation}
with $\tau_e=\tau_a=\tau_r=0.05\,$s for the flight control surfaces and $\tau_t=0.40\,$s for the propeller. 

\section{Classical guidance cascaded autopilot gains}
\label{app:autopilot_gains}
The gains of the cascaded autopilot used in the baseline classical guidance controller are given in table \ref{tab_autopilot_gains}. Every integrator is protected against windup by clamping its magnitude, to 0.35, 0.55 and 0.40 of full travel in the roll rate, pitch rate and sideslip loops respectively and to 30 m s$^{-2}$ in the normal acceleration loop. Accumulation is suspended whenever the surface command exceeds actuator limits and the current error would drive it further beyond the limits. Integration resumes as soon as the error reverses, so recovery from a saturated maneuver is not delayed by an integrator that must first unwind.

\begin{table}[!t]
\centering
\small
\caption{Gains and constants of the cascaded acceleration autopilot of figure
\ref{fig_autopilot_blockdiagram}. Surface commands are normalised to
$[-1,1]$, so gains acting on a surface are dimensionless.}
\label{tab_autopilot_gains}
\begin{tabular}{llll}
\toprule
Symbol & Value & Unit & Description \\
\midrule
\multicolumn{4}{l}{\textit{Roll-rate loop (inner, aileron)}} \\
$K_{ff,p}$  & 0.134 & s\,rad$^{-1}$   & rate feedforward, $1/p_{ss}$ \\
$K_{P,p}$  & 0.25  & s\,rad$^{-1}$   & proportional gain \\
$K_{I,p}$             & 0.15  & rad$^{-1}$      & integral gain \\
\addlinespace
\multicolumn{4}{l}{\textit{Pitch-rate loop (inner, elevator)}} \\
$K_{ff,q}$  & 0.300 & s\,rad$^{-1}$   & rate feedforward ($1/q_{ss} =$ 0.405) \\
$K_{P,q}$  & 0.70  & s\,rad$^{-1}$   & proportional gain \\
$K_{I,q}$             & 2.00  & rad$^{-1}$      & integral gain \\
\addlinespace
\multicolumn{4}{l}{\textit{Sideslip regulator (inner, rudder)}} \\
$K_{P,\beta}$         & 1.2   & rad$^{-1}$      & proportional gain \\
$K_{I,\beta}$         & 0.8   & rad$^{-1}$s$^{-1}$ & integral gain \\
$K_{D,r}$             & 0.30  & s\,rad$^{-1}$   & yaw damping about $r_{\mathrm{coord}}$ \\
\addlinespace
\multicolumn{4}{l}{\textit{Roll channel (outer)}} \\
$T_{\phi}$            & 0.15  & s               & roll time constant; $P = 1/T_{\phi}$ \\
\addlinespace
\multicolumn{4}{l}{\textit{Normal-acceleration channel (outer)}} \\
$K_{P,a}$             & 2.0   & --              & open-loop gain \\
$K_{I,a}$             & 4.0   & s$^{-1}$        & integral gain \\
\addlinespace
\multicolumn{4}{l}{\textit{Applying across loops}} \\
$V_{\mathrm{stab}}$   & 25.0  & m\,s$^{-1}$     & reference for $(V_{\mathrm{stab}}/V)^{2}$ schedule \\
$\tau_{f}$            & 0.08  & s               & low-pass on specific force \\
$f_{\min}$            & 1.47  & m\,s$^{-2}$     & roll demand frozen below this \\
\bottomrule
\end{tabular}
\end{table}

\section{Reward function coefficients}

Table \ref{tab:reward_coeffs} shows the reward function coefficients used in the training of the ultimate policy that was flown in the real flight tests.
\begin{table}[t]
\centering
\caption{Coefficients and parameters of the reward function.}
\label{tab:reward_coeffs}
\begin{tabular}{llr}
Symbol & Description & Value \\
\multicolumn{3}{l}{\textit{Term weights}}\\
$c_d$            & Distance-progress reward     & $1.0$ \\
$c_{\text{int}}$ & Interception bonus           & $0.5$ \\
$c_c$            & Camera-centering reward      & $0.4$ \\
$c_\Phi$         & Field-of-view penalty        & $1.5$ \\
$c_\Omega$       & Body-rate penalty            & $0.7$ \\
$c_u$            & Action-smoothness penalty    & $1.5$ \\
$c_\alpha$       & Angle-of-attack penalty      & $2.0$ \\
\multicolumn{3}{l}{\textit{Shaping parameters}}\\
$v_{cl,\text{safe}}$   & Safe closing speed          & $10\ \mathrm{m/s}$    \\
$\Omega_{r_{\max}}$    & Body-rate normalizer        & $2.0\ \mathrm{rad/s}$ \\
$\alpha_{\text{warn}}$ & AOA warning angle           & $10^\circ$            \\
$\alpha_{\max}$        & AOA cutoff limit            & $60^\circ$            \\
$K$                    & AOA penalty curvature       & $-3$                 \\
$\tau_{\max}$          & Target-loss horizon         & $200$ steps          \\
$R_{\max}$             & Return normalization budget & $1000$               \\
\end{tabular}
\end{table}

\label{app: reward function}

\section{Curriculum learning stages}

Tabel \ref{tab:curriculum} describes all the changes per stage during the training curriculum. Maximum range is deliberately lowered in the latest two stages, as it was that observed that the camera noise became too large at greater ranges to reliably detect the target.

\begin{sidewaystable}
\centering
\small
\caption{Curriculum schedule used in training the policy describing the variables that were changed for each stage.}
\label{tab:curriculum}
\begin{tabular}{l*{10}{c}}
Stage & 0 & 1 & 2 & 3 & 4 & 5 & 6 & 7 & 8 & 9 \\
\multicolumn{11}{l}{\textit{Target dynamics}}\\
Max spawn range [m]              & 60 & 70 & 80 & 90 & 100 & 110 & 120 & 120 & 100 & 80 \\
Trajectory\textsuperscript{a}    & CV & CV & CA & CA & CA & CA & CA & CA & CA & CA \\
Max vel.\ $xy$ [m/s]             & 1 & 2 & 3 & 4 & 5 & 6 & 7 & 8 & 9 & 10 \\
Max vel.\ $z$ [m/s]              & 0.1 & 0.2 & 0.3 & 0.4 & 0.5 & 0.6 & 0.7 & 0.8 & 0.9 & 1.0 \\
Max accel.\ $xy$ [m/s\textsuperscript{2}] & 0.0 & 0.1 & 0.2 & 0.3 & 0.4 & 0.5 & 0.6 & 0.7 & 0.8 & 1.0 \\
Max accel.\ $z$ [m/s\textsuperscript{2}]  & 0.0 & 0.0 & 0.0 & 0.0 & 0.0 & 0.1 & 0.1 & 0.2 & 0.2 & 0.3 \\
\multicolumn{11}{l}{\textit{Environment \& domain randomization}}\\
Airframe DR [\%]                 & 1 & 2 & 3 & 5 & 10 & 15 & 20 & 20 & 25 & 30 \\
Target-size rand.\ [\%]          & 0 & 0 & 0 & 0 & 5 & 10 & 10 & 15 & 17 & 20 \\
Control freq.\ [Hz]              & 44--46 & 43--47 & 42--48 & 41--49 & 40--50 & 40--50 & 40--50 & 40--50 & 40--50 & 40--50 \\
Wind speed max [m/s]             & 0.2 & 0.4 & 0.6 & 0.8 & 1.0 & 1.3 & 1.7 & 2.3 & 4.5 & 6.0 \\
Turbulence [m/s]                 & 0 & 0 & 0 & 0 & 0 & 0 & 0.10 & 0.20 & 0.40 & 0.60 \\
\multicolumn{11}{l}{\textit{Camera realism}}\\
Dropout mean [s]                 & 0.20 & 0.25 & 0.30 & 0.35 & 0.40 & 0.45 & 0.50 & 0.60 & 0.70 & 0.70 \\
Bbox bias std [--]               & 0.04 & 0.05 & 0.06 & 0.07 & 0.08 & 0.09 & 0.10 & 0.11 & 0.12 & 0.12 \\
Detection latency [s]            & 0.02 & 0.02 & 0.02 & 0.04 & 0.04 & 0.04 & 0.06 & 0.06 & 0.06 & 0.06 \\
\multicolumn{11}{l}{\textit{Training}}\\
Episode length [s]               & 20 & 20 & 25 & 25 & 30 & 30 & 35 & 40 & 45 & 45 \\
\end{tabular}

\vspace{2pt}
{\footnotesize \textsuperscript{a}CV = constant velocity; CA = constant
velocity + constant acceleration.}
\end{sidewaystable}

\label{app: curriculum}

\section{Aerodynamic aircraft coefficients}

In table \ref{tab:coefficients_in_sim}, the aerodynamic aircraft coefficients that were used in training of the RL policy are given. 

\begin{table}[ht]
\centering
\caption{Aerodynamic Stability and Control Derivatives}
\label{tab:aero_coefficients}
\begin{tabular}{llr}
\hline
\textbf{Category} & \textbf{Coefficient} & \textbf{Value} \\
\hline

\multirow{3}{*}{Lift}
    & $C_{L0}$              & 0.31 \\
    & $C_{L_\alpha}$        & 2.90 \\
    & $C_{L_{\delta_e}}$    & 0.50 \\

\hline
\multirow{2}{*}{Drag}
    & $C_{D0}$              & 0.12 \\
    & $k_{\mathrm{induced}}$& 0.18 \\

\hline
\multirow{2}{*}{Side Force}
    & $C_{Y_\beta}$         & -0.096244 \\
    & $C_{Y_{\delta_r}}$    & -0.18 \\

\hline
\multirow{4}{*}{Pitching Moment}
    & $C_{m0}$              & -0.011 \\
    & $C_{m_\alpha}$        & -0.18 \\
    & $C_{m_q}$             & -2.60 \\
    & $C_{m_{\delta_e}}$    & 0.30 \\

\hline
\multirow{5}{*}{Rolling Moment}
    & $C_{l_\beta}$         & -0.0170 \\
    & $C_{l_p}$             & -0.092 \\
    & $C_{l_r}$             & 0.075 \\
    & $C_{l_{\delta_a}}$    & 0.03 \\
    & $C_{l_{\delta_r}}$    & 0.006506 \\

\hline
\multirow{5}{*}{Yawing Moment}
    & $C_{n_\beta}$         & 0.02 \\
    & $C_{n_p}$             & -0.051 \\
    & $C_{n_r}$             & -0.035 \\
    & $C_{n_{\delta_a}}$    & 0.007 \\
    & $C_{n_{\delta_r}}$    & 0.028 \\

\hline
\end{tabular}
\label{tab:coefficients_in_sim}
\end{table}

\label{app:sysid}

\section{RL policy performance in the old training environment}

\begin{figure}[!t]
\centering
\includegraphics[width=1\columnwidth]{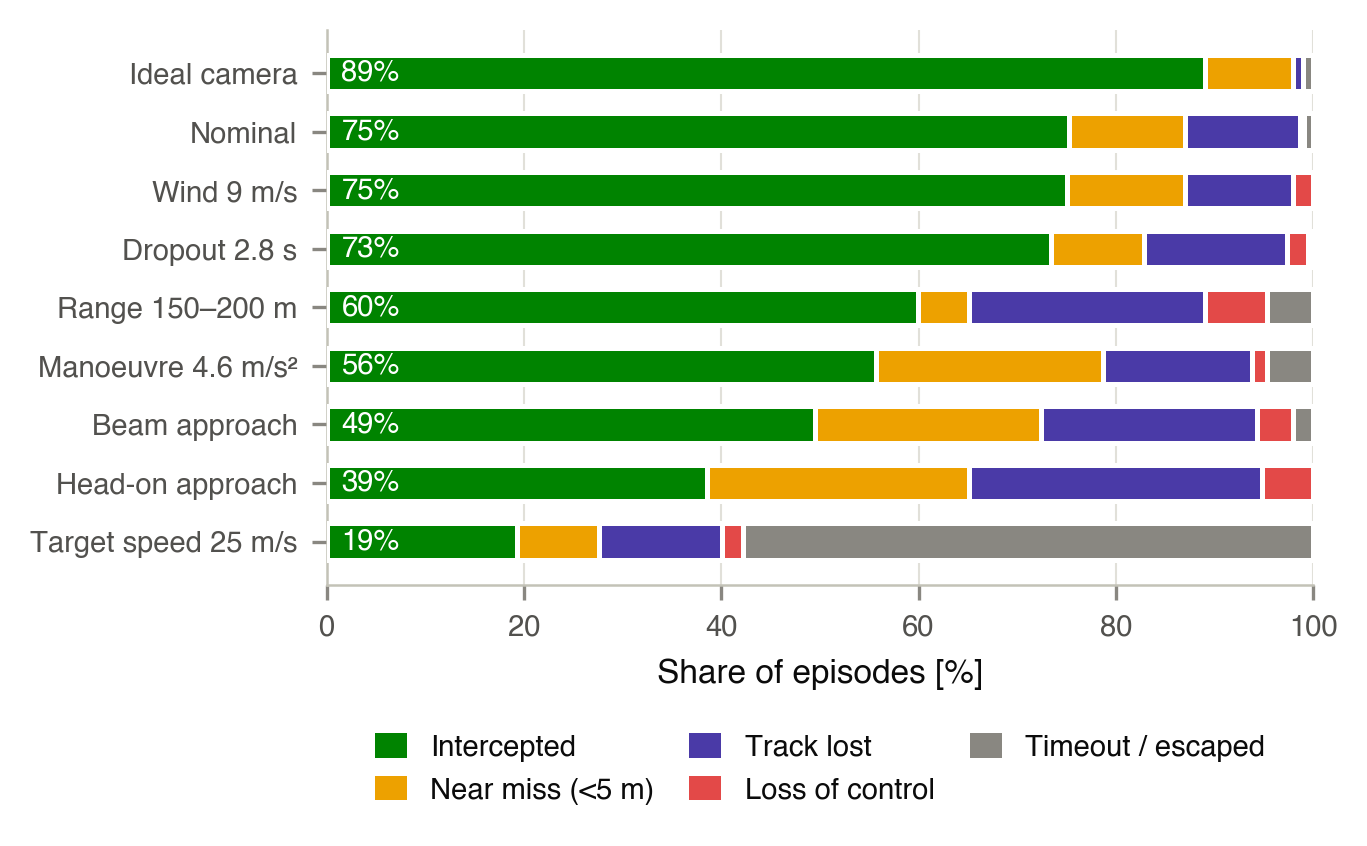}
\caption{A breakdown of the reasons episodes were terminated, categorized by condition. Green, purple, orange, red and grey represent the fraction of episodes terminated due to successful interception, target loss, near misses, loss of control and timeout. All conditions except for nominal are in some way beyond the training envelope, giving insight into the robustness of the policy. These variations were measured against stage 8 nominal conditions instead of stage 5 nominal conditions as in in figure \ref{fig_Outcomes_all}.}
\label{fig_outcomes_old}
\end{figure}

The performance of the RL policy in the old simulation environment (with the bug not fixed) is shown in figure \ref{fig_outcomes_old}.

\label{app:pre-bug-fix}

\section*{Acknowledgment}
This work was funded by TU Delft and Dremian BV. The latter provided the computer vision algorithm used to detect real targets in real flight experiments.

\bibliographystyle{unsrtnat}
\bibliography{references}

\end{document}